\documentclass{ametsocV6.1}

\usepackage{rotating}

\graphicspath{{figs/}}

\title{BG4Sea: Biogeochemical Seasonal Forecastability via Progressive Information Scaling}

\authors{Gabriela Martinez Balbontin,\aff{a,b}\correspondingauthor{Gabriela Martinez Balbontin, gmartinezbalbontin@mercator-ocean.fr}
Anastase Charantonis,\aff{c}
Dominique Bereziat,\aff{b}
and Stefano Ciavatta\aff{a}
}

\affiliation{\aff{a}{Mercator Ocean International}\\
\aff{b}{LIP6, Sorbonne Universite, Paris, France}\\
\aff{c}{INRIA (ARCHES), Paris, France}
}

\abstract{Marine biogeochemical forecasting is increasingly important for managing marine ecosystems and tracking the ocean's role in the carbon cycle, yet global, seasonal forecast products lag far behind physical oceanography, held back by the complexity of the processes involved and by data scarcity. We introduce BG4Sea, which to our knowledge is the first global, data-driven system to produce multivariate seasonal forecasts of the marine biogeochemical state. BG4Sea is a modular architecture with a column autoencoder that compresses the vertical column into a low-dimensional latent space, a latent forecaster that propagates this representation forward in time, a surface-forcing conditioner that injects physical boundary information via Feature-wise Linear Modulation (FiLM), and a horizontal-coupling module that incorporates neighboring-column context through cross-attention. The model is trained and evaluated on the global ocean reanalysis BIORYS4 (NEMO/PISCES), and produces six-month forecasts at $1/4^\circ$, monthly resolution for nutrients and oxygen, biology, and carbon-pool variables, outperforming persistence and climatology across most variables and lead times. We position BG4Sea as an interpretable baseline for future, more expressive approaches, and discuss predictability attribution to each component, alongside the model's structural limitations.} 

\begin{document}

\maketitle

%
%
%

%
\section{Introduction}

Biogeochemical (BGC) cycles govern the movement of chemical elements throughout the Earth's ecosystems. In the ocean, these cycles help regulate the Earth's climate and control how the ocean supports life~\citep{eppleyParticulateOrganicMatter1979, martinGlacialInterglacialCO21990, zehrNitrogenCyclingOcean2002}. To understand and predict the evolution of these systems, scientists use process-based models. These can be used to quantify the effects of global warming, acidification, deoxygenation, eutrophication, and intensifying resource use~\citep{berardi_21st-century_2020, gehlen_building_2015, ismail_applications_2023, bopp_multiple_2013, gruberWarmingTurningSour2011}, for mechanistic attribution, and to explore consequences of these changes on marine ecosystems, climate feedbacks, and human livelihoods~\citep{park_seasonal_2019, fennel_ocean_2022, moraBioticHumanVulnerability2013}.  For an in-depth overview of BGC modeling, we refer the reader to~\citep{fennel_ocean_2022}.

These same pressures make forecasting increasingly relevant for fisheries, conservation, carbon-cycle tracking, and coastal adaptation~\citep{linkWhyWeNeed2023}, but global-scale forecast products, particularly those targeting policy-relevant, seasonal timescales, remain scarce relative to their counterparts in numerical weather prediction (NWP) and physical oceanography.

Like physical ocean and atmospheric models, BGC models are based on mostly deterministic approximations of nonlinear, interconnected processes. Incorrect parameterization introduces systematic biases that increasingly degrade performance with forecast lead time~\citep{seferianTrackingImprovementSimulated2020, terhaarAssessmentGlobalOceanBiogeochemistry2024,hagstromImpactDynamicPhytoplanktonStoichiometry2024, fuEvaluationOceanBiogeochemistryCMIP2022}. They are also computationally expensive, making it difficult to generate ensembles, produce long-term forecasts, or explore uncertainty at scale. Furthermore, BGC cycles are sensitive to a wide range of climate and physical ocean drivers, which remain imperfectly represented in process-based models~\citep{payne_uncertainties_projecting_2016, frolicher_sources_uncertainties_2016}.

The trajectory of NWP offers a compelling, if imperfect, analogy for the future of BGC forecasting. For decades, weather prediction relied exclusively on physics-based models that, while powerful, shared many of the same limitations. Data-driven alternatives (\citep{pathak_fourcastnet_2022, lam_graphcast_2023, bi_accurate_2023} to cite a few), have since demonstrated that neural networks trained on reanalyses can match or exceed state-of-the-art NWP skill at a fraction of the cost. Nonetheless, marine BGC presents a steeper challenge: it involves many weakly constrained processes that are sensitive to a wide range of physical, chemical, and biological drivers, and its observing system remains far sparser than those available for physical variables.  Yet this is precisely why it could represent an exciting new frontier for the data-driven modeling revolution.

The field of deep learning has a tendency to favor the maximization of architectural expressivity over systems encoding domain-specific prior knowledge,\footnote{As famously articulated in Sutton's ``bitter lesson''~\citep{suttonBitterLesson2019}.} and in weather forecasting this has proven a successful approach. However, given the unique constraints discussed above, the added value of additional parameters and capacity drops quickly. We argue that in this data-scarce, high complexity setting, explicit structural assumptions could prove effective, at least for the establishment of initial baselines.

Concretely, we propose BG4Sea: an architecture that forecasts the marine BGC state through dedicated components for the depthwise, temporal, and spatial structure of the ocean. A column autoencoder compresses the vertical structure of the water column into a low-dimensional latent space (achieving over 23-fold compression), aligned with the depth-resolved nature of in-situ BGC observations~\citep{claustre2020}. A latent forecaster then propagates this representation forward in time. Context enters beyond the local column through surface physical forcing, which has the additional benefit of a comparatively richer observational record, and mirrors how process-based BGC models receive physical boundary conditions~\citep{fennel_ocean_2022}. Horizontal coupling, via cross-attention over neighbouring columns, further restores the transport pathways that 1D models neglect.

This is, to our knowledge, the first global data-driven system that produces multivariate, seasonal BGC forecasts. The resulting system produces skillful 1-6 month forecasts, and our approach demonstrates that local structure carries substantial seasonal information. The system lets us localise where predictability is gained and where it saturates, providing a concrete starting point for incorporating additional observational and dynamical constraints. We position BG4Sea as a structured, interpretable architecture, which can serve as a baseline for future, more expressive models.

\section{Methods}
\label{sec:methods}

\subsection*{Data}

Given the physical constraints that drive biogeochemistry, we include in our study both BGC and physical variables to train and evaluate our model. We use 29 BGC variables (24 prognostic and 5 diagnostic variables) from the BIORYS4 global BGC reanalysis~\citep{gregoire_review_2026} and temperature and salinity from the GLORYS12 global physical reanalysis~\citep{lellouche_copernicus_2021, cmems_phys}. BIORYS4 is based on the PISCES model~\citep{aumont_pisces-v2_2015}, which simulates BGC and ecosystem dynamics of the low trophic levels. The model is further constrained by the assimilation of surface chlorophyll, dissolved inorganic carbon, alkalinity and pH. GLORYS12 is based on the NEMO dynamical model and it assimilates sea surface temperature, sea level anomaly, and direct measurements of temperature and salinity in the water column~\citep{lellouche_copernicus_2021}. We extract 24 vertical levels (down to 220 m depth) from the original 75-level vertical structure. The full list of variables and depths is given in Table~\ref{tab:all_vars}. While some of these tracers might be considered redundant, we let the network see the full state of BIORYS4 rather than a preselected subset to avoid baking in any assumptions and to fully mirror the parent model.

\begin{table}[h]
  \centering
  \caption{Depth-column variables (31), grouped by process. An asterisk ($*$) marks variables used in scoring.}
  \label{tab:all_vars}
  \begin{tabular}{p{0.24\textwidth} p{0.24\textwidth} p{0.24\textwidth} p{0.22\textwidth}}
    \topline
    \textbf{Nutrients and oxygen}           & \textbf{Biology}                             & \textbf{Carbon pools}                               & \textbf{Physics \& light}           \\
    \midline
    Nitrate\textsuperscript{*}              & Nanophytoplankton biomass\textsuperscript{*} & Alkalinity\textsuperscript{*}                       & Temperature                         \\
    Phosphate\textsuperscript{*}            & Diatom biomass\textsuperscript{*}            & Dissolved inorganic carbon\textsuperscript{*}       & Salinity                            \\
    Ammonium                                & Total phytoplankton biomass                  & Dissolved organic carbon\textsuperscript{*}         & Photosynthetically active radiation \\
    Total dissolved iron\textsuperscript{*} & Total chlorophyll\textsuperscript{*}         & Calcium carbonate (calcite)\textsuperscript{*}      &                                     \\
    Diatom iron content                     & Nanophytoplankton chlorophyll                & pH                                                  &                                     \\
    Nanophytoplankton iron content          & Diatom chlorophyll                           & Small particulate organic carbon\textsuperscript{*} &                                     \\
    Large particulate iron                  & Net primary production\textsuperscript{*}    & Large particulate organic carbon                    &                                     \\
    Small particulate iron                  & Microzooplankton biomass\textsuperscript{*}  &                                                     &                                     \\
    Silicate\textsuperscript{*}             & Mesozooplankton biomass\textsuperscript{*}   &                                                     &                                     \\
    Diatom silicon content                  &                                              &                                                     &                                     \\
    Sinking biogenic silica                 &                                              &                                                     &                                     \\
    Oxygen\textsuperscript{*}               &                                              &                                                     &                                     \\
    \botline
  \end{tabular}
\end{table}

Table~\ref{tab:surface_cond} lists the optional surface conditioning fields. We use a monthly temporal resolution, and a global grid with $1/4^\circ$ horizontal resolution. We split the data into training (2010--2017), validation (2018--2019), and test (2020--2022) periods. To direct model capacity at residual variance rather than the mean seasonal cycle, we train on monthly anomalies relative to a monthly climatology. Anomalies are subsequently standardised to zero mean and unit variance using statistics computed over the same period (details in Appendix, see Eqs.~\ref{eq:clim}--\ref{eq:anomaly}).

\begin{table}[h]
  \centering
  \begin{tabular}{rrrrrr}
    \topline
    \textbf{Nominal depth levels (24) on the model vertical grid}        \\
    \midline
    0.5$^{*}$  & 1.5         & 3.8   & 5.1$^{*}$   & 7.9   & 11.4        \\
    13.5$^{*}$ & 15.8        & 18.5  & 21.6$^{*}$  & 25.2  & 29.4        \\
    34.4$^{*}$ & 40.3        & 47.4  & 55.8$^{*}$  & 65.8  & 77.9        \\
    92.3$^{*}$ & 109.7$^{*}$ & 130.7 & 155.9$^{*}$ & 186.1 & 222.5$^{*}$ \\
    \botline
  \end{tabular}
  \caption{Nominal depth levels (24) on the model vertical grid. An asterisk ($*$) marks levels used for scoring.}\label{tab:depths}
\end{table}

\begin{table}[h]
  \centering
  \begin{tabular}{l}
    \topline
    \textbf{Surface-only conditioning} \\
    \midline
    Mixed-layer depth                  \\
    Sea-surface height                 \\
    Sea-surface salinity               \\
    Sea-surface temperature            \\
    Particulate dust                   \\
    \botline
  \end{tabular}
  \caption{Optional surface-only physics forcing.}\label{tab:surface_cond}
\end{table}

The model is trained on samples anchored at individual depth profiles from latitude-longitude grid points subsampled from the global ocean, with the final model additionally conditioned on the surrounding $3\times3$ horizontal neighbourhood (Section~\ref{sec:horizontal_coupling}). Subsampling is done in order to balance global coverage with computational constraints. We use equal-area sampling, which is achieved by binning the data according to the sine of the latitude in order to ensure a uniform distribution of samples across the globe. Batches were formed by grouping samples from the same time step to maximize data efficiency.

\subsection*{Architecture}

\begin{figure}[ht]
  \centering
  \includegraphics[width=\textwidth]{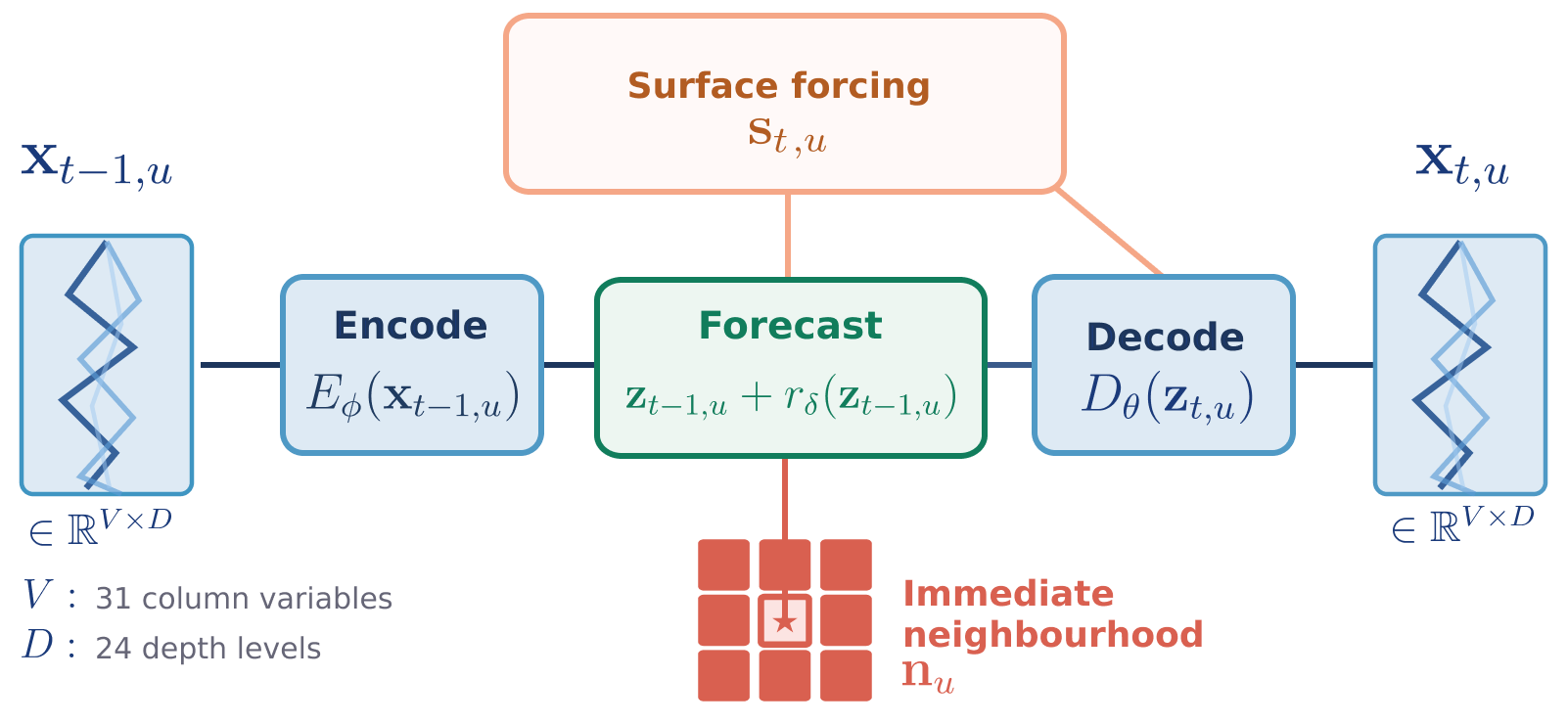}
  \caption{Modular architecture composed of a column autoencoder (shown in blue), a latent forecaster (in green), and optional surface forcings (orange) and horizontal coupling (red).}
  \label{fig:architecture}
\end{figure}

Our final forecasting system compresses each water column into a latent code, propagates that code forward in time, and conditions this propagation on surface physical forcing and on the horizontal neighbourhood (Figure~\ref{fig:architecture}). We arrive at it in stages, introducing one source of information at a time and keeping representation and dynamics separate throughout: an autoencoder defines a latent space, and a latent forecaster is trained within it.

\subsubsection*{Compression}
\label{sec:compressing_the_water_column}

We first trained a deterministic, 1D convolutional autoencoder $(E_\phi, D_\theta)$ to compress the water column. Each depth profile  $\mathbf{x}_{t, u} \in \mathbb{R}^{V \times D}$ (where $V = 31$ variables across $D = 24$ depth levels at a specific time $t$ and horizontal location $u = (i, j)$), is mapped to a low-dimensional latent code $\mathbf{z}_{t, u} \in \mathbb{R}^{L}$ (where $L$ corresponds to the dimensionality of the latent space) and reconstructed back by the decoder:

\begin{equation}
  \mathbf{z}_{t, u} = E_\phi(\mathbf{x}_{t, u}) \in \mathbb{R}^L, \qquad \hat{\mathbf{x}}_{t, u} = D_\theta(\mathbf{z}_{t, u}) \in \mathbb{R}^{V \times D}
  \label{eq:ae_def}
\end{equation}
This corresponds to a compression ratio of $V \times D / L$.

Each decoder stage is conditioned on a spatiotemporal embedding $\mathbf{e}_{D}$, via Feature-wise Linear Modulation (FiLM)~\citep{perezFiLMVisualReasoning2017}: $\mathbf{e}_{D}$ is linearly projected to a scale $\boldsymbol{\gamma}$ and a shift $\boldsymbol{\beta}$ parameters, which modulate each convolutional feature map $\mathbf{h}$:
\begin{equation}
  \mathrm{FiLM}(\mathbf{h},\, \mathbf{e}_{D}) = \boldsymbol{\gamma}(\mathbf{e}_{D}) \odot \mathbf{h} + \boldsymbol{\beta}(\mathbf{e}_{D}),
  \label{eq:film}
\end{equation}
where $\odot$ denotes element-wise multiplication. FiLM was chosen because it can convert heterogeneous context directly into scale and shift "knobs" that modulate the network's internal representation, instead of requiring the network to implicitly learn how to combine unrelated types of information like concatenation would. Implementation details can be found in Appendix~\ref{sec:appendix_ae}.

The training objective was set to the mean squared reconstruction error (MSE), averaged over the $N_{\text{train}}$ training samples (i.e. number of sampled locations per time step $\times$ the number of training months):
\begin{equation}
  \mathcal{L} = \frac{1}{N_{\text{train}}} \sum_{t, u} \left\| \mathbf{x}_{t, u} - \hat{\mathbf{x}}_{t, u} \right\|_F^2
  \label{eq:ae_loss}
\end{equation}
Equal-area sampling was used to yield $N \approx 780$ ocean locations per time step. The data split is summarised in Table~\ref{tab:data_split}.

\subsubsection*{Latent forecasting}

Having established the latent representation of the column structure, we turned to one-step forecasting in latent space $\mathbb{R}^L$. The latent dimensionality was set to $L=32$ as a compromise between representational capacity and compression. This corresponds to the number of components required to explain 87.51\% of the anomaly variance in the training data according to Principal Components Analysis run on the same anomaly fields. Forecasting is achieved through a learned residual $r_\delta$, so that the prediction is given by:

\begin{equation}
  \hat{\mathbf{z}}_{t,u} = \mathbf{z}_{t-1,u} + r_\delta(\mathbf{z}_{t-1,u})
  \label{eq:res_for}
\end{equation}

The column-only variant forecasts within the base autoencoder described above, by obtaining $\mathbf{z}_{t-1,u} = E_\phi(\mathbf{x}_{t-1,u})$ and subsequently decoding $\hat{\mathbf{x}}_{t,u} = D_\theta(\hat{\mathbf{z}}_{t,u})$. The training objective is the mean squared error between the reconstructed prediction and the target anomaly field, as given by Eq.~\ref{eq:ae_loss}. Only the forecaster weights are optimised during training. Each sample consists of a pair of consecutive time steps, and to accommodate the increased memory footprint, we used a sparser spatial sampling strategy ($N \approx 500$) compared to the autoencoder. The data split is summarised in Table~\ref{tab:data_split}, and additional architecture and training details can be found in Appendix~\ref{sec:appendix_forecaster}.

\begin{table}[h]
  \centering
  \begin{tabular}{llccc}
    \topline
    \textbf{Split}        & \textbf{Period} & \textbf{Months}    & \textbf{Autoencoder} & \textbf{Forecaster} \\
    \midline
    Training              & 2010--2017      & All                & $N \approx 780$      & $N \approx 500$     \\
    Validation            & 2018--2019      & All                & $N \approx 780$      & $N \approx 500$     \\
    \midline
    Test (reconstruction) & 2020            & Mar, Jun, Sep, Dec & Full grid            & Full grid           \\
    Test (forecasting)    & 2020--2022      & Mar, Jun, Sep, Dec & Full grid            & Full grid           \\
    \botline
  \end{tabular}
  \caption{Training, validation, and test splits for the column autoencoder and latent forecaster. $N$ indicates the number of subsampled points.}\label{tab:data_split}
\end{table}

\subsubsection*{Surface-level conditioning}
\label{sec:surface_conditioning}

The surface-level variables listed in Table~\ref{tab:surface_cond} capture physical forcing and state information that is absent from, or only partially represented in, the depth-column inputs. We introduce the conditioning at two stages:

\begin{itemize}
  \item \textbf{Surface-informed decoding:} We learn a new autoencoder conditioned on surface fields: instead of the positional embedding $\mathbf{e}_{D}$ alone, we use a learned surface conditioner $\mathbf{s} = f_{\omega}(\mathbf{b},\, \mathbf{e}_{D})$, where $\mathbf{b}$ denotes the surface variables in Table~\ref{tab:surface_cond}. As in Section~\ref{sec:compressing_the_water_column}, each decoder stage is modulated via FiLM (Eq.~\ref{eq:film}), but now $\mathbf{s}$ replaces $\mathbf{e}_{D}$ as the conditioning input thats modulates the decoding of the forecasted latent back into physical space.
        
  \item \textbf{Surface-informed latent dynamics:} A forecaster is trained on the conditioned autoencoder described above, with the learned embedding $\mathbf{s}$ additionally concatenated with the latent state such that $ \hat{\mathbf{z}}_{t,u} = \mathbf{z}_{t-1,u} + r_{\delta s}(\mathbf{z}_{t-1,u};\, \mathbf{s}_{t,u})$
        
\end{itemize}

Similarly to the previous section, only the forecaster weights are optimised during training. The training objective is the same MSE as Eq.~\ref{eq:ae_loss}, applied to the decoded prediction.

\paragraph*{Forecasting the conditioning:} At the target time $t$, the surface variables $\mathbf{b}$ are not necessarily known in advance, and BGC models such as BIORYS4 do not simulate ocean circulation itself, but receive physical boundary conditions from a physical ocean model at each timestep. We thus consider two ways of supplying $\mathbf{b}_t$: a prescribed forcing taken from an external product (in our case the GLORYS12 reanalysis; operationally, a forecast product such as SEAS5~\citep{johnson_seas5_2019}), or a self-generated approximation $\hat{\mathbf{b}}_t$ produced by the model itself, removing the dependency on an external forecast altogether. For the latter, we introduce a lightweight forecaster that advances $\mathbf{b}$ forward in time before calculating the conditioning embedding:

\begin{equation}
  \hat{\mathbf{b}}_{t,u} = \mathbf{b}_{t-1,u} + r_{\delta b}(\mathbf{b}_{t-1,u}),
  \label{eq:surf_prop}
\end{equation}
where $r_{\delta b}$ is a learned residual that is independently optimised using gradients only from its own surface loss (which is also defined as an MSE). Architecture and training details are given in Appendix~\ref{sec:appendix_surface}.

\subsubsection*{Horizontal coupling}
\label{sec:horizontal_coupling}

So far, we have ignored horizontal transport. But the local regime has a significant impact on BGC variability. We address this by adding horizontal context vector $\mathbf{n}_u$ to the latent forecaster:

\begin{equation}
  \hat{\mathbf{z}}_{t,u} = \mathbf{z}_{t-1,u} + r_{\delta sn}(\mathbf{z}_{t-1,u};\, \mathbf{s}_{t,u};\, \mathbf{n}_u)
  \label{eq:horizontal_coupling},
\end{equation}
where $\mathbf{n}_u$ is comprised of a column and surface components $\mathbf{n}^z_{t-1,u}$ and $\mathbf{n}^s_{t,u}$, respectively. These are computed via cross-attention over the latent representations of the neighbouring features, calculated at $t-1$ for the latent columns and target time $t$ for the surface. The smallest possible neighbourhood ($N=8$ neighbours in a $3\times 3$ grid) is chosen in order to keep the computational footprint low. In our $1/4^\circ$, regular grid, this spans about 80 km $\times$ 80 km at the Equator. This variant reuses the surface-informed autoencoder from the previous section. Further architecture and training details can be found in Appendix~\ref{sec:appendix_horizontal}.

\subsection*{Evaluation}

\paragraph*{Compression}

Reconstruction quality is evaluated on the test set (see Table~\ref{tab:data_split}) as a function of the latent dimension $L \in \{8, 16, 32\}$, using the coefficient of determination $R^2$ (Eq.~\ref{eq:r2}) reported in both the anomaly and physical space and computed over all variables and depth levels (Tables~\ref{tab:all_vars}--\ref{tab:depths}).

We compare against PCA applied to the flattened column vectors ($C \times D$) and truncated to the same number of components, fitted on the same anomaly training set. We additionally report root mean squared error (RMSE) and normalised root mean squared error (NRMSE; Eqs.~\ref{eq:rmse}--\ref{eq:nrmse}).

\paragraph*{Forecasting}

Forecast skill for the selected latent dimension $L=32$ is assessed on the entire horizontal grid for the variable-depth pairs selected for evaluation (marked with an asterisk in Tables~\ref{tab:all_vars} and~\ref{tab:depths}). Each model variant is evaluated autoregressively for lead times one, three and six months, meaning that for lead times $LT> 1$, the last predicted latent state is passed as input to the next prediction. Decoded predictions are transformed back to physical space by reversing the standardisation and adding back the climatology. We evaluate each variant using four months per year: March, June, September and December, for 2020--2022 (Table~\ref{tab:data_split}). We consider this a reasonable compromise between inference costs and inter-annual coverage.

Predictions are evaluated with respect the reference model (BIORYS4). Our primary evaluation metric is the anomaly correlation coefficient (ACC; Eq.~\ref{eq:acc}), and we additionally report $R^2$ (in the physical space), RMSE, and NRMSE (Eqs.~\ref{eq:r2}--\ref{eq:nrmse}). Metrics are calculated by pooling across all points in the grid, aggregated over the spatial domain using area weights to avoid representation biases associated with the non-uniform cell area of the regular grid (details in Appendix~\ref{sec:supp_preprocessing_metrics}). Summary scores are then obtained by averaging within three main groups: nutrients and oxygen, biology, and carbon pools. We compare each model variant to two baselines: the monthly climatology $\bar{\mathbf{k}}_m$, and initialisation persistence.

We use contemporaneous surface conditioning $\mathbf{s}_{t}$ to demonstrate an upper bound on skill when assessing the contribution of the different modules, analogous to how a process-based model receives physical forcings at each timestep. Finally, we evaluate the surface forecaster autoregressively on the final model, and assess its performance on temperature and salinity predictions.

\section{Results}

\subsection*{Compression}

\begin{figure}[htbp]
  \centering
  \includegraphics[width=\textwidth]{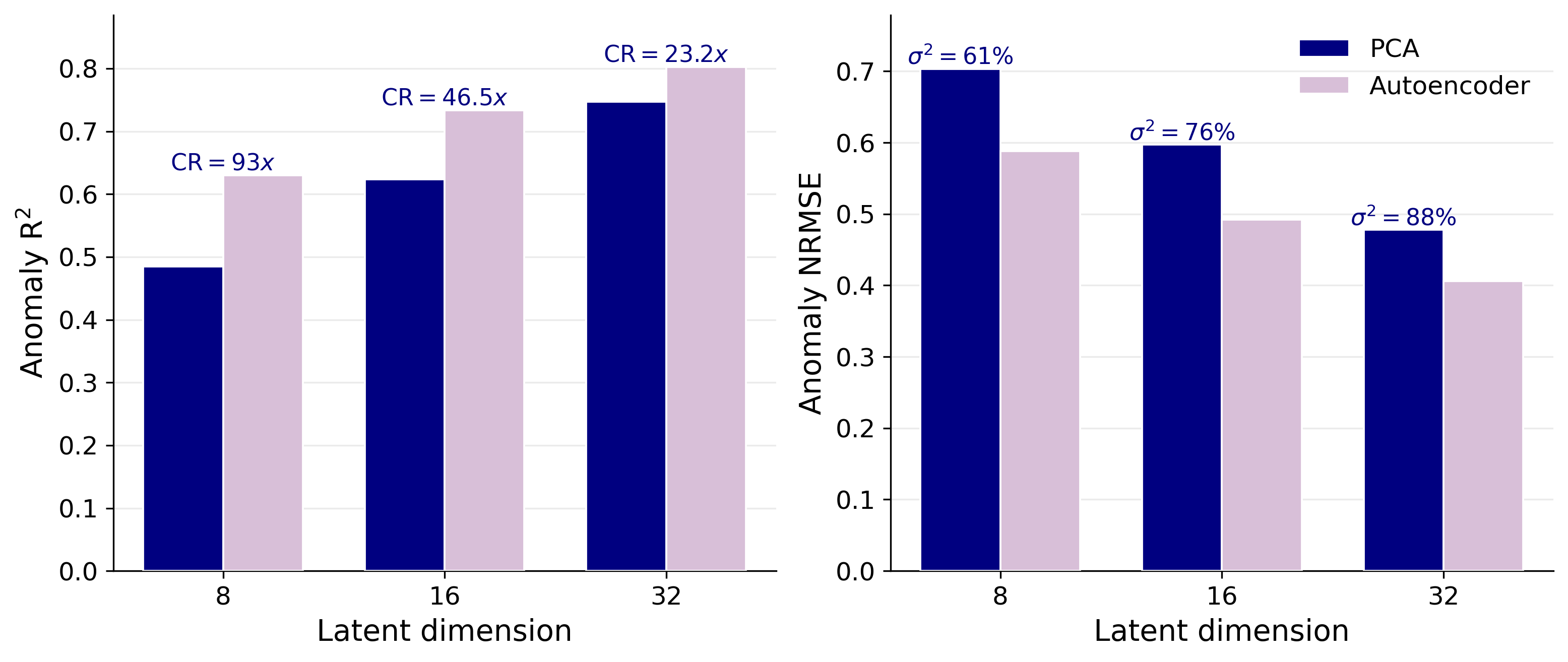}
  \caption{R$^2$ and NRMSE in anomaly space as a function of latent dimension. The autoencoder is shown in light purple, and an equivalent PCA baseline is shown in dark purple. $\sigma^2$ shows the cumulative explained variance for each PCA variant and $CR= \frac{V \times D}{L}$ shows the compression ratio.}
  \label{fig:latent_dim_metrics_2020}
\end{figure}

Figure~\ref{fig:latent_dim_metrics_2020} shows R$^2$ and NRMSE (both in anomaly space) as a function of latent dimension $L$, evaluated on the test set described in Table~\ref{tab:data_split} and averaged across all 31 input variables. We also show an equivalent PCA truncated to the same number of components, and its cumulative explained variance ($\sigma^2$), which was used to determine the latent dimension: $L=32$ explains 87.51\% of the anomaly variance.

Tables~\ref{tab:ae_metrics_chem}--\ref{tab:ae_metrics_phys} show the per-variable reconstruction metrics ($R^2$ in anomaly and physical space, and RMSE in physical units) for the selected autoencoder ($L=32$). We can preliminarily conclude that the autoencoder is able to reconstruct most of the BGC anomalies, with the exception of iron and silica related variables (and to a lesser-extent, particulate organic carbon), which show lower anomaly reconstruction quality in spite of a good physical $R^2$. While a higher $L$ is likely to improve reconstruction quality for most variables (as evidenced by the trend in Figure~\ref{fig:latent_dim_metrics_2020}), these results indicate that iron and silica would likely benefit from being encoded through a different strategy.

\begin{table}[h]
  \centering
  \begin{tabular}{lrrl}
    \topline
    \textbf{Variable}              & \textbf{$R^2$ (anom.)} & \textbf{$R^2$ (phys.)} & \textbf{RMSE (units)}                    \\
    \midline
    Nitrate                        & 0.947                  & 0.999                  & 0.243 (mmol m$^{-3}$)                    \\
    Phosphate                      & 0.944                  & 0.999                  & 0.018 (mmol m$^{-3}$)                    \\
    Ammonium                       & 0.900                  & 0.991                  & 0.031 (mmol m$^{-3}$)                    \\
    Total dissolved iron           & 0.747                  & 0.991                  & $5.2 \times 10^{-5}$ ($\mu$mol m$^{-3}$) \\
    Diatom iron content            & 0.822                  & 0.981                  & $2.1 \times 10^{-6}$ ($\mu$mol m$^{-3}$) \\
    Nanophytoplankton iron content & 0.877                  & 0.986                  & $1.6 \times 10^{-6}$ ($\mu$mol m$^{-3}$) \\
    Large particulate iron         & 0.564                  & 0.940                  & $7.9 \times 10^{-7}$ ($\mu$mol m$^{-3}$) \\
    Small particulate iron         & 0.310                  & 0.903                  & $8.8 \times 10^{-6}$ ($\mu$mol m$^{-3}$) \\
    Silicate                       & 0.264                  & 0.962                  & 3.34 (mmol m$^{-3}$)                     \\
    Diatom silicon content         & 0.831                  & 0.980                  & 0.037 (mmol m$^{-3}$)                    \\
    Sinking biogenic silica        & 0.203                  & 0.874                  & 0.037 (mmol m$^{-3}$)                    \\
    Oxygen                         & 0.908                  & 0.999                  & 2.02 (mmol m$^{-3}$)                     \\
    \botline
  \end{tabular}
  \caption{Per-variable reconstruction metrics (latent dimension 32): nutrients and oxygen. Columns show anomaly $R^2$, physical $R^2$, and RMSE.}\label{tab:ae_metrics_chem}
\end{table}

\begin{table}[h]
  \centering
  \begin{tabular}{lrrl}
    \topline
    \textbf{Variable}             & \textbf{$R^2$ (anom.)} & \textbf{$R^2$ (phys.)} & \textbf{RMSE (units)}         \\
    \midline
    Nanophytoplankton biomass     & 0.904                  & 0.989                  & 0.045 (mmol m$^{-3}$)         \\
    Diatom biomass                & 0.878                  & 0.983                  & 0.083 (mmol m$^{-3}$)         \\
    Total phytoplankton biomass   & 0.904                  & 0.989                  & 0.096 (mmol m$^{-3}$)         \\
    Total chlorophyll             & 0.889                  & 0.985                  & 0.030 (mg m$^{-3}$)           \\
    Nanophytoplankton chlorophyll & 0.909                  & 0.987                  & 0.012 (mg m$^{-3}$)           \\
    Diatom chlorophyll            & 0.840                  & 0.974                  & 0.027 (mg m$^{-3}$)           \\
    Net primary production        & 0.867                  & 0.982                  & 1.01 (mmol m$^{-3}$ d$^{-1}$) \\
    Microzooplankton biomass      & 0.932                  & 0.992                  & 0.034 (mmol m$^{-3}$)         \\
    Mesozooplankton biomass       & 0.935                  & 0.994                  & 0.024 (mmol m$^{-3}$)         \\
    \botline
  \end{tabular}
  \caption{Per-variable reconstruction metrics (latent dimension 32): biology.}\label{tab:ae_metrics_bio}
\end{table}

\begin{table}[h]
  \centering
  \begin{tabular}{lrrl}
    \topline
    \textbf{Variable}                & \textbf{$R^2$ (anom.)} & \textbf{$R^2$ (phys.)} & \textbf{RMSE (units)}                \\
    \midline
    Alkalinity                       & 0.880                  & 0.998                  & 3.18 (mmol m$^{-3}$)                 \\
    Dissolved inorganic carbon       & 0.919                  & 0.998                  & $3.3 \times 10^{-3}$ (mmol m$^{-3}$) \\
    Dissolved organic carbon         & 0.948                  & 0.999                  & $2.6 \times 10^{-4}$ (mmol m$^{-3}$) \\
    Calcium carbonate (calcite)      & 0.935                  & 0.990                  & $2.2 \times 10^{-3}$ (mmol m$^{-3}$) \\
    pH                               & 0.896                  & 0.996                  & $4.6 \times 10^{-3}$ (--)            \\
    Small particulate organic carbon & 0.598                  & 0.946                  & 0.208 (mmol m$^{-3}$)                \\
    Large particulate organic carbon & 0.743                  & 0.961                  & 0.028 (mmol m$^{-3}$)                \\
    \botline
  \end{tabular}
  \caption{Per-variable reconstruction metrics (latent dimension 32): carbon pools.}\label{tab:ae_metrics_carbon}
\end{table}

\begin{table}[h]
  \centering
  \begin{tabular}{lrrl}
    \topline
    \textbf{Variable}                   & \textbf{$R^2$ (anom.)} & \textbf{$R^2$ (phys.)} & \textbf{RMSE (units)} \\
    \midline
    Temperature                         & 0.913                  & 0.999                  & 0.237 ($^\circ$C)     \\
    Salinity                            & 0.819                  & 0.995                  & 0.105 (PSU)           \\
    Photosynthetically active radiation & 0.856                  & 0.990                  & 1.81 (W m$^{-2}$)     \\
    \botline
  \end{tabular}
  \caption{Per-variable reconstruction metrics (latent dimension 32): physics and light.}\label{tab:ae_metrics_phys}
\end{table}

\clearpage

\subsection*{Forecast skill through progressive information scaling}

Throughout the following sections, we evaluate seasonal forecast skill at lead times up to six months for the selected variables and depth levels (Table~\ref{tab:all_vars} and~\ref{tab:depths}) of three main BGC groups. Nutrients and oxygen includes nitrate, phosphate, silicate, dissolved iron, and oxygen (NO3, PO4, Si, Fe, O2). Biology includes chlorophyll-a, net primary production, and diatom, nanophytoplankton, microzooplankton, and mesozooplankton biomass (chl-a, NPP, diatom, nanopkt, microzoo, mesozoo). Carbon pools include alkalinity, dissolved organic and inorganic carbon, calcite, and small particulate organic carbon (alk, DOC, DIC, CaCO$_3$, POC).

Our final forecasting system combines column compression, surface conditioning, and horizontal coupling (Figure~\ref{fig:architecture}); here we quantify what each information source contributes. The column-only model propagates its latent state forward in time using no external information, measuring how much of the future BGC state is recoverable from the vertical column alone. The surface-forced variant additionally integrates the surface physical state through conditioning, and the horizontally-coupled variant adds local regime variability via cross-attention over the immediate neighbourhood states.

\begin{figure}[htbp]
  \centering
  \includegraphics[width=\textwidth]{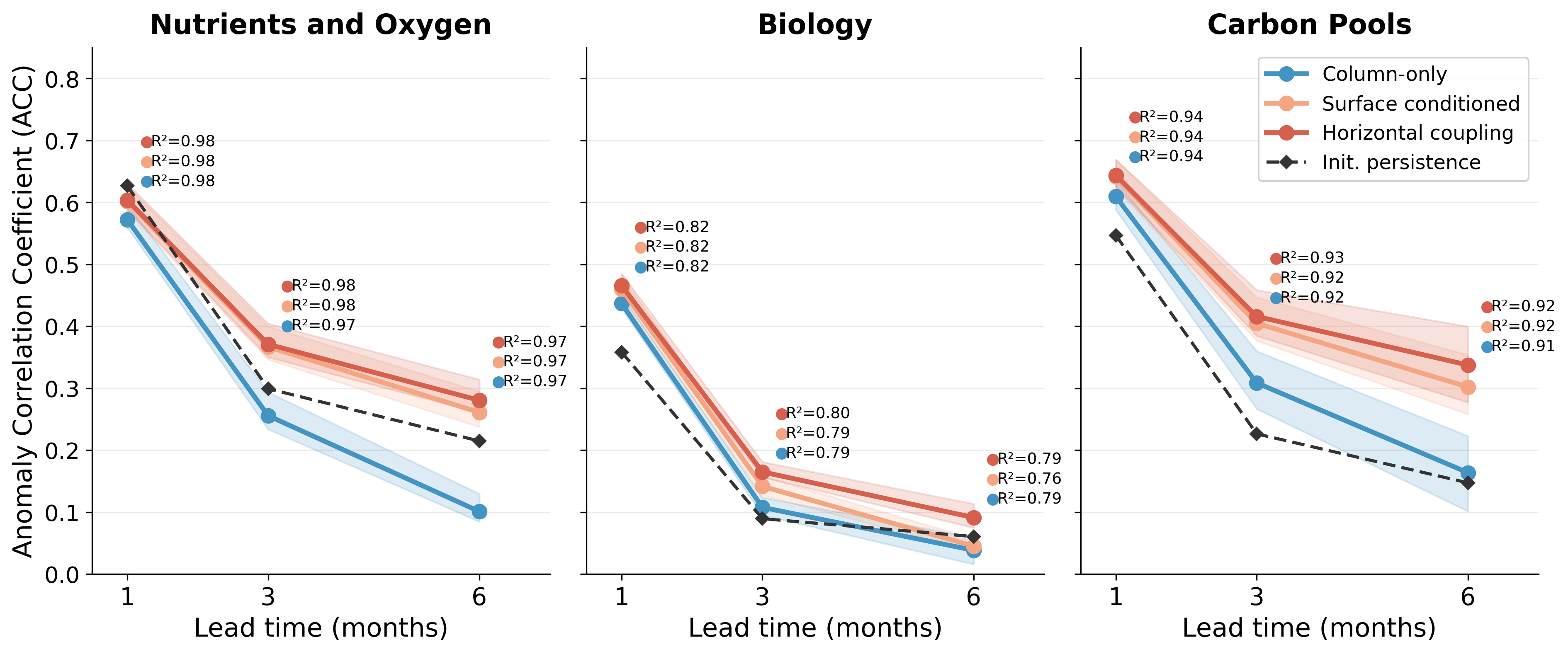}
  \caption{ACC by BGC group (see Table~\ref{tab:all_vars}) for each model variant. The column-only model is shown in blue, the surface-forced variant in orange, and the surface-forced with horizontal coupling model in red. The shading indicates the minimum and maximum scores across the three evaluated years. Persistence (black dashed lines) is shown for reference.}
  \label{fig:forecast_acc_by_group}
\end{figure}

Figure~\ref{fig:forecast_acc_by_group} compares the three proposed variants in terms of the anomaly correlation coefficient (ACC, Eq.~\ref{eq:acc}), with persistence used as a baseline. We also show $R^2$ (in the physical space) for reference. A breakdown of these results is shown in Table~\ref{tab:forecast_groups}, and variable level details are shown in the Appendix (Tables~\ref{tab:forecast_acc_by_var}--\ref{tab:forecast_varratio_by_var})

The column-only variant establishes a floor: how far the vertical state can carry a forecast with no external information. It beats persistence for biology and carbon pools at short lead times, but trails persistence for nutrients and oxygen at all lead times, and for biology by lead time six. Surface physical forcing contributes the largest single increment in skill, consistent with the strong physical control on BGC variability, while horizontal coupling adds a smaller but systematic gain that grows with lead time, becoming most pronounced at lead time six. The final horizontally-coupled model outperforms the other two (and persistence) for all groups and lead times, with the exception of the nutrients and oxygen group, where persistence outperforms all models at lead time one.

Nutrients and oxygen, and the carbon pools show the highest skill at lead time six, with average ACC scores of $\approx 0.3$ and $\approx 0.34$, and $R^2$ of 0.97 and $\approx 0.92$ respectively. The gap between the most basic model (blue) and the surface-forced, horizontally-coupled variant (red) widens with lead time, indicating that the added information improves the dynamical representation of the system.

In contrast, the biology group loses almost all anomaly skill by six months, and additional complexity barely nudges this score. Nonetheless, the most complex variant manages to outperform persistence at all lead times. There is an apparent contradiction between the relatively high $R^2 \approx 0.79$ and the low ACC, but this is due to the strength of the seasonal cycle; the climatology explains most of the variance (Table~\ref{tab:forecast_groups}) and surpassing becomes increasingly difficult with lead time, particularly at a monthly temporal resolution. 

\begin{figure}[htbp]
  \centering
  \includegraphics[width=\textwidth]{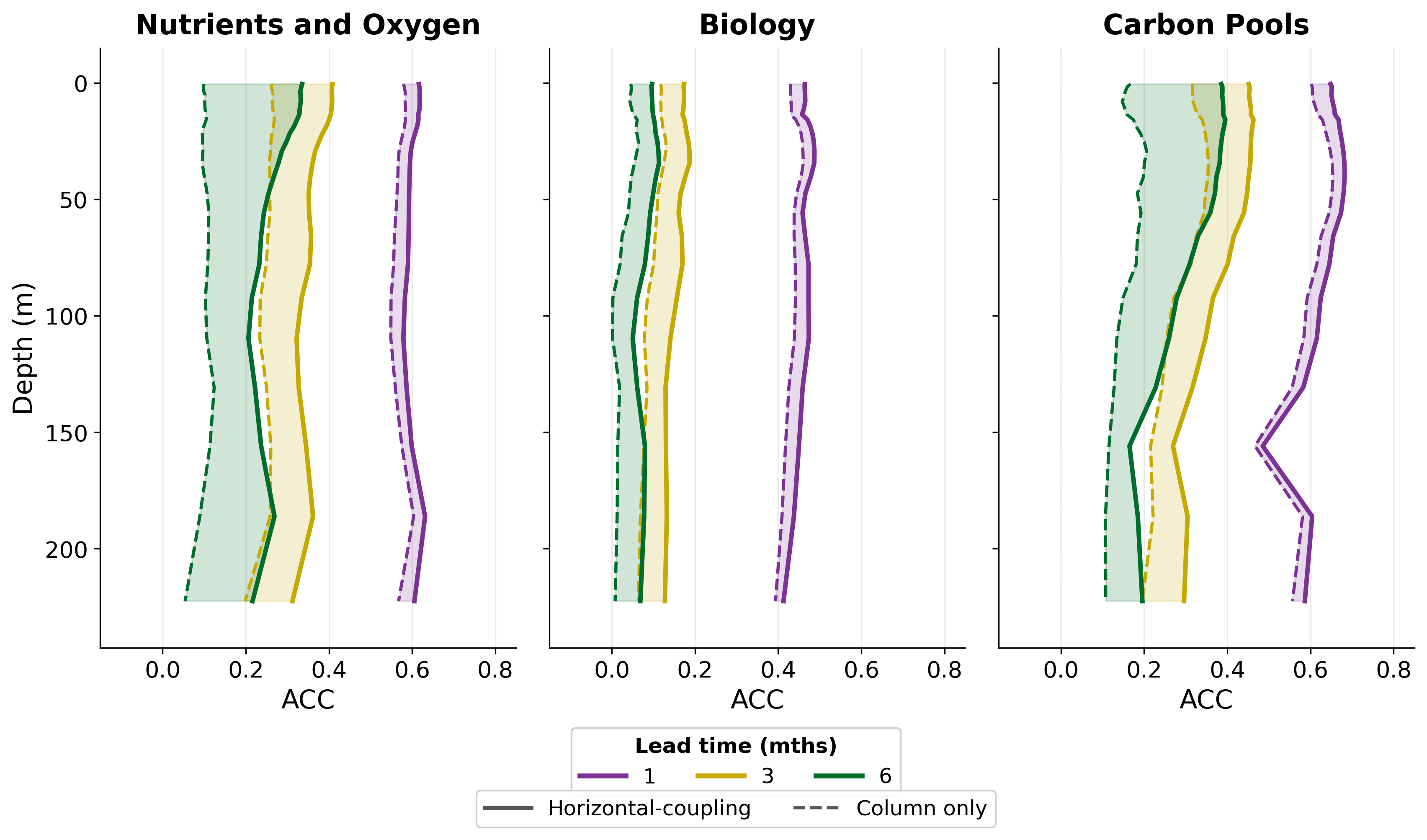}
  \caption{ACC by BGC group and depth level for the column-only (dashed lines) and surface-forced, horizontally-coupled (solid lines) variants by group. Lead times one, three and six are shown in purple, yellow and green, respectively. Shading is used to highlight the gain in ACC relative to the column-only model.}
  \label{fig:forecast_acc_by_depth}
\end{figure}

Figure~\ref{fig:forecast_acc_by_depth} shows how ACC varies with depth across all 24 depth levels, comparing the column-only model with the most complex variant at the three evaluated lead times. The gain from added complexity is most pronounced at longer lead times (shown in green), and for the nutrients and oxygen and carbon pools groups. This is similar to the results shown in Figure~\ref{fig:forecast_acc_by_group}. Improvements are most marked near the surface for those two groups, while they are more uniform for biology. Skill is largely preserved at depth, although it is possible to observe some changes between 150--200\,m.

\begin{table}[h]
  \centering
  \begin{tabular}{lccc}
    \topline
                          & \textbf{Nutrients and oxygen}                                        & \textbf{Biology}                                                     & \textbf{Carbon pools}                                                \\
    \midline
    \textbf{ACC}          &                                                                      &                                                                      &                                                                      \\
    Persistence           & \textbf{0.627}\,$\bullet$\,0.299\,$\bullet$\,0.215                   & 0.358\,$\bullet$\,0.089\,$\bullet$\,0.061                            & 0.547\,$\bullet$\,0.226\,$\bullet$\,0.147                            \\
    Column-only           & 0.572\,$\bullet$\,0.256\,$\bullet$\,0.101                            & 0.437\,$\bullet$\,0.108\,$\bullet$\,0.038                            & 0.609\,$\bullet$\,0.309\,$\bullet$\,0.164                            \\
    + Surface-forcing     & 0.602\,$\bullet$\,0.367\,$\bullet$\,0.261                            & 0.460\,$\bullet$\,0.142\,$\bullet$\,0.045                            & 0.641\,$\bullet$\,0.405\,$\bullet$\,0.302                            \\
    + Horizontal-coupling & 0.604\,$\bullet$\,\textbf{0.371}\,$\bullet$\,\textbf{0.280}          & \textbf{0.466}\,$\bullet$\,\textbf{0.165}\,$\bullet$\,\textbf{0.091} & \textbf{0.644}\,$\bullet$\,\textbf{0.416}\,$\bullet$\,\textbf{0.337} \\
    \midline
    \textbf{R$^2$}        &                                                                      &                                                                      &                                                                      \\
    Climatology           & 0.973\,$\bullet$\,0.973\,$\bullet$\,0.973                            & \textbf{0.821}\,$\bullet$\,\textbf{0.821}\,$\bullet$\,\textbf{0.821} & 0.918\,$\bullet$\,0.918\,$\bullet$\,0.918                            \\
    Persistence           & \textbf{0.980}\,$\bullet$\,0.932\,$\bullet$\,0.899                   & 0.669\,$\bullet$\,0.089\,$\bullet$\,$-$0.203                         & 0.875\,$\bullet$\,0.637\,$\bullet$\,0.471                            \\
    Column-only           & \textbf{0.980}\,$\bullet$\,0.974\,$\bullet$\,0.973                   & 0.817\,$\bullet$\,0.794\,$\bullet$\,0.788                            & 0.940\,$\bullet$\,0.921\,$\bullet$\,0.915                            \\
    + Surface-forcing     & \textbf{0.980}\,$\bullet$\,0.975\,$\bullet$\,\textbf{0.974}          & 0.816\,$\bullet$\,0.789\,$\bullet$\,0.762                            & 0.942\,$\bullet$\,0.924\,$\bullet$\,0.917                            \\
    + Horizontal-coupling & \textbf{0.980}\,$\bullet$\,\textbf{0.976}\,$\bullet$\,\textbf{0.974} & 0.815\,$\bullet$\,0.796\,$\bullet$\,0.791                            & \textbf{0.943}\,$\bullet$\,\textbf{0.926}\,$\bullet$\,\textbf{0.921} \\
    \midline
    \textbf{NRMSE}        &                                                                      &                                                                      &                                                                      \\
    Climatology           & 0.152\,$\bullet$\,0.152\,$\bullet$\,0.152                            & 0.418\,$\bullet$\,\textbf{0.418}\,$\bullet$\,\textbf{0.418}          & 0.255\,$\bullet$\,0.255\,$\bullet$\,0.255                            \\
    Persistence           & 0.132\,$\bullet$\,0.246\,$\bullet$\,0.298                            & 0.568\,$\bullet$\,0.947\,$\bullet$\,1.089                            & 0.296\,$\bullet$\,0.517\,$\bullet$\,0.625                            \\
    Column-only           & 0.128\,$\bullet$\,0.148\,$\bullet$\,0.153                            & 0.392\,$\bullet$\,0.432\,$\bullet$\,0.440                            & 0.208\,$\bullet$\,0.247\,$\bullet$\,0.257                            \\
    + Surface-forcing     & \textbf{0.125}\,$\bullet$\,\textbf{0.143}\,$\bullet$\,0.148          & 0.388\,$\bullet$\,0.433\,$\bullet$\,0.454                            & 0.203\,$\bullet$\,0.238\,$\bullet$\,0.250                            \\
    + Horizontal-coupling & \textbf{0.125}\,$\bullet$\,\textbf{0.143}\,$\bullet$\,\textbf{0.147} & \textbf{0.387}\,$\bullet$\,0.428\,$\bullet$\,0.435                   & \textbf{0.202}\,$\bullet$\,\textbf{0.236}\,$\bullet$\,\textbf{0.244} \\
    \botline
  \end{tabular}
  \caption{Group-mean metrics at lead times 1, 3, and 6 months (shown as $x$\,$\bullet$\,$x$\,$\bullet$\,$x$). Persistence and climatology are shown for reference. Best values (higher ACC and $R^2$; lower NRMSE) at each lead time are shown in bold.}\label{tab:forecast_groups}
\end{table}

\clearpage

\subsection*{Evaluating spatiotemporal skill}

\begin{figure}[htbp]
  \centering
  \includegraphics[width=\textwidth]{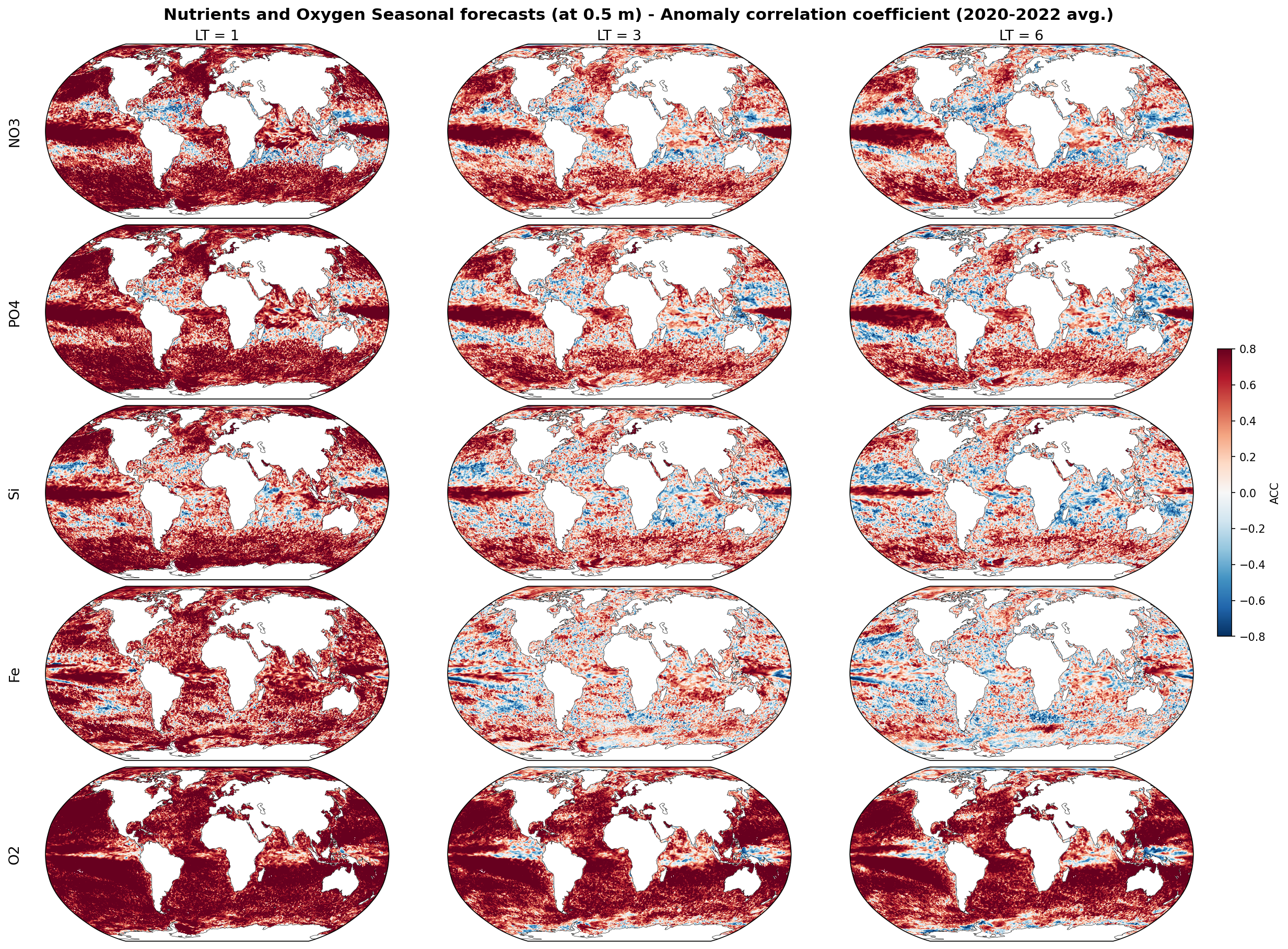}
  \caption{Surface anomaly correlation coefficient (ACC) for nitrate (NO$_3$), phosphate (PO$_4$), silicate (Si), dissolved iron (Fe), and oxygen (O$_2$) at lead times 1, 3, 6 months.}
  \label{fig:acc_chemsurf}
\end{figure}

\begin{figure}[htbp]
  \centering
  \includegraphics[width=\textwidth]{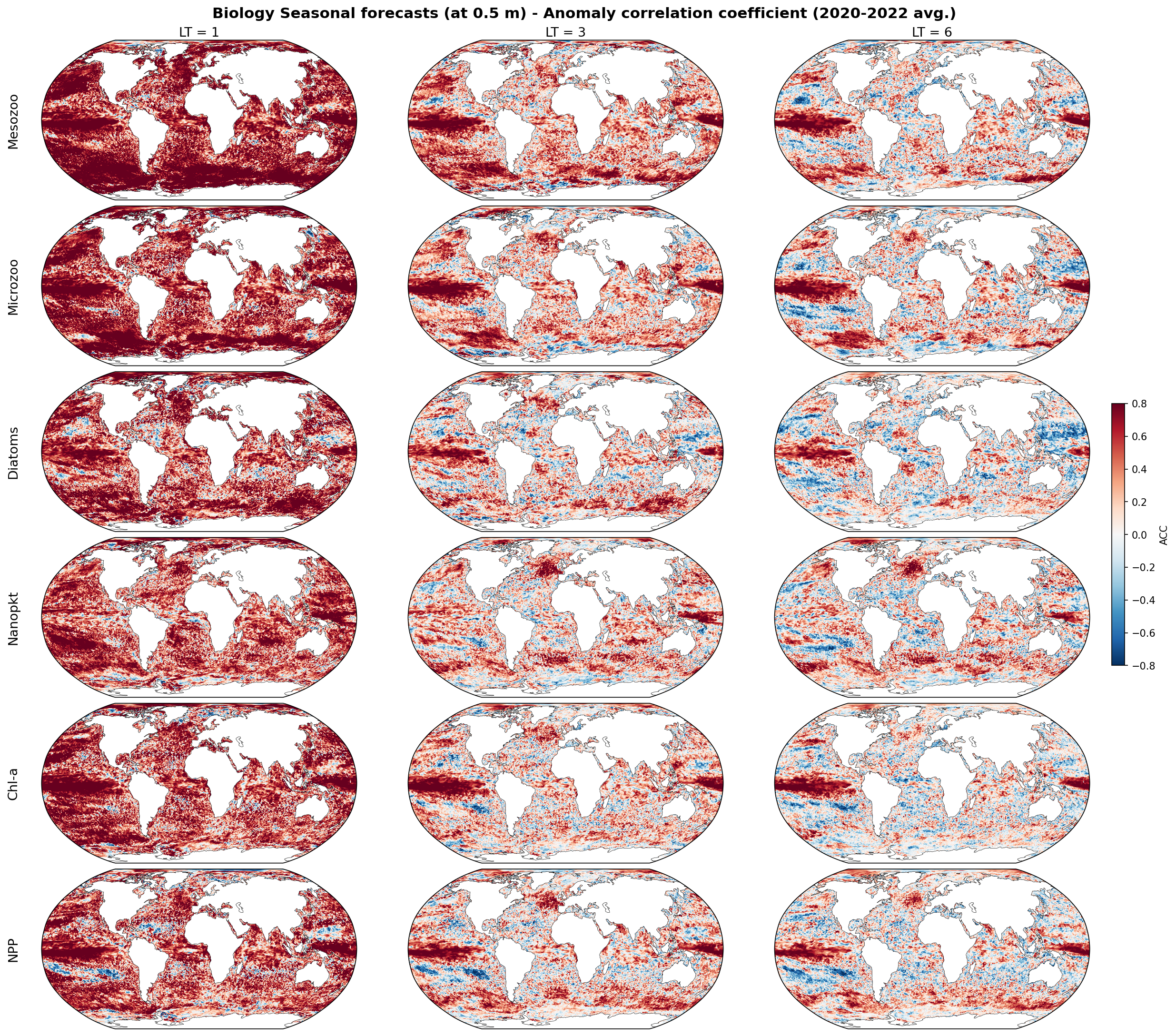}
  \caption{Surface ACC for chlorophyll-a (chl-a), net primary production (NPP), nanophytoplankton (nanopkt), microzooplankton (microzoo), and mesozooplankton (mesozoo) at lead times 1, 3, 6 months.}
  \label{fig:acc_biologysurf}
\end{figure}

\begin{figure}[htbp]
  \centering
  \includegraphics[width=\textwidth]{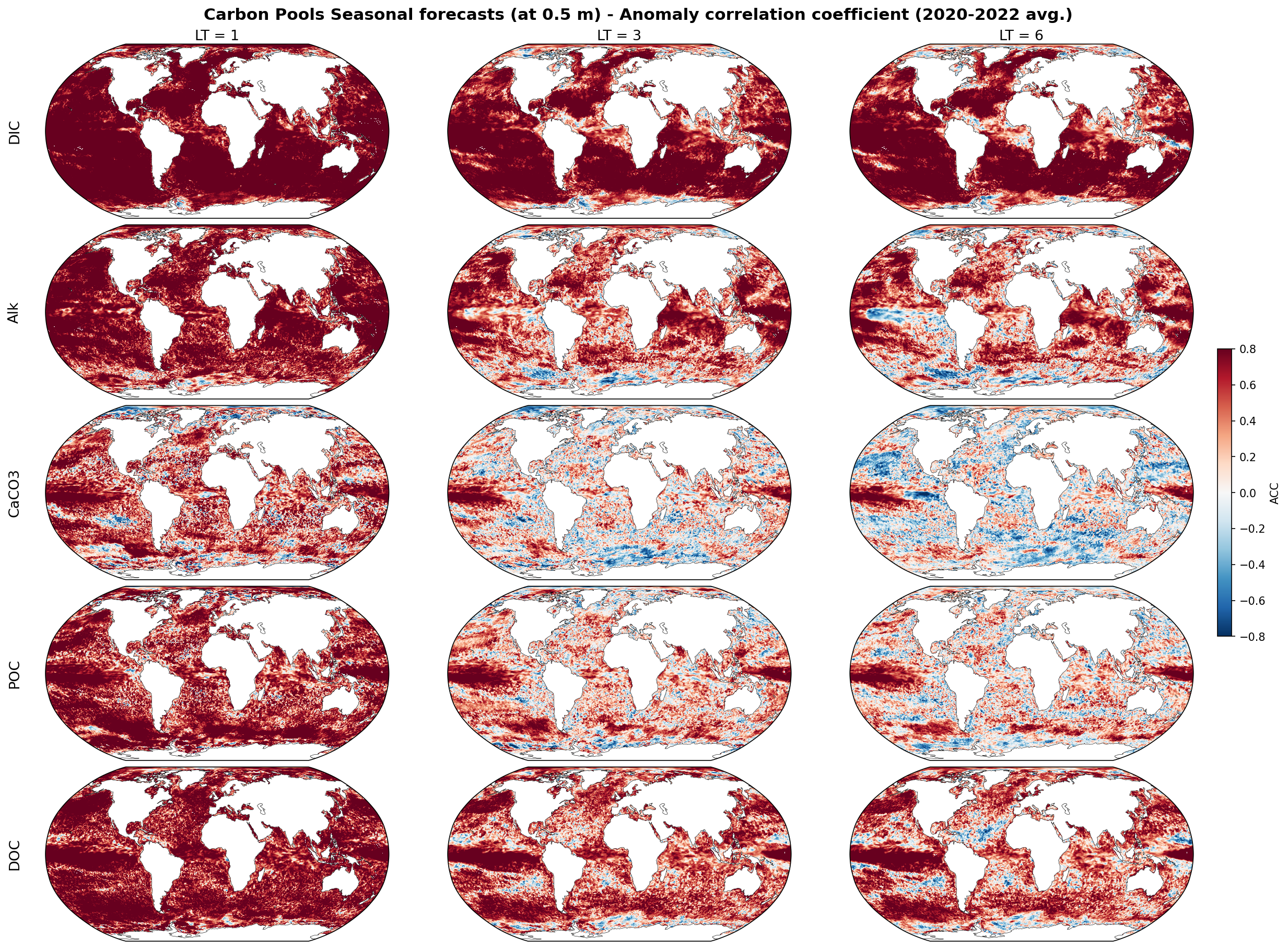}
  \caption{Surface ACC for alkalinity (alk), dissolved organic carbon (DOC), dissolved inorganic carbon (DIC), calcite or calcium carbonate (CaCO$_3$), and small particulate organic carbon (POC) at lead times 1, 3, 6 months.}
  \label{fig:acc_carbonsurf}
\end{figure}

The remainder of the results section focuses on the surface-forced, horizontally-coupled model. Figures~\ref{fig:acc_chemsurf},~\ref{fig:acc_biologysurf}, and~\ref{fig:acc_carbonsurf} show per-pixel ACC at the surface for the nutrients and oxygen, biology, and carbon variables respectively when forecasted at lead times one, three and six. These were calculated by averaging temporal ACC over the test period (Table~\ref{tab:data_split}) at each spatial grid point. Equivalent maps at 220\,m depth are shown in the Appendix (Figs.~\ref{fig:acc_chem220},~\ref{fig:acc_biology220}, and~\ref{fig:acc_carbon220}).

The ACC figures show that at lead time one, most of the global ocean displays high positive ACC (deep reds, > 0.6-0.8), indicating strong short-range predictability. By lead time six, however, skill becomes more heterogenous, with the weakest performance in the subtropical gyres and mid-latitudes.

\paragraph*{Nutrients and oxygen} At lead time one (Figure~\ref{fig:acc_chemsurf}), NO$_3$, PO$_4$, and Si show strong positive ACC in the equatorial bands, poles, higher latitudes, and the Southern Ocean, while the Tropical Atlantic and the gyres are the least well represented regions. Skill degrades significantly by lead time six, but some performance is retained in the equatorial bands and the Southern Ocean. Iron shows more heterogeneous performance from lead time one, and degrades faster than the other nutrients. Oxygen is the most persistently skillful variable, retaining high ACC across most of the global ocean even at higher lead times, although some oxygen minimum zones (Eastern Tropical Pacific, Indian Ocean and the Philippine Sea) show degradation. Performance is relatively consistent at 220\,m depth (see Appendix, Fig.~\ref{fig:acc_chem220}), with the exception of oxygen, which loses most of its skill by lead time six at this depth (although the Eastern Tropical Pacific is better represented).

\paragraph*{Biology} These variables show the fastest degradation in skill (Figure~\ref{fig:acc_biologysurf}), which is expected given that the shorter time scales of plankton dynamics are likely to be poorly captured at a monthly resolution. Most variables retain some skill in the Equatorial Pacific even at lead time six, with the exception of nanophytoplankton. Some patches of moderate skill are retained at southern latitudes for the plankton groups and NPP. The zooplankton groups show better performance compared to the other variables, likely due to their stronger persistence (see Table~\ref{tab:forecast_acc_by_var}). At 220\,m depth (see Appendix, Fig.~\ref{fig:acc_biology220}), most variables show deteriorated performance. Zooplankton loses its edge and NPP is deteriorated from lead time one (although it is effectively absent below the euphotic zone, leaving little anomaly variance on which to base a meaninful ACC).

\paragraph*{Carbon pools} Alkalinity, DIC and DOC show nearly saturated ACC (0.8+) at lead time one, and retain high skill even at six months lead time, with DIC demonstrating one of the longest memory timescales. CaCO$_3$ and (small) POC show more heterogeneous performance, with CaCO$_3$ showing almost no skill by lead time six. POC is slightly better, but its degradation is consistent with the shorter memory timescales of biologically linked processes. At 220\,m (see Appendix, Fig.~\ref{fig:acc_carbon220}), DIC loses some skill, but remains relatively well represented. The rest of the variables deteriorate, with the exception of CaCO$_3$, which shows slight improvements over its performance at the surface.

\begin{figure}[htbp]
  \centering
  \includegraphics[width=\textwidth]{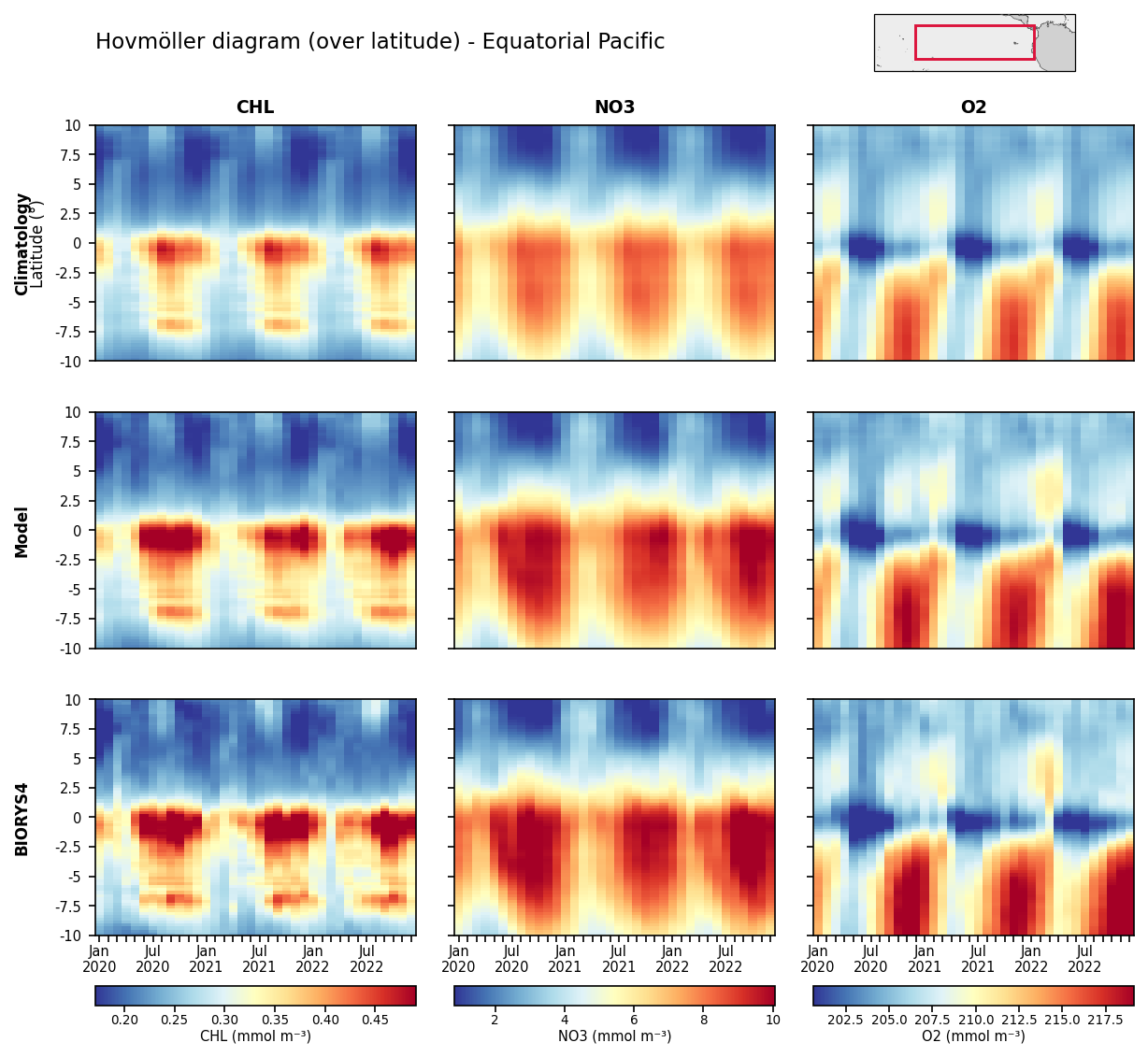}
  \caption{Hovmöller diagram of the zonal mean at the surface of chlorophyll-a, nitrate and oxygen in the Equatorial Pacific at lead time one. The region is shown at the top of the figure, highlighted with a red box.}
  \label{fig:hovmoller_eqpac_lat}
\end{figure}

\begin{figure}[htbp]
  \centering
  \includegraphics[width=\textwidth]{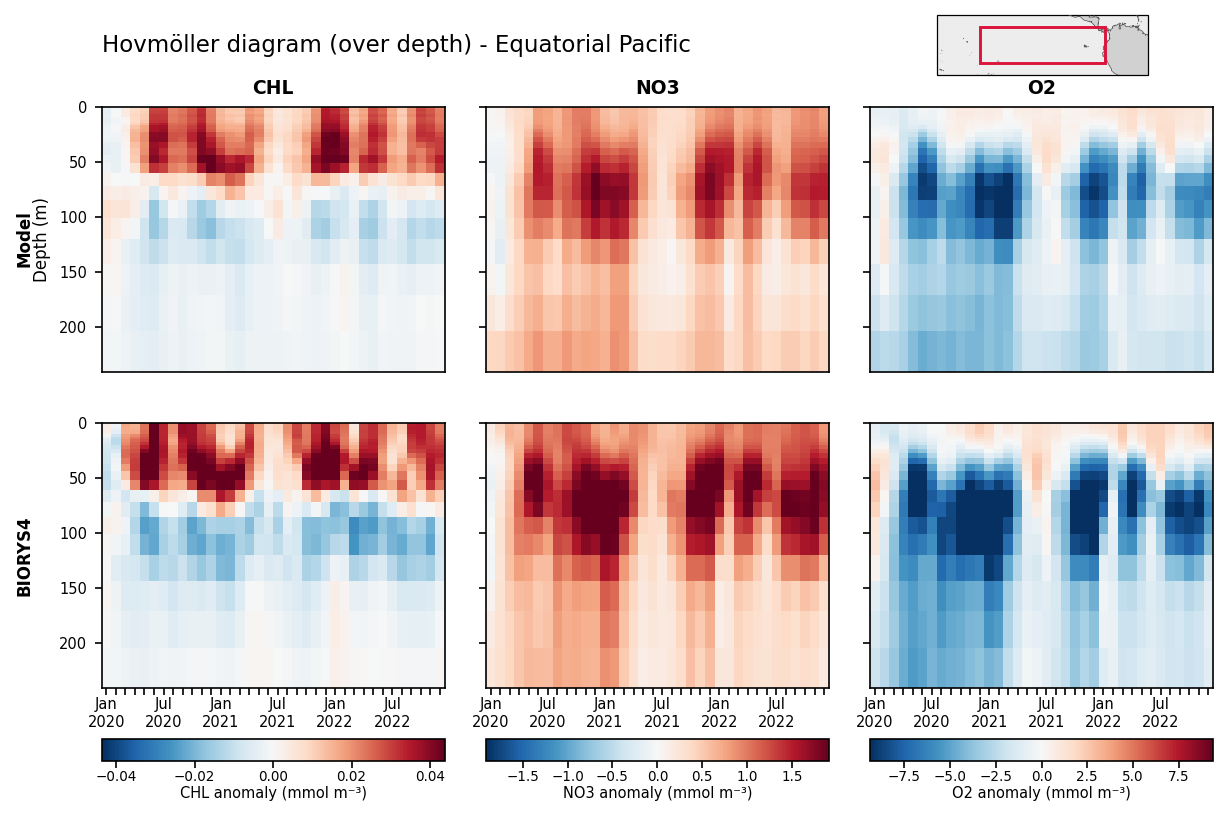}
  \caption{Hovmöller diagram of the regional-mean vertical anomalies of the fields shown in Figure~\ref{fig:hovmoller_eqpac_lat}.}
  \label{fig:hovmoller_eqpac_anom}
\end{figure}

\paragraph*{Hovmöller diagrams} Figures (\ref{fig:hovmoller_eqpac_lat},~\ref{fig:hovmoller_eqpac_anom}) show Hovmöller diagrams, plotting evolution along one spatial axis against time, of chlorophyll-a, nitrate and oxygen concentrations (over latitude) and corresponding anomalies (over depth) in the Equatorial Pacific. We show all 36 months between 2020--2022 at lead time one. Figure~\ref{fig:hovmoller_eqpac_lat} shows the latitude-time series over the region, calculated by averaging surface values over the longitude range. The figure illustrates how well not only the model, but also the climatology, represent the observed seasonal patterns in the region. Nonetheless, the model is able to match the reference time series more closely: high and low concentrations are better represented, and we see less smoothing. Figure~\ref{fig:hovmoller_eqpac_anom} shows the regional-mean depth-time series of the anomalies (i.e. after having subtracted the climatology), calculated by averaging over both longitude and latitude ranges at each depth. This illustrates performance beyond the seasonal cycle, showcasing the model's ability to capture the anomalies of the regional water column. Similar results are shown in the Appendix to illustrate performance in the North Atlantic Subtropical Gyre (Figures~\ref{fig:hovmoller_NASTG_lat},~\ref{fig:hovmoller_NASTG_anom}).

\paragraph*{Forecast samples}

\begin{figure}[htbp]
  \centering
  \includegraphics[width=\textwidth]{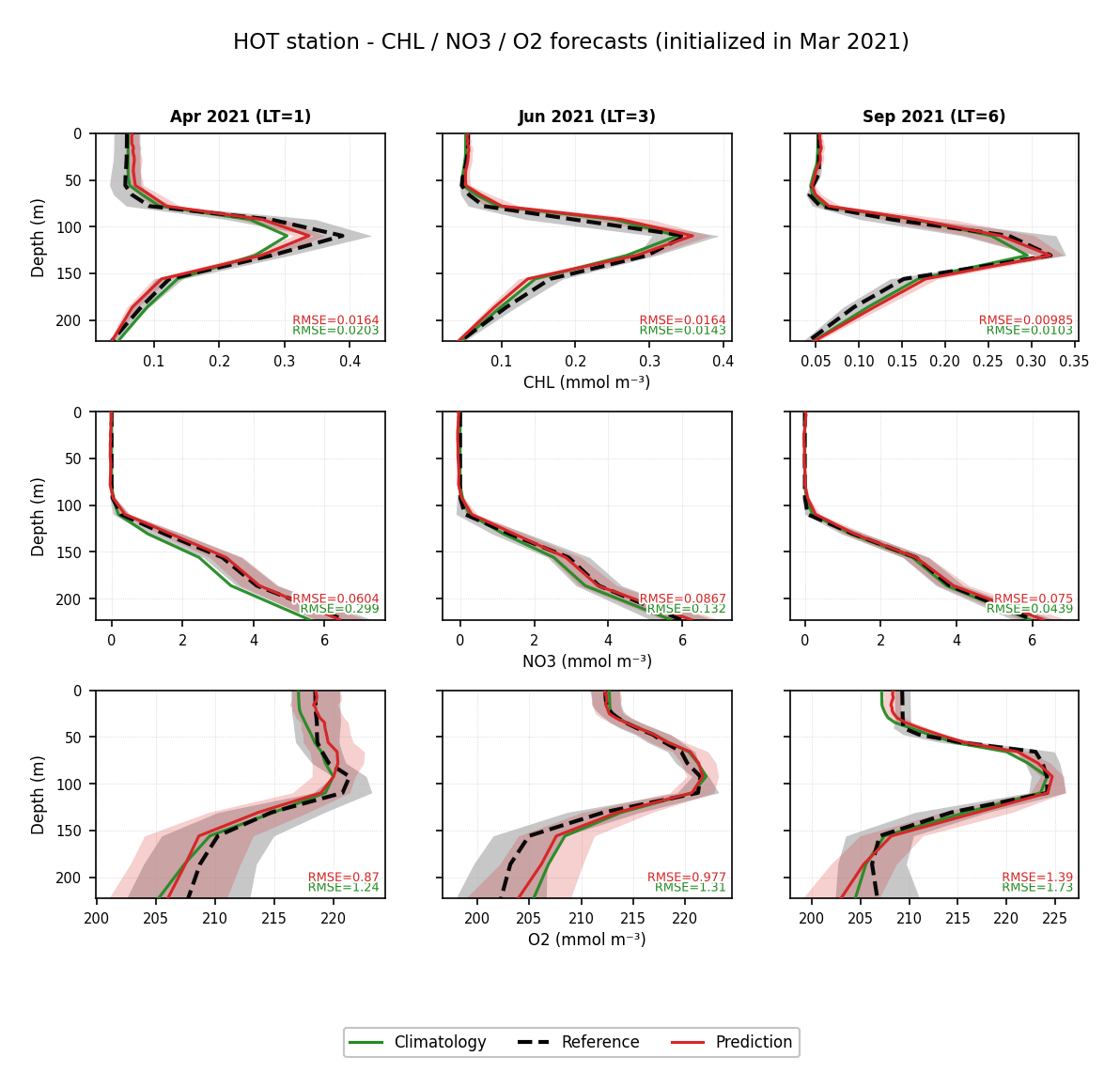}
  \caption{Chlorophyll-a, nitrate and oxygen samples from a global forecast at the HOT station. Shading illustrates the standard deviation for the reference and prediction calculated over a region spanning $\pm 2^\circ$ from the station.}
  \label{fig:HOT_profs}
\end{figure}

\begin{figure}[htbp]
  \centering
  \includegraphics[width=\textwidth]{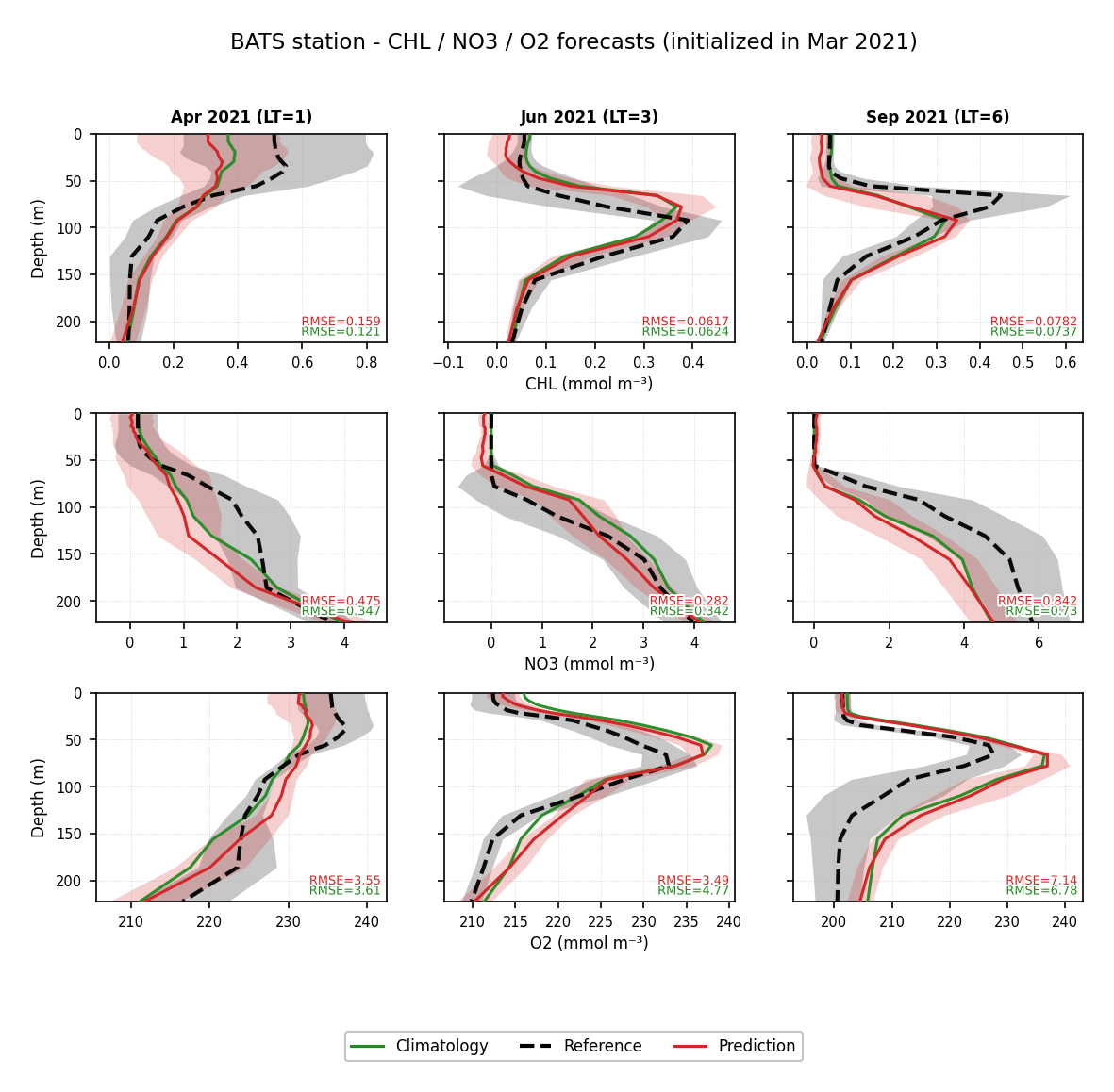}
  \caption{Samples from the same forecast as Figure~\ref{fig:HOT_profs} at the BATS station. Shading illustrates the regional standard deviation for the reference and prediction.}
  \label{fig:BATS_profs}
\end{figure}

Figures~\ref{fig:HOT_profs} and~\ref{fig:BATS_profs} illustrate chlorophyll-a, nitrate and oxygen depth profiles from a forecast  initialized on March 2021, showing lead times one (April 2021), three (June 2021), and six (September 2021) at the HOT (Hawaii Ocean Time-series) and BATS (Bermuda Atlantic Time-series) stations, respectively. The figures illustrate the prediction (red), the BIORYS4 reference (black), and the climatology (green), together with the standard deviation of the reference and prediction calculated over a region spanning $\pm 2^\circ$ in longitude and latitude from the station to illustrate regional spatial variability.

The figures show that the model is able to reproduce the overall vertical structure and magnitude of the reference profiles reasonably well. At HOT, RMSE values are lower than those of the climatology for the majority of months and lead times, while at BATS they are generally within the same range. For a global forecast evaluated at single point locations, this level of agreement is encouraging. The standard deviations further indicate that the prediction captures regional spatial variability comparable to that of the reference, with HOT matching the reference more closely than BATS.

\begin{figure}[htbp]
  \centering
  \includegraphics[width=\textwidth]{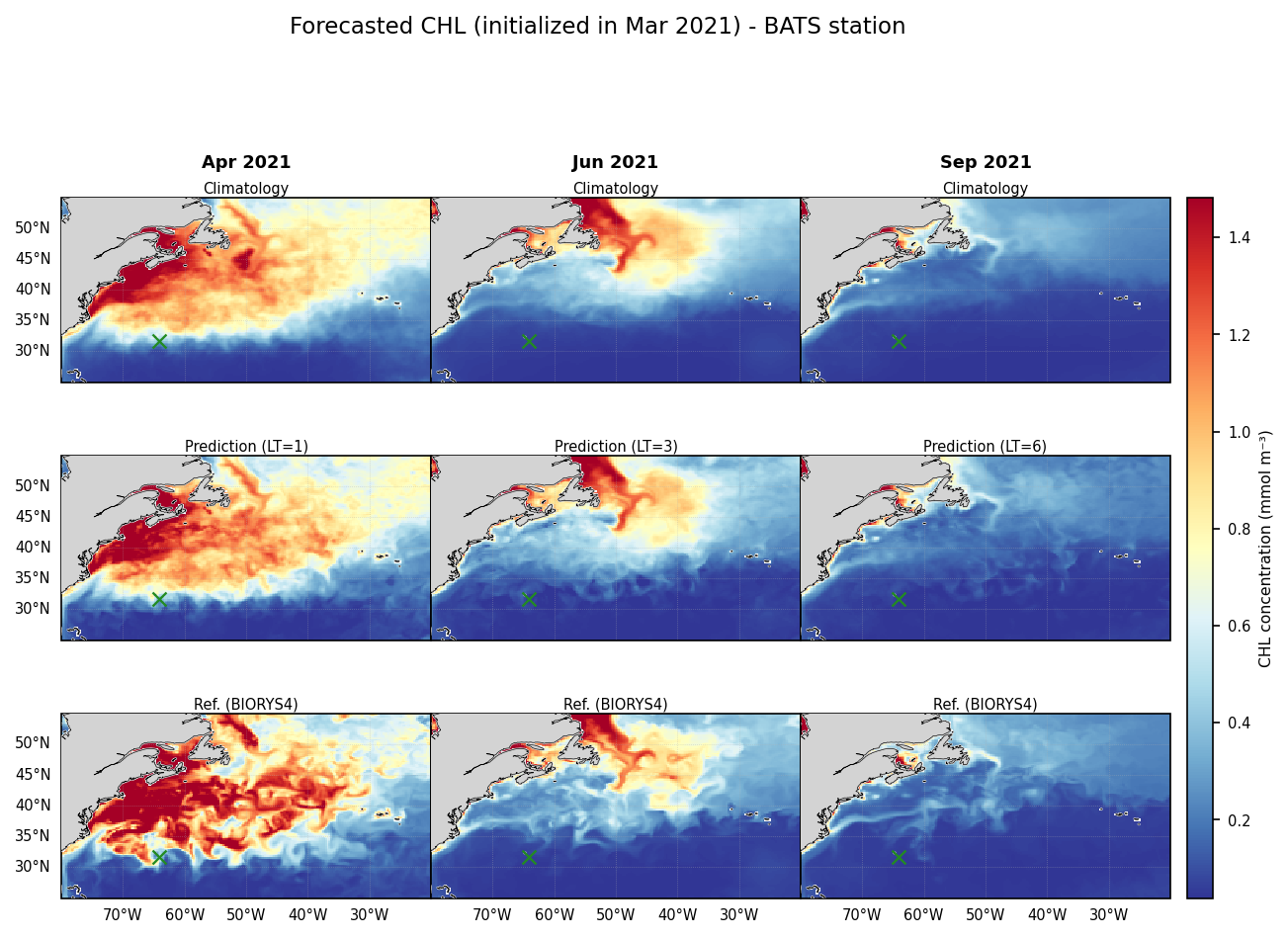}
  \caption{Forecasted, reference and climatology chlorophyll-a fields around the BATS station (marked with a green cross for reference).}
  \label{fig:BATS_chl}
\end{figure}

\begin{figure}[htbp]
  \centering
  \includegraphics[width=\textwidth]{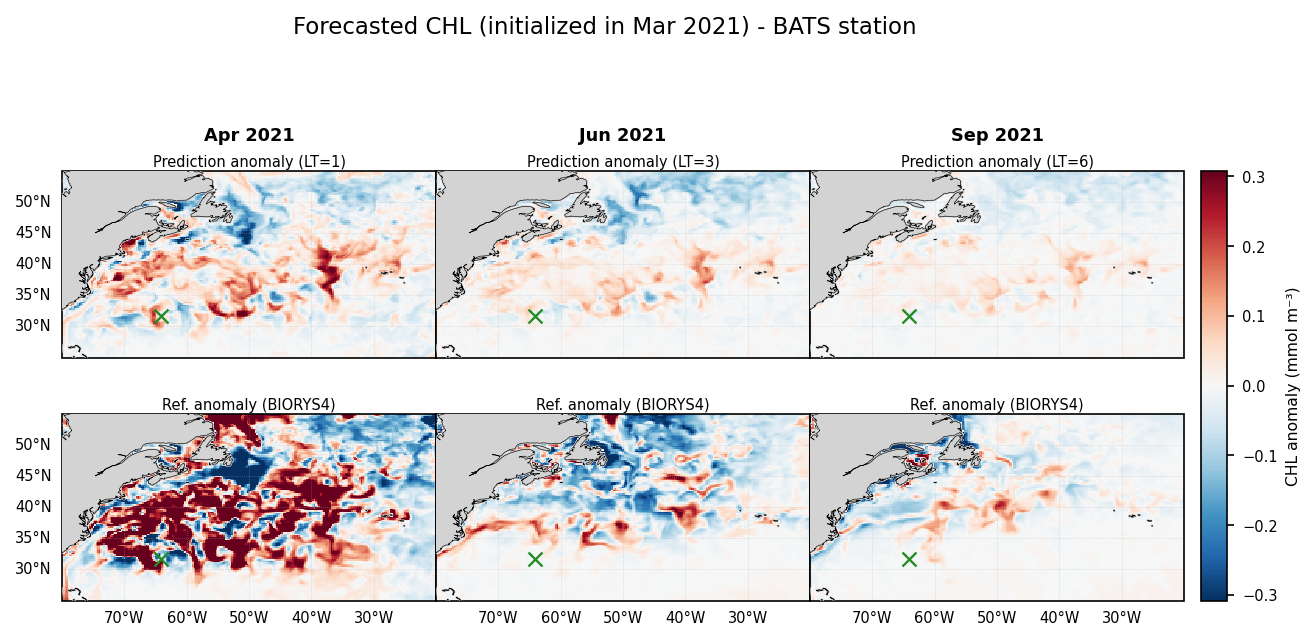}
  \caption{Forecasted and reference chlorophyll-a anomalies around the BATS station (marked with a green cross for reference).}
  \label{fig:BATS_chl_anom}
\end{figure}

Figures~\ref{fig:BATS_chl}--\ref{fig:BATS_chl_anom} illustrate the chlorophyll-a fields from the same forecast around the BATS station, which is marked with a green cross in the figures. The model captures the broad spatial pattern of the bloom anomalies reasonably well, accurately placing the general boundaries, but it underestimates the concentration peaks, and tends to converge towards the climatology at longer lead times (as illustrated by the smoothing that appears in lead time six in Fig.~\ref{fig:BATS_chl_anom}). In the Appendix, Figures~\ref{fig:BATS_no3}--\ref{fig:BATS_o2_anom} show a similar comparison for nitrate and oxygen in BATS, while Figures~\ref{fig:HOT_chl}--\ref{fig:HOT_o2_anom} illustrate the three variables in the HOT station. The figures show that while the model is able to capture the macro-level trend of the anomalies, it struggles to represent the smaller spatial structures, especially at longer lead times. In some cases (i.e. nitrate at lead times three and six, or chlorophyll-a in HOT), the mean anomaly is overestimated.

\paragraph*{Derived quantities}

BGC forecasts can be used to predict the ocean's carbon uptake capacity. Surface partial pressure of dissolved CO$_2$ (spCO$_2$) is a useful indicator of this because the difference in pressure between the ocean surface and atmosphere controls the direction and magnitude of air-sea exchange. While this variable is produced by BIORYS4, it is absent from the training data, as we treat is as a diagnostic: we calculate predicted spCO$_2$ from BG4Sea's alkalinity, dissolved inorganic carbon, Si and PO$_4$ forecasts, as well as prescribed temperature and salinity (from GLORYS12). This is done with the open-source PyCO2SYS package \citep{humphreysPyCO2SYSV182022} that is used to solve the carbonate system.

At lead time one (see Table~\ref{tab:spco2_metrics}), persistence dominates (ACC 0.723 vs.\ 0.521), but lead time six, however, the model-derived spCO$_2$ outperforms persistence in ACC (0.287 vs.\ 0.275), and outperforms both persistence and climatology in $R^2$. Given that spCO$_2$ is never used as a training target, these results demonstrate that the model can propagate internally consistent carbon state estimates long enough to anticipate the ocean carbon sink. We show a similar consistency check in the Appendix (Table~\ref{tab:dn_ratios}) that shows that the global mean diatom and nanophytoplankton fractions of total biomass and chlorophyll-a are well preserved by the model.

\subsection*{Evaluating the physics forecaster}

In Section~\ref{sec:surface_conditioning} we introduce a naïve forecaster that propagates the physical variables before passing them through the conditioning mechanism with a residual network, mimicking the latent column forecaster achitecture. Here we evaluate performance for this forecaster and its effect on the forecasted biogeochemistry.

\begin{table}[h]
  \centering
  \begin{tabular}{lcccc}
    \topline
                         & \textbf{Nutrients and oxygen}                                        & \textbf{Biology}                                                     & \textbf{Carbon pools}                                                & \textbf{Temp. \& Sal.}                                               \\
    \midline
    \textbf{ACC}         &                                                                      &                                                                      &                                                                      &                                                                      \\
    Persistence          & \textbf{0.627}\,$\bullet$\,\textbf{0.299}\,$\bullet$\,\textbf{0.215} & 0.358\,$\bullet$\,0.089\,$\bullet$\,\textbf{0.061}                   & 0.547\,$\bullet$\,0.226\,$\bullet$\,0.147                            & 0.604\,$\bullet$\,0.278\,$\bullet$\,0.199                            \\
    Horizontally-coupled & 0.604\,$\bullet$\,0.371\,$\bullet$\,0.280                            & 0.466\,$\bullet$\,0.165\,$\bullet$\,0.091                            & 0.644\,$\bullet$\,0.416\,$\bullet$\,0.337                            & N/A                                                                  \\
    +Forecasted forcings & 0.573\,$\bullet$\,0.277\,$\bullet$\,0.158                            & \textbf{0.442}\,$\bullet$\,\textbf{0.121}\,$\bullet$\,0.048          & \textbf{0.613}\,$\bullet$\,\textbf{0.337}\,$\bullet$\,\textbf{0.234} & \textbf{0.649}\,$\bullet$\,\textbf{0.377}\,$\bullet$\,\textbf{0.253} \\
    \midline
    \textbf{R$^2$}       &                                                                      &                                                                      &                                                                      &                                                                      \\
    Climatology          & 0.973\,$\bullet$\,0.973\,$\bullet$\,\textbf{0.973}                   & \textbf{0.821}\,$\bullet$\,\textbf{0.821}\,$\bullet$\,\textbf{0.821} & 0.918\,$\bullet$\,0.918\,$\bullet$\,0.918                            & 0.983\,$\bullet$\,0.983\,$\bullet$\,\textbf{0.983}                   \\
    Persistence          & \textbf{0.980}\,$\bullet$\,0.932\,$\bullet$\,0.899                   & 0.669\,$\bullet$\,0.089\,$\bullet$\,$-$0.203                         & 0.875\,$\bullet$\,0.637\,$\bullet$\,0.471                            & 0.986\,$\bullet$\,0.950\,$\bullet$\,0.923                            \\
    Horizontally-coupled & 0.980\,$\bullet$\,0.976\,$\bullet$\,0.974                            & 0.815\,$\bullet$\,0.796\,$\bullet$\,0.791                            & 0.943\,$\bullet$\,0.926\,$\bullet$\,0.921                            & N/A                                                                  \\
    +Forecasted forcings & \textbf{0.980}\,$\bullet$\,\textbf{0.975}\,$\bullet$\,\textbf{0.974} & 0.818\,$\bullet$\,0.803\,$\bullet$\,0.793                            & \textbf{0.940}\,$\bullet$\,\textbf{0.923}\,$\bullet$\,\textbf{0.919} & \textbf{0.990}\,$\bullet$\,\textbf{0.985}\,$\bullet$\,\textbf{0.983} \\
    \midline
    \textbf{NRMSE}       &                                                                      &                                                                      &                                                                      &                                                                      \\
    Climatology          & 0.152\,$\bullet$\,0.152\,$\bullet$\,\textbf{0.152}                   & 0.418\,$\bullet$\,\textbf{0.418}\,$\bullet$\,\textbf{0.418}          & 0.255\,$\bullet$\,0.255\,$\bullet$\,0.255                            & 0.127\,$\bullet$\,0.127\,$\bullet$\,0.127                            \\
    Persistence          & 0.132\,$\bullet$\,0.246\,$\bullet$\,0.298                            & 0.568\,$\bullet$\,0.947\,$\bullet$\,1.089                            & 0.296\,$\bullet$\,0.517\,$\bullet$\,0.625                            & 0.118\,$\bullet$\,0.219\,$\bullet$\,0.269                            \\
    Horizontally-coupled & 0.125\,$\bullet$\,0.143\,$\bullet$\,0.147                            & 0.387\,$\bullet$\,0.428\,$\bullet$\,0.435                            & 0.202\,$\bullet$\,0.236\,$\bullet$\,0.244                            & N/A                                                                  \\
    +Forecasted forcings & \textbf{0.128}\,$\bullet$\,\textbf{0.147}\,$\bullet$\,\textbf{0.151} & \textbf{0.390}\,$\bullet$\,0.427\,$\bullet$\,0.435                   & \textbf{0.208}\,$\bullet$\,\textbf{0.243}\,$\bullet$\,\textbf{0.251} & \textbf{0.096}\,$\bullet$\,\textbf{0.118}\,$\bullet$\,\textbf{0.124} \\
    \botline
  \end{tabular}
  \caption{Group-mean forecast metrics at lead times 1, 3, and 6 months (shown as $x$\,$\bullet$\,$x$\,$\bullet$\,$x$) for the coupled model (+Forecasted forcings). Persistence, climatology, and the original architecture (with prescribed physical forcings) are shown for reference. Best metric values, excluding the prescribed forcings model, are shown in bold.}\label{tab:forecast_groups_coupled}
\end{table}

Table~\ref{tab:forecast_groups_coupled} compares performance for the forecaster that receives the prescribed physics embedding (corresponding to the results discussed above, see Table~\ref{tab:forecast_groups}), and the forecaster that internally propagates said forcings forward in time. We also examine forecast skill for temperature and salinity for this variant. The forecaster shows acceptable performance for its simplicity, outperforming persistence and climatology for the forecasted variables (i.e. temperature and salinity) at all lead times. Spatially (Appendix, Figures~\ref{fig:acc_physsurf}--\ref{fig:acc_phys220}), both variables show some skill at the surface by lead time three, but this deteriorates by lead time six. The degradation is more pronounced for both variables at 220\,m.

The impact on BGC skill can be observed when examining ACC: nutrients and oxygen are now worse than persistence at all lead times and by lead time six their ACC has degraded from 0.280 to 0.158. The carbon pools degrade from 0.337 to 0.234, and biology from 0.091 to 0.048. These results suggest that our prototype might be insufficient for the downstream task, but more importantly, that poorly constrained physical forcings result in a significant degradation of BGC anomaly skill.

\section{Discussion}

This work introduces BG4Sea, which we believe to be the first global, data-driven system to produce multivariate seasonal forecasts of the marine BGC state, and which was developed as an instrument for decomposing where seasonal BGC predictability comes from. We focus the rest of this section on where this baseline succeeds, and where it falls short.

\subsection*{Modularity and sources of skill}

The column-only model lets us examine how much of the water column can be compressed into a low-dimensional state, and how much of that can be predicted from its previous state. The autoencoder's reconstruction errors (Tables~\ref{tab:ae_metrics_chem}--\ref{tab:ae_metrics_phys}) anticipate which variables will perform better or worse in the forecast, simoultaneously highlighting which processes are more or less coupled to the dominant modes captured by the compression. Silicate is a clear example of a poorly represented variable: it has the worst anomaly reconstruction in the entire dataset $R^2 = 0.264$, and it is the only nutrient for which persistence outperforms at every lead time (Table~\ref{tab:forecast_acc_by_var}). One would expect this strong persistence to translate into higher predictability; a variable that changes slowly should be easier to predict. However, because silicate shares a low-dimensional bottleneck with tracers that are more strongly interconnected with one another than with silicate itself, its distinct variability is likely suppressed when the model minimizes the aggregate reconstruction error. A natural next step would be to give these processes a more differentiated latent representation in order to better retain their variability.

Nitrate illustrates the opposite case: with anomaly reconstruction $R^2 = 0.947$, its variability is well-captured by the shared bottleneck, and the model outperforms persistence by lead time three, reaching ACC $= 0.384$ at six months against persistence's $0.129$. Total dissolved iron shows a distinct and interpretable pattern: its anomaly reconstruction ($R^2 = 0.747$) is lower than most nutrients, but unlike silicate, the forecast retains modest skill over persistence. This gap is not primarily a bottleneck-competition effect, but rather linked to the specific dynamics of this variable. Iron's variability is driven by episodic, regionally concentrated dust-deposition events, and our model's performance is consistent with the broader difficulty BGC models have in capturing these processes~\citep{tagliabueHowWellGlobal2016}. 

Turning to the skill contributed by each module, we find that every component yields a measurable and interpretable gain. Physical surface forcings provide the largest improvement in ACC: at lead time six, the nutrients and oxygen group rises from $0.101 \rightarrow 0.261$ while the carbon pools improve from $0.164 \rightarrow 0.302$. When prescribed physics are replaced by the model's own forecast, skill degrades back to 0.158 for nutrients and oxygen and 0.234 for carbon pools, highlighting how tightly BGC processes depend on well-resolved physics. 

Finally, while coupling to the immediate horizontal neighbourhood adds only modest skill, it still shows that lateral structure matters for predictability. Except for nutrients and oxygen at lead time one, every group performs best at every lead time once these lateral interactions are included; the same pattern holds at the variable level (Table~\ref{tab:forecast_acc_by_var}). Extending the neighbourhood to more distant grid points would be a natural next step, but preliminary experiments showed that the model struggled to use the extra information, likely because the training window is short and the data are monthly. We saw a similar outcome when lengthening the forecaster's temporal history: overfitting increased and many runs failed to converge.

Nonetheless, despite its limitations, the modular design achieves its main goal: each source of information is  attributable, and the progression from column-only to surface-forced to horizontally-coupled coherently shows where seasonal BGC predictability originates.

\subsection*{Model skill and intrinsic predictability}

Most variables show a combination of high $R^2$ and rapidly decaying ACC. At lead time six, the three groups have ACC below 0.34, but $R^2$ near 0.8, because the seasonal cycle dominates total variance while anomaly persistence tends to be weak. The biology group is the clearest example. The autoencoder reconstructs its anomalies well ($R^2 \approx 0.9$) but we see a rapid degradation of predictability: by lead time three persistence reaches only $ACC=0.09$ and the column-only model only $ACC=0.108$. Adding physical forcings and then horizontal coupling improves skill modestly; by lead time six, the full model still achieves only 0.09, compared to 0.061 for persistence, all while $R^2$ remains at 0.79. These results point to intrinsic predictability limits for biological variables at monthly resolution, where bloom onset is fast and the system is highly sensitive to initial conditions.

At the same time, BG4Sea's design encourages the forecast to regress toward the mean as lead time increases. Training with a pointwise MSE loss tends to push predictions to the conditional mean of the target, a well known limitation in the machine learning forecasting literature~\citep{price_gencast_2025}. Table~\ref{tab:forecast_varratio_by_var} shows that the predicted anomaly variance ratio falls rapidly with lead time, approaching zero as anomalies shrink toward climatology (which itself has zero anomaly variance). The same behaviour appears in Figures~\ref{fig:BATS_chl}--\ref{fig:BATS_chl_anom}, and~\ref{fig:BATS_no3}--\ref{fig:HOT_o2_anom}:  forecasts become smoother at longer horizons, capturing the sign of some anomalies but not their amplitude. This was a known limitation from the outset; our first goal was to establish a minimalist baseline before moving to more computationally demanding probabilistic extensions, but we recognise that variance collapse might contribute to the limited forecast skill reported above.

\subsection*{Limitations and future work}

\paragraph*{Deterministic approach for poorly constrained fields}
One of the main limitations follows directly from the variance collapse discussed above. A model that learns a deterministic function to map a poorly constrained field will, by construction, produce overconfident estimations, because it is forced to choose a single outcome when multiple plausible possibilities exist. It may suppress extremes by averaging over the tails of the distribution, failing precisely when outcomes are most consequential. Probabilistic extensions would likely reduce this problem, improving predictive skill, but we nevertheless decided to begin with a deterministic architecture because it provides a simple benchmark. We plan to pursue generative forecasting approaches in future work.

\paragraph*{Limits to modularity}
BG4Sea is built from two main modules trained separately (Section~\ref{sec:methods}): each forecaster is trained against a frozen, previously trained autoencoder. This staged design is cheaper and faster, each block can be trained, ablated, and reused independently, which also makes the system more generalisable in principle. The trade-off is that autoencoder learns what is optimal only for its own objective, not necessarily for downstream outputs. While it performs well at reconstructing most variables (Tables~\ref{tab:ae_metrics_chem}--\ref{tab:ae_metrics_phys}), and skill can improve when additional information is added (Table~\ref{tab:forecast_groups}), the frozen bottleneck limits how much the dynamics modules can recover.

\paragraph*{Limited data}
We use monthly averaged fields from BIORYS4 at a $1/4^\circ$ resolution. This means that our training and validation period only include 18 years $\times$ 12 months time steps. We further reduce the temporal window for our evaluation to only four months per year. While this was done in order to limit the computational footprint of global forecasts for different model variants and lead times, it also limits our capacity to evaluate the system as robustly as we would like. We compensate by pooling anomalies across all spatiotemporal grid points to provide more degrees of freedom to our metric estimates, but the
short verification window means we can only characterise short-range predictability. Data scarcity limits the model as well: preliminary experiments that expanded the horizontal neighbourhood or lengthened the input temporal history were more prone to overfitting and frequently failed to converge.

A related concern is non-stationarity. BG4Sea learns from a limited temporal sample, yet the ocean has continued to warm, acidify, and deoxygenate, with some of these trends accelerating. While our 2020--2022 evaluation period provides some evidence of robustness to near-term drift, as these trends accelerate, the relationships that the model learned during training could change, and periodic updating will be needed to maintain robustness.

Finally, because training and evaluation are done entirely on BIORYS4, BG4Sea functions as an emulator of the PISCES-based reanalysis rather than direct observations. This includes its biases and structural limitations. Priority next steps include validation against independent observations, and training across more varied data sources.

\section{Conclusion}

BG4Sea shows that seasonal marine BGC forecasting can be meaningfully decomposed into separable, learnable sources of skill: local column structure, surface physical forcing, and horizontal coupling. Each contributes measurable improvement, and the resulting system, to our knowledge, the first global, multivariate, data-driven seasonal BGC forecaster, beats persistence and climatology across most variables and lead times. We hope BG4Sea can help provide a starting point and a diagnostic tool for the more expressive architectures that will likely follow.

\clearpage
\acknowledgments
We thank Coralie Perruche and Julien Lamouroux for their guidance and for facilitating access to the BIORYS4 dataset used in this work. This research was supported by the European Union's Horizon Europe Programme under Grant Agreement no.\ 101112823 (DTO-BioFlow), and by the NECCTON project (``New Copernicus Capability for Trophic Ocean Networks''), funded under Horizon Europe RIA grant no.\ 101081273.

%
%
\datastatement
The Global Ocean Physics Reanalysis (GLOBAL\_MULTIYEAR\_PHY\_001\_030; \citealt{cmems_phys}) was used as the source of physical forcing variables. The biogeochemical fields used for training and evaluation are derived from an intermediate simulation extending the Global Ocean Biogeochemistry Analysis and Forecast (GLOBAL\_ANALYSISFORECAST\_BGC\_001\_028; \citealt{cmems_bio}), containing the full set of biogeochemical tracers produced by the NEMO/PISCES configuration.
This intermediate product is not yet publicly available (expected release through CMEMS by the end of 2026); it is discussed in \citep{gregoire_review_2026}.





\appendix[A]

\appendixtitle{Additional results}

\subsection*{Forecast skill through progressive information scaling}

\begin{sidewaystable}
  \centering
  \begin{tabular}{lcccccc}
    \topline
                                  & \textbf{Climatology}                & \textbf{Persistence}                                                 & \shortstack{\textbf{Column-}                                                                                                                                                                                                                                   \\\textbf{only}} & \shortstack{\textbf{+ Surface-}\\\textbf{forcing}} & \shortstack{\textbf{+ Horizontal-}\\\textbf{coupling}} & \shortstack{\textbf{+ Forecasted}\\\textbf{forcings}} \\
    \midline
    \textbf{Nutrients and oxygen} &                                     &                                                                      &                                           &                                                                      &                                                                      &                                                                      \\
    Nitrate                       & ---\,$\bullet$\,---\,$\bullet$\,--- & 0.608\,$\bullet$\,0.225\,$\bullet$\,0.129                            & 0.657\,$\bullet$\,0.311\,$\bullet$\,0.112 & 0.706\,$\bullet$\,0.475\,$\bullet$\,0.355                            & \textbf{0.709}\,$\bullet$\,\textbf{0.48}\,$\bullet$\,\textbf{0.384}  & 0.662\,$\bullet$\,0.343\,$\bullet$\,0.208                            \\
    Phosphate                     & ---\,$\bullet$\,---\,$\bullet$\,--- & 0.637\,$\bullet$\,0.257\,$\bullet$\,0.158                            & 0.659\,$\bullet$\,0.271\,$\bullet$\,0.045 & 0.704\,$\bullet$\,0.465\,$\bullet$\,0.35                             & \textbf{0.706}\,$\bullet$\,\textbf{0.469}\,$\bullet$\,\textbf{0.368} & 0.665\,$\bullet$\,0.351\,$\bullet$\,0.22                             \\
    Silicate                      & ---\,$\bullet$\,---\,$\bullet$\,--- & \textbf{0.902}\,$\bullet$\,\textbf{0.731}\,$\bullet$\,\textbf{0.635} & 0.457\,$\bullet$\,0.282\,$\bullet$\,0.186 & 0.456\,$\bullet$\,0.296\,$\bullet$\,0.199                            & 0.45\,$\bullet$\,0.289\,$\bullet$\,0.208                             & 0.442\,$\bullet$\,0.268\,$\bullet$\,0.172                            \\
    Total dissolved iron          & ---\,$\bullet$\,---\,$\bullet$\,--- & 0.463\,$\bullet$\,0.126\,$\bullet$\,0.065                            & 0.47\,$\bullet$\,0.156\,$\bullet$\,0.061  & 0.473\,$\bullet$\,0.167\,$\bullet$\,0.066                            & \textbf{0.481}\,$\bullet$\,\textbf{0.176}\,$\bullet$\,\textbf{0.083} & 0.472\,$\bullet$\,0.152\,$\bullet$\,0.057                            \\
    Oxygen                        & ---\,$\bullet$\,---\,$\bullet$\,--- & 0.524\,$\bullet$\,0.157\,$\bullet$\,0.086                            & 0.619\,$\bullet$\,0.259\,$\bullet$\,0.101 & 0.67\,$\bullet$\,0.432\,$\bullet$\,0.336                             & \textbf{0.674}\,$\bullet$\,\textbf{0.442}\,$\bullet$\,\textbf{0.358} & 0.622\,$\bullet$\,0.27\,$\bullet$\,0.135                             \\
    \textit{Group mean}           & ---\,$\bullet$\,---\,$\bullet$\,--- & \textbf{0.627}\,$\bullet$\,0.299\,$\bullet$\,0.215                   & 0.572\,$\bullet$\,0.256\,$\bullet$\,0.101 & 0.602\,$\bullet$\,0.367\,$\bullet$\,0.261                            & 0.604\,$\bullet$\,\textbf{0.371}\,$\bullet$\,\textbf{0.28}           & 0.573\,$\bullet$\,0.277\,$\bullet$\,0.158                            \\
    \midline
    \textbf{Biology}              &                                     &                                                                      &                                           &                                                                      &                                                                      &                                                                      \\
    Mesozooplankton               & ---\,$\bullet$\,---\,$\bullet$\,--- & 0.431\,$\bullet$\,0.128\,$\bullet$\,0.083                            & 0.533\,$\bullet$\,0.151\,$\bullet$\,0.058 & 0.556\,$\bullet$\,0.211\,$\bullet$\,0.069                            & \textbf{0.564}\,$\bullet$\,\textbf{0.247}\,$\bullet$\,\textbf{0.146} & 0.544\,$\bullet$\,0.183\,$\bullet$\,0.081                            \\
    Microzooplankton              & ---\,$\bullet$\,---\,$\bullet$\,--- & 0.377\,$\bullet$\,0.096\,$\bullet$\,0.072                            & 0.501\,$\bullet$\,0.126\,$\bullet$\,0.058 & 0.527\,$\bullet$\,0.192\,$\bullet$\,0.087                            & \textbf{0.53}\,$\bullet$\,\textbf{0.209}\,$\bullet$\,\textbf{0.141}  & 0.508\,$\bullet$\,0.159\,$\bullet$\,0.088                            \\
    Diatoms                       & ---\,$\bullet$\,---\,$\bullet$\,--- & 0.354\,$\bullet$\,0.079\,$\bullet$\,\textbf{0.054}                   & 0.439\,$\bullet$\,0.095\,$\bullet$\,0.025 & 0.461\,$\bullet$\,0.118\,$\bullet$\,0.029                            & \textbf{0.463}\,$\bullet$\,\textbf{0.124}\,$\bullet$\,0.049          & 0.44\,$\bullet$\,0.098\,$\bullet$\,0.026                             \\
    Nanophytoplankton             & ---\,$\bullet$\,---\,$\bullet$\,--- & 0.259\,$\bullet$\,0.044\,$\bullet$\,0.025                            & 0.365\,$\bullet$\,0.076\,$\bullet$\,0.028 & 0.382\,$\bullet$\,0.094\,$\bullet$\,0.052                            & \textbf{0.393}\,$\bullet$\,\textbf{0.128}\,$\bullet$\,\textbf{0.079} & 0.368\,$\bullet$\,0.077\,$\bullet$\,0.027                            \\
    Chlorophyll                   & ---\,$\bullet$\,---\,$\bullet$\,--- & 0.344\,$\bullet$\,0.078\,$\bullet$\,\textbf{0.058}                   & 0.424\,$\bullet$\,0.089\,$\bullet$\,0.025 & 0.448\,$\bullet$\,0.118\,$\bullet$\,0.028                            & \textbf{0.452}\,$\bullet$\,\textbf{0.133}\,$\bullet$\,0.057          & 0.423\,$\bullet$\,0.094\,$\bullet$\,0.02                             \\
    Net primary production        & ---\,$\bullet$\,---\,$\bullet$\,--- & 0.385\,$\bullet$\,0.111\,$\bullet$\,0.072                            & 0.359\,$\bullet$\,0.108\,$\bullet$\,0.037 & 0.385\,$\bullet$\,0.117\,$\bullet$\,0.005                            & \textbf{0.391}\,$\bullet$\,\textbf{0.149}\,$\bullet$\,\textbf{0.076} & 0.369\,$\bullet$\,0.116\,$\bullet$\,0.045                            \\
    \textit{Group mean}           & ---\,$\bullet$\,---\,$\bullet$\,--- & 0.358\,$\bullet$\,0.089\,$\bullet$\,0.061                            & 0.437\,$\bullet$\,0.108\,$\bullet$\,0.038 & 0.46\,$\bullet$\,0.142\,$\bullet$\,0.045                             & \textbf{0.466}\,$\bullet$\,\textbf{0.165}\,$\bullet$\,\textbf{0.091} & 0.442\,$\bullet$\,0.121\,$\bullet$\,0.048                            \\
    \midline
    \textbf{Carbon Pools}         &                                     &                                                                      &                                           &                                                                      &                                                                      &                                                                      \\
    Dissolved inorganic carbon    & ---\,$\bullet$\,---\,$\bullet$\,--- & 0.682\,$\bullet$\,0.336\,$\bullet$\,0.232                            & 0.75\,$\bullet$\,0.483\,$\bullet$\,0.231  & \textbf{0.793}\,$\bullet$\,\textbf{0.651}\,$\bullet$\,\textbf{0.596} & 0.792\,$\bullet$\,0.641\,$\bullet$\,0.581                            & 0.759\,$\bullet$\,0.562\,$\bullet$\,0.49                             \\
    Alkalinity                    & ---\,$\bullet$\,---\,$\bullet$\,--- & 0.653\,$\bullet$\,0.311\,$\bullet$\,0.198                            & 0.654\,$\bullet$\,0.359\,$\bullet$\,0.209 & 0.685\,$\bullet$\,0.448\,$\bullet$\,0.345                            & \textbf{0.686}\,$\bullet$\,\textbf{0.449}\,$\bullet$\,\textbf{0.368} & 0.658\,$\bullet$\,0.37\,$\bullet$\,0.245                             \\
    Calcium carbonate             & ---\,$\bullet$\,---\,$\bullet$\,--- & 0.354\,$\bullet$\,0.117\,$\bullet$\,0.084                            & 0.433\,$\bullet$\,0.167\,$\bullet$\,0.096 & 0.455\,$\bullet$\,0.185\,$\bullet$\,0.074                            & \textbf{0.46}\,$\bullet$\,\textbf{0.218}\,$\bullet$\,\textbf{0.134}  & 0.432\,$\bullet$\,0.168\,$\bullet$\,0.094                            \\
    Particulate organic carbon    & ---\,$\bullet$\,---\,$\bullet$\,--- & 0.422\,$\bullet$\,0.127\,$\bullet$\,0.088                            & 0.526\,$\bullet$\,0.142\,$\bullet$\,0.049 & 0.545\,$\bullet$\,0.21\,$\bullet$\,0.078                             & \textbf{0.549}\,$\bullet$\,\textbf{0.238}\,$\bullet$\,\textbf{0.16}  & 0.531\,$\bullet$\,0.189\,$\bullet$\,0.098                            \\
    Dissolved organic carbon      & ---\,$\bullet$\,---\,$\bullet$\,--- & 0.622\,$\bullet$\,0.24\,$\bullet$\,0.134                             & 0.683\,$\bullet$\,0.392\,$\bullet$\,0.234 & 0.729\,$\bullet$\,0.53\,$\bullet$\,0.417                             & \textbf{0.731}\,$\bullet$\,\textbf{0.534}\,$\bullet$\,\textbf{0.445} & 0.684\,$\bullet$\,0.395\,$\bullet$\,0.244                            \\
    \textit{Group mean}           & ---\,$\bullet$\,---\,$\bullet$\,--- & 0.547\,$\bullet$\,0.226\,$\bullet$\,0.147                            & 0.609\,$\bullet$\,0.309\,$\bullet$\,0.164 & 0.641\,$\bullet$\,0.405\,$\bullet$\,0.302                            & \textbf{0.644}\,$\bullet$\,\textbf{0.416}\,$\bullet$\,\textbf{0.337} & 0.613\,$\bullet$\,0.337\,$\bullet$\,0.234                            \\
    \midline
    \textbf{Physics}              &                                     &                                                                      &                                           &                                                                      &                                                                      &                                                                      \\
    Temperature                   & ---\,$\bullet$\,---\,$\bullet$\,--- & 0.51\,$\bullet$\,0.178\,$\bullet$\,0.119                             & ---\,$\bullet$\,---\,$\bullet$\,---       & ---\,$\bullet$\,---\,$\bullet$\,---                                  & ---\,$\bullet$\,---\,$\bullet$\,---                                  & \textbf{0.651}\,$\bullet$\,\textbf{0.352}\,$\bullet$\,\textbf{0.227} \\
    Salinity                      & ---\,$\bullet$\,---\,$\bullet$\,--- & \textbf{0.697}\,$\bullet$\,0.378\,$\bullet$\,\textbf{0.278}          & ---\,$\bullet$\,---\,$\bullet$\,---       & ---\,$\bullet$\,---\,$\bullet$\,---                                  & ---\,$\bullet$\,---\,$\bullet$\,---                                  & 0.647\,$\bullet$\,\textbf{0.401}\,$\bullet$\,\textbf{0.278}          \\
    \textit{Group mean}           & ---\,$\bullet$\,---\,$\bullet$\,--- & 0.604\,$\bullet$\,0.278\,$\bullet$\,0.199                            & ---\,$\bullet$\,---\,$\bullet$\,---       & ---\,$\bullet$\,---\,$\bullet$\,---                                  & ---\,$\bullet$\,---\,$\bullet$\,---                                  & \textbf{0.649}\,$\bullet$\,\textbf{0.377}\,$\bullet$\,\textbf{0.253} \\
    \botline
  \end{tabular}
  \caption{Per-variable ACC at lead times 1, 3, and 6 months ($x$\,$\bullet$\,$x$\,$\bullet$\,$x$), by group and model variant, averaged over 2020--2022. Climatology is not applicable for ACC. Physics variables evaluated only for persistence and forecasted forcings. Best value at each lead time in bold.}\label{tab:forecast_acc_by_var}
\end{sidewaystable}

\begin{sidewaystable}
  \centering
  \begin{tabular}{lcccccc}
    \topline
                                  & \textbf{Climatology}                                                 & \textbf{Persistence}                                        & \shortstack{\textbf{Column-}                                                                                                                                                                                                                                   \\\textbf{only}} & \shortstack{\textbf{+ Surface-}\\\textbf{forcing}} & \shortstack{\textbf{+ Horizontal-}\\\textbf{coupling}} & \shortstack{\textbf{+ Forecasted}\\\textbf{forcings}} \\
    \midline
    \textbf{Nutrients and oxygen} &                                                                      &                                                             &                                           &                                                                      &                                                                      &                                                                      \\
    Nitrate                       & 0.115\,$\bullet$\,0.115\,$\bullet$\,0.115                            & 0.11\,$\bullet$\,0.209\,$\bullet$\,0.254                    & 0.086\,$\bullet$\,0.109\,$\bullet$\,0.115 & 0.082\,$\bullet$\,\textbf{0.101}\,$\bullet$\,\textbf{0.107}          & \textbf{0.081}\,$\bullet$\,\textbf{0.101}\,$\bullet$\,\textbf{0.107} & 0.086\,$\bullet$\,0.108\,$\bullet$\,0.112                            \\
    Phosphate                     & 0.12\,$\bullet$\,0.12\,$\bullet$\,0.12                               & 0.11\,$\bullet$\,0.205\,$\bullet$\,0.246                    & 0.091\,$\bullet$\,0.116\,$\bullet$\,0.125 & \textbf{0.086}\,$\bullet$\,\textbf{0.107}\,$\bullet$\,\textbf{0.113} & \textbf{0.086}\,$\bullet$\,\textbf{0.107}\,$\bullet$\,\textbf{0.113} & 0.09\,$\bullet$\,0.113\,$\bullet$\,0.118                             \\
    Silicate                      & 0.217\,$\bullet$\,0.217\,$\bullet$\,0.217                            & \textbf{0.086}\,$\bullet$\,\textbf{0.169}\,$\bullet$\,0.217 & 0.199\,$\bullet$\,0.211\,$\bullet$\,0.215 & 0.198\,$\bullet$\,0.209\,$\bullet$\,\textbf{0.213}                   & 0.199\,$\bullet$\,0.21\,$\bullet$\,0.214                             & 0.2\,$\bullet$\,0.211\,$\bullet$\,0.215                              \\
    Total dissolved iron          & 0.203\,$\bullet$\,0.203\,$\bullet$\,\textbf{0.203}                   & 0.235\,$\bullet$\,0.409\,$\bullet$\,0.488                   & 0.179\,$\bullet$\,0.201\,$\bullet$\,0.204 & 0.179\,$\bullet$\,0.201\,$\bullet$\,0.205                            & \textbf{0.178}\,$\bullet$\,\textbf{0.2}\,$\bullet$\,\textbf{0.203}   & 0.179\,$\bullet$\,0.201\,$\bullet$\,\textbf{0.203}                   \\
    Oxygen                        & 0.106\,$\bullet$\,0.106\,$\bullet$\,0.106                            & 0.118\,$\bullet$\,0.236\,$\bullet$\,0.286                   & 0.084\,$\bullet$\,0.103\,$\bullet$\,0.106 & 0.08\,$\bullet$\,0.097\,$\bullet$\,0.101                             & \textbf{0.079}\,$\bullet$\,\textbf{0.096}\,$\bullet$\,\textbf{0.099} & 0.083\,$\bullet$\,0.103\,$\bullet$\,0.106                            \\
    \textit{Group mean}           & 0.152\,$\bullet$\,0.152\,$\bullet$\,0.152                            & 0.132\,$\bullet$\,0.246\,$\bullet$\,0.298                   & 0.128\,$\bullet$\,0.148\,$\bullet$\,0.153 & \textbf{0.125}\,$\bullet$\,\textbf{0.143}\,$\bullet$\,0.148          & \textbf{0.125}\,$\bullet$\,\textbf{0.143}\,$\bullet$\,\textbf{0.147} & 0.128\,$\bullet$\,0.147\,$\bullet$\,0.151                            \\
    \midline
    \textbf{Biology}              &                                                                      &                                                             &                                           &                                                                      &                                                                      &                                                                      \\
    Mesozooplankton               & 0.358\,$\bullet$\,0.358\,$\bullet$\,\textbf{0.358}                   & 0.457\,$\bullet$\,0.839\,$\bullet$\,0.983                   & 0.304\,$\bullet$\,0.361\,$\bullet$\,0.377 & 0.299\,$\bullet$\,0.357\,$\bullet$\,0.389                            & \textbf{0.297}\,$\bullet$\,\textbf{0.35}\,$\bullet$\,0.36            & 0.301\,$\bullet$\,0.355\,$\bullet$\,0.365                            \\
    Microzooplankton              & 0.406\,$\bullet$\,0.406\,$\bullet$\,\textbf{0.406}                   & 0.558\,$\bullet$\,0.991\,$\bullet$\,1.171                   & 0.354\,$\bullet$\,0.413\,$\bullet$\,0.428 & 0.348\,$\bullet$\,0.409\,$\bullet$\,0.429                            & \textbf{0.347}\,$\bullet$\,\textbf{0.403}\,$\bullet$\,0.414          & 0.352\,$\bullet$\,0.406\,$\bullet$\,0.415                            \\
    Diatoms                       & 0.435\,$\bullet$\,\textbf{0.435}\,$\bullet$\,\textbf{0.435}          & 0.609\,$\bullet$\,1.037\,$\bullet$\,1.18                    & 0.39\,$\bullet$\,0.438\,$\bullet$\,0.443  & \textbf{0.384}\,$\bullet$\,0.437\,$\bullet$\,0.452                   & \textbf{0.384}\,$\bullet$\,0.436\,$\bullet$\,0.442                   & 0.388\,$\bullet$\,0.437\,$\bullet$\,0.444                            \\
    Nanophytoplankton             & 0.438\,$\bullet$\,\textbf{0.438}\,$\bullet$\,\textbf{0.438}          & 0.656\,$\bullet$\,0.999\,$\bullet$\,1.131                   & 0.409\,$\bullet$\,0.441\,$\bullet$\,0.445 & 0.406\,$\bullet$\,0.446\,$\bullet$\,0.457                            & \textbf{0.404}\,$\bullet$\,0.439\,$\bullet$\,0.445                   & 0.408\,$\bullet$\,0.442\,$\bullet$\,0.448                            \\
    Chlorophyll                   & 0.415\,$\bullet$\,\textbf{0.415}\,$\bullet$\,\textbf{0.415}          & 0.573\,$\bullet$\,0.945\,$\bullet$\,1.087                   & 0.375\,$\bullet$\,0.417\,$\bullet$\,0.423 & \textbf{0.37}\,$\bullet$\,0.418\,$\bullet$\,0.435                    & \textbf{0.37}\,$\bullet$\,0.416\,$\bullet$\,0.422                    & 0.375\,$\bullet$\,0.418\,$\bullet$\,0.424                            \\
    Net primary production        & \textbf{0.455}\,$\bullet$\,\textbf{0.455}\,$\bullet$\,\textbf{0.455} & 0.557\,$\bullet$\,0.869\,$\bullet$\,0.983                   & 0.517\,$\bullet$\,0.521\,$\bullet$\,0.522 & 0.52\,$\bullet$\,0.531\,$\bullet$\,0.559                             & 0.52\,$\bullet$\,0.523\,$\bullet$\,0.525                             & 0.514\,$\bullet$\,0.505\,$\bullet$\,0.515                            \\
    \textit{Group mean}           & 0.418\,$\bullet$\,\textbf{0.418}\,$\bullet$\,\textbf{0.418}          & 0.568\,$\bullet$\,0.947\,$\bullet$\,1.089                   & 0.392\,$\bullet$\,0.432\,$\bullet$\,0.44  & 0.388\,$\bullet$\,0.433\,$\bullet$\,0.454                            & \textbf{0.387}\,$\bullet$\,0.428\,$\bullet$\,0.435                   & 0.39\,$\bullet$\,0.427\,$\bullet$\,0.435                             \\
    \midline
    \textbf{Carbon Pools}         &                                                                      &                                                             &                                           &                                                                      &                                                                      &                                                                      \\
    Dissolved inorganic carbon    & 0.174\,$\bullet$\,0.174\,$\bullet$\,0.174                            & 0.154\,$\bullet$\,0.301\,$\bullet$\,0.37                    & 0.116\,$\bullet$\,0.153\,$\bullet$\,0.169 & \textbf{0.107}\,$\bullet$\,\textbf{0.134}\,$\bullet$\,\textbf{0.142} & \textbf{0.107}\,$\bullet$\,0.135\,$\bullet$\,0.144                   & 0.114\,$\bullet$\,0.145\,$\bullet$\,0.154                            \\
    Alkalinity                    & 0.133\,$\bullet$\,0.133\,$\bullet$\,0.133                            & 0.124\,$\bullet$\,0.22\,$\bullet$\,0.265                    & 0.098\,$\bullet$\,0.123\,$\bullet$\,0.132 & \textbf{0.095}\,$\bullet$\,\textbf{0.117}\,$\bullet$\,0.127          & \textbf{0.095}\,$\bullet$\,\textbf{0.117}\,$\bullet$\,\textbf{0.123} & 0.098\,$\bullet$\,0.123\,$\bullet$\,0.129                            \\
    Calcium carbonate             & 0.42\,$\bullet$\,0.42\,$\bullet$\,0.42                               & 0.551\,$\bullet$\,0.856\,$\bullet$\,1.019                   & 0.379\,$\bullet$\,0.414\,$\bullet$\,0.418 & 0.375\,$\bullet$\,0.412\,$\bullet$\,0.419                            & \textbf{0.373}\,$\bullet$\,\textbf{0.41}\,$\bullet$\,\textbf{0.416}  & 0.379\,$\bullet$\,0.414\,$\bullet$\,0.418                            \\
    Particulate organic carbon    & 0.377\,$\bullet$\,0.377\,$\bullet$\,\textbf{0.377}                   & 0.488\,$\bullet$\,0.899\,$\bullet$\,1.093                   & 0.321\,$\bullet$\,0.381\,$\bullet$\,0.397 & 0.316\,$\bullet$\,0.375\,$\bullet$\,0.397                            & \textbf{0.315}\,$\bullet$\,\textbf{0.37}\,$\bullet$\,0.379           & 0.32\,$\bullet$\,0.374\,$\bullet$\,0.382                             \\
    Dissolved organic carbon      & 0.174\,$\bullet$\,0.174\,$\bullet$\,0.174                            & 0.162\,$\bullet$\,0.308\,$\bullet$\,0.376                   & 0.128\,$\bullet$\,0.161\,$\bullet$\,0.17  & \textbf{0.12}\,$\bullet$\,0.15\,$\bullet$\,0.164                     & \textbf{0.12}\,$\bullet$\,\textbf{0.149}\,$\bullet$\,\textbf{0.159}  & 0.128\,$\bullet$\,0.162\,$\bullet$\,0.172                            \\
    \textit{Group mean}           & 0.255\,$\bullet$\,0.255\,$\bullet$\,0.255                            & 0.296\,$\bullet$\,0.517\,$\bullet$\,0.625                   & 0.208\,$\bullet$\,0.247\,$\bullet$\,0.257 & 0.203\,$\bullet$\,0.238\,$\bullet$\,0.25                             & \textbf{0.202}\,$\bullet$\,\textbf{0.236}\,$\bullet$\,\textbf{0.244} & 0.208\,$\bullet$\,0.243\,$\bullet$\,0.251                            \\
    \midline
    \textbf{Physics}              &                                                                      &                                                             &                                           &                                                                      &                                                                      &                                                                      \\
    Temperature                   & 0.093\,$\bullet$\,0.093\,$\bullet$\,0.093                            & 0.106\,$\bullet$\,0.221\,$\bullet$\,0.283                   & ---\,$\bullet$\,---\,$\bullet$\,---       & ---\,$\bullet$\,---\,$\bullet$\,---                                  & ---\,$\bullet$\,---\,$\bullet$\,---                                  & \textbf{0.071}\,$\bullet$\,\textbf{0.088}\,$\bullet$\,\textbf{0.091} \\
    Salinity                      & 0.161\,$\bullet$\,0.161\,$\bullet$\,0.161                            & 0.13\,$\bullet$\,0.216\,$\bullet$\,0.254                    & ---\,$\bullet$\,---\,$\bullet$\,---       & ---\,$\bullet$\,---\,$\bullet$\,---                                  & ---\,$\bullet$\,---\,$\bullet$\,---                                  & \textbf{0.121}\,$\bullet$\,\textbf{0.147}\,$\bullet$\,\textbf{0.156} \\
    \textit{Group mean}           & 0.127\,$\bullet$\,0.127\,$\bullet$\,0.127                            & 0.118\,$\bullet$\,0.219\,$\bullet$\,0.269                   & ---\,$\bullet$\,---\,$\bullet$\,---       & ---\,$\bullet$\,---\,$\bullet$\,---                                  & ---\,$\bullet$\,---\,$\bullet$\,---                                  & \textbf{0.096}\,$\bullet$\,\textbf{0.118}\,$\bullet$\,\textbf{0.124} \\
    \botline
  \end{tabular}
  \caption{Per-variable NRMSE at lead times 1, 3, and 6 months ($x$\,$\bullet$\,$x$\,$\bullet$\,$x$), by group and model variant, averaged over 2020--2022. Physics variables evaluated only for climatology, persistence, and forecasted forcings. Best value at each lead time in bold: lower NRMSE.}\label{tab:forecast_nrmse_by_var}
\end{sidewaystable}

\begin{sidewaystable}
  \centering
  \begin{tabular}{lcccccc}
    \topline
                                  & \textbf{Climatology}                                                 & \textbf{Persistence}                                               & \shortstack{\textbf{Column-}                                                                                                                                                                                                                                                     \\\textbf{only}} & \shortstack{\textbf{+ Surface-}\\\textbf{forcing}} & \shortstack{\textbf{+ Horizontal-}\\\textbf{coupling}} & \shortstack{\textbf{+ Forecasted}\\\textbf{forcings}} \\
    \midline
    \textbf{Nutrients and oxygen} &                                                                      &                                                                    &                                                             &                                                                      &                                                                      &                                                                      \\
    Nitrate                       & 0.986\,$\bullet$\,0.986\,$\bullet$\,0.986                            & 0.988\,$\bullet$\,0.956\,$\bullet$\,0.934                          & 0.992\,$\bullet$\,0.988\,$\bullet$\,0.986                   & \textbf{0.993}\,$\bullet$\,\textbf{0.989}\,$\bullet$\,\textbf{0.988} & \textbf{0.993}\,$\bullet$\,\textbf{0.989}\,$\bullet$\,\textbf{0.988} & 0.992\,$\bullet$\,0.988\,$\bullet$\,0.987                            \\
    Phosphate                     & 0.985\,$\bullet$\,0.985\,$\bullet$\,0.985                            & 0.988\,$\bullet$\,0.957\,$\bullet$\,0.938                          & \textbf{0.992}\,$\bullet$\,0.986\,$\bullet$\,0.984          & \textbf{0.992}\,$\bullet$\,\textbf{0.988}\,$\bullet$\,\textbf{0.987} & \textbf{0.992}\,$\bullet$\,\textbf{0.988}\,$\bullet$\,\textbf{0.987} & \textbf{0.992}\,$\bullet$\,0.987\,$\bullet$\,0.986                   \\
    Silicate                      & 0.948\,$\bullet$\,0.948\,$\bullet$\,0.948                            & \textbf{0.993}\,$\bullet$\,\textbf{0.97}\,$\bullet$\,\textbf{0.95} & 0.956\,$\bullet$\,0.951\,$\bullet$\,0.949                   & 0.955\,$\bullet$\,0.951\,$\bullet$\,\textbf{0.95}                    & 0.955\,$\bullet$\,0.951\,$\bullet$\,0.949                            & 0.955\,$\bullet$\,0.95\,$\bullet$\,0.949                             \\
    Total dissolved iron          & 0.958\,$\bullet$\,0.958\,$\bullet$\,\textbf{0.958}                   & 0.944\,$\bullet$\,0.831\,$\bullet$\,0.759                          & \textbf{0.967}\,$\bullet$\,\textbf{0.959}\,$\bullet$\,0.957 & \textbf{0.967}\,$\bullet$\,\textbf{0.959}\,$\bullet$\,0.957          & \textbf{0.967}\,$\bullet$\,\textbf{0.959}\,$\bullet$\,\textbf{0.958} & \textbf{0.967}\,$\bullet$\,\textbf{0.959}\,$\bullet$\,\textbf{0.958} \\
    Oxygen                        & 0.988\,$\bullet$\,0.988\,$\bullet$\,0.988                            & 0.986\,$\bullet$\,0.943\,$\bullet$\,0.915                          & \textbf{0.993}\,$\bullet$\,0.989\,$\bullet$\,0.988          & \textbf{0.993}\,$\bullet$\,\textbf{0.99}\,$\bullet$\,\textbf{0.989}  & \textbf{0.993}\,$\bullet$\,\textbf{0.99}\,$\bullet$\,\textbf{0.989}  & \textbf{0.993}\,$\bullet$\,0.989\,$\bullet$\,0.988                   \\
    \textit{Group mean}           & 0.973\,$\bullet$\,0.973\,$\bullet$\,0.973                            & \textbf{0.98}\,$\bullet$\,0.932\,$\bullet$\,0.899                  & \textbf{0.98}\,$\bullet$\,0.974\,$\bullet$\,0.973           & \textbf{0.98}\,$\bullet$\,0.975\,$\bullet$\,\textbf{0.974}           & \textbf{0.98}\,$\bullet$\,\textbf{0.976}\,$\bullet$\,\textbf{0.974}  & \textbf{0.98}\,$\bullet$\,0.975\,$\bullet$\,\textbf{0.974}           \\
    \midline
    \textbf{Biology}              &                                                                      &                                                                    &                                                             &                                                                      &                                                                      &                                                                      \\
    Mesozooplankton               & 0.869\,$\bullet$\,0.869\,$\bullet$\,\textbf{0.869}                   & 0.79\,$\bullet$\,0.292\,$\bullet$\,0.03                            & 0.904\,$\bullet$\,0.864\,$\bullet$\,0.85                    & 0.907\,$\bullet$\,0.868\,$\bullet$\,0.841                            & \textbf{0.909}\,$\bullet$\,\textbf{0.874}\,$\bullet$\,0.866          & 0.906\,$\bullet$\,0.87\,$\bullet$\,0.862                             \\
    Microzooplankton              & 0.832\,$\bullet$\,0.832\,$\bullet$\,\textbf{0.832}                   & 0.688\,$\bullet$\,0.015\,$\bullet$\,$-$0.375                       & 0.871\,$\bullet$\,0.824\,$\bullet$\,0.807                   & 0.875\,$\bullet$\,0.827\,$\bullet$\,0.806                            & \textbf{0.876}\,$\bullet$\,\textbf{0.832}\,$\bullet$\,0.822          & 0.873\,$\bullet$\,0.831\,$\bullet$\,0.822                            \\
    Diatoms                       & 0.807\,$\bullet$\,\textbf{0.807}\,$\bullet$\,\textbf{0.807}          & 0.626\,$\bullet$\,$-$0.079\,$\bullet$\,$-$0.395                    & 0.846\,$\bullet$\,0.804\,$\bullet$\,0.799                   & \textbf{0.851}\,$\bullet$\,0.805\,$\bullet$\,0.791                   & \textbf{0.851}\,$\bullet$\,0.806\,$\bullet$\,0.8                     & 0.847\,$\bullet$\,0.805\,$\bullet$\,0.798                            \\
    Nanophytoplankton             & 0.806\,$\bullet$\,\textbf{0.806}\,$\bullet$\,\textbf{0.806}          & 0.553\,$\bullet$\,$-$0.015\,$\bullet$\,$-$0.302                    & 0.831\,$\bullet$\,0.804\,$\bullet$\,0.801                   & 0.833\,$\bullet$\,0.8\,$\bullet$\,0.79                               & \textbf{0.835}\,$\bullet$\,\textbf{0.806}\,$\bullet$\,0.801          & 0.831\,$\bullet$\,0.803\,$\bullet$\,0.798                            \\
    Chlorophyll                   & 0.827\,$\bullet$\,\textbf{0.827}\,$\bullet$\,\textbf{0.827}          & 0.67\,$\bullet$\,0.102\,$\bullet$\,$-$0.19                         & 0.859\,$\bullet$\,0.825\,$\bullet$\,0.818                   & 0.862\,$\bullet$\,0.823\,$\bullet$\,0.807                            & \textbf{0.863}\,$\bullet$\,0.826\,$\bullet$\,0.82                    & 0.858\,$\bullet$\,0.824\,$\bullet$\,0.818                            \\
    Net primary production        & \textbf{0.786}\,$\bullet$\,\textbf{0.786}\,$\bullet$\,\textbf{0.786} & 0.684\,$\bullet$\,0.222\,$\bullet$\,0.014                          & 0.59\,$\bullet$\,0.643\,$\bullet$\,0.655                    & 0.567\,$\bullet$\,0.613\,$\bullet$\,0.538                            & 0.56\,$\bullet$\,0.632\,$\bullet$\,0.641                             & 0.59\,$\bullet$\,0.684\,$\bullet$\,0.66                              \\
    \textit{Group mean}           & \textbf{0.821}\,$\bullet$\,\textbf{0.821}\,$\bullet$\,\textbf{0.821} & 0.669\,$\bullet$\,0.089\,$\bullet$\,$-$0.203                       & 0.817\,$\bullet$\,0.794\,$\bullet$\,0.788                   & 0.816\,$\bullet$\,0.789\,$\bullet$\,0.762                            & 0.815\,$\bullet$\,0.796\,$\bullet$\,0.791                            & 0.818\,$\bullet$\,0.803\,$\bullet$\,0.793                            \\
    \midline
    \textbf{Carbon Pools}         &                                                                      &                                                                    &                                                             &                                                                      &                                                                      &                                                                      \\
    Dissolved inorganic carbon    & 0.969\,$\bullet$\,0.969\,$\bullet$\,0.969                            & 0.976\,$\bullet$\,0.906\,$\bullet$\,0.856                          & 0.986\,$\bullet$\,0.976\,$\bullet$\,0.971                   & \textbf{0.988}\,$\bullet$\,\textbf{0.981}\,$\bullet$\,\textbf{0.979} & \textbf{0.988}\,$\bullet$\,\textbf{0.981}\,$\bullet$\,\textbf{0.979} & 0.987\,$\bullet$\,0.979\,$\bullet$\,0.976                            \\
    Alkalinity                    & 0.982\,$\bullet$\,0.982\,$\bullet$\,0.982                            & 0.984\,$\bullet$\,0.947\,$\bullet$\,0.921                          & 0.99\,$\bullet$\,0.985\,$\bullet$\,0.983                    & \textbf{0.991}\,$\bullet$\,\textbf{0.986}\,$\bullet$\,0.984          & \textbf{0.991}\,$\bullet$\,\textbf{0.986}\,$\bullet$\,\textbf{0.985} & 0.99\,$\bullet$\,0.985\,$\bullet$\,0.983                             \\
    Calcium carbonate             & 0.821\,$\bullet$\,0.821\,$\bullet$\,0.821                            & 0.685\,$\bullet$\,0.254\,$\bullet$\,$-$0.061                       & 0.853\,$\bullet$\,0.826\,$\bullet$\,0.823                   & 0.857\,$\bullet$\,0.827\,$\bullet$\,0.822                            & \textbf{0.858}\,$\bullet$\,\textbf{0.829}\,$\bullet$\,\textbf{0.824} & 0.853\,$\bullet$\,0.826\,$\bullet$\,0.823                            \\
    Particulate organic carbon    & 0.851\,$\bullet$\,0.851\,$\bullet$\,\textbf{0.851}                   & 0.758\,$\bullet$\,0.177\,$\bullet$\,$-$0.217                       & 0.89\,$\bullet$\,0.847\,$\bullet$\,0.831                    & \textbf{0.893}\,$\bullet$\,0.852\,$\bullet$\,0.831                   & \textbf{0.893}\,$\bullet$\,\textbf{0.856}\,$\bullet$\,0.849          & 0.891\,$\bullet$\,0.854\,$\bullet$\,0.847                            \\
    Dissolved organic carbon      & 0.967\,$\bullet$\,0.967\,$\bullet$\,0.967                            & 0.972\,$\bullet$\,0.902\,$\bullet$\,0.855                          & 0.981\,$\bullet$\,0.971\,$\bullet$\,0.968                   & \textbf{0.983}\,$\bullet$\,\textbf{0.974}\,$\bullet$\,0.969          & \textbf{0.983}\,$\bullet$\,\textbf{0.974}\,$\bullet$\,\textbf{0.971} & 0.981\,$\bullet$\,0.97\,$\bullet$\,0.966                             \\
    \textit{Group mean}           & 0.918\,$\bullet$\,0.918\,$\bullet$\,0.918                            & 0.875\,$\bullet$\,0.637\,$\bullet$\,0.471                          & 0.94\,$\bullet$\,0.921\,$\bullet$\,0.915                    & 0.942\,$\bullet$\,0.924\,$\bullet$\,0.917                            & \textbf{0.943}\,$\bullet$\,\textbf{0.926}\,$\bullet$\,\textbf{0.921} & 0.94\,$\bullet$\,0.923\,$\bullet$\,0.919                             \\
    \midline
    \textbf{Physics}              &                                                                      &                                                                    &                                                             &                                                                      &                                                                      &                                                                      \\
    Temperature                   & 0.991\,$\bullet$\,0.991\,$\bullet$\,\textbf{0.991}                   & 0.989\,$\bullet$\,0.949\,$\bullet$\,0.915                          & ---\,$\bullet$\,---\,$\bullet$\,---                         & ---\,$\bullet$\,---\,$\bullet$\,---                                  & ---\,$\bullet$\,---\,$\bullet$\,---                                  & \textbf{0.994}\,$\bullet$\,\textbf{0.992}\,$\bullet$\,\textbf{0.991} \\
    Salinity                      & 0.974\,$\bullet$\,0.974\,$\bullet$\,0.974                            & 0.982\,$\bullet$\,0.95\,$\bullet$\,0.93                            & ---\,$\bullet$\,---\,$\bullet$\,---                         & ---\,$\bullet$\,---\,$\bullet$\,---                                  & ---\,$\bullet$\,---\,$\bullet$\,---                                  & \textbf{0.985}\,$\bullet$\,\textbf{0.978}\,$\bullet$\,\textbf{0.975} \\
    \textit{Group mean}           & 0.982\,$\bullet$\,0.982\,$\bullet$\,0.982                            & 0.986\,$\bullet$\,0.950\,$\bullet$\,0.923                          & ---\,$\bullet$\,---\,$\bullet$\,---                         & ---\,$\bullet$\,---\,$\bullet$\,---                                  & ---\,$\bullet$\,---\,$\bullet$\,---                                  & \textbf{0.99}\,$\bullet$\,\textbf{0.985}\,$\bullet$\,\textbf{0.983}  \\
    \botline
  \end{tabular}
  \caption{Per-variable R$^2$ at lead times 1, 3, and 6 months ($x$\,$\bullet$\,$x$\,$\bullet$\,$x$), by group and model variant, averaged over 2020--2022. Physics variables evaluated only for climatology, persistence, and forecasted forcings. Best value at each lead time in bold: higher R$^2$.}\label{tab:forecast_r2_by_var}
\end{sidewaystable}

\begin{sidewaystable}
  \centering
  \begin{tabular}{lcccccc}
    \topline
                                  & \textbf{Climatology} & \textbf{Persistence}                                       & \shortstack{\textbf{Column-}                                                                                                                                                                                                                                           \\\textbf{only}} & \shortstack{\textbf{+ Surface-}\\\textbf{forcing}} & \shortstack{\textbf{+ Horizontal-}\\\textbf{coupling}} & \shortstack{\textbf{+ Forecasted}\\\textbf{forcings}} \\
    \midline
    \textbf{Nutrients and oxygen} &                      &                                                            &                                                   &                                                                      &                                                                      &                                                                      \\
    Nitrate                       & 0                    & \textbf{1.381}\,$\bullet$\,3.486\,$\bullet$\,5.28          & 0.421\,$\bullet$\,0.111\,$\bullet$\,0.03          & 0.431\,$\bullet$\,0.159\,$\bullet$\,\textbf{0.1}                     & 0.452\,$\bullet$\,\textbf{0.167}\,$\bullet$\,0.083                   & 0.406\,$\bullet$\,0.112\,$\bullet$\,0.036                            \\
    Phosphate                     & 0                    & \textbf{1.299}\,$\bullet$\,2.888\,$\bullet$\,4.181         & 0.399\,$\bullet$\,0.106\,$\bullet$\,0.031         & 0.412\,$\bullet$\,0.148\,$\bullet$\,0.08                             & 0.44\,$\bullet$\,\textbf{0.165}\,$\bullet$\,\textbf{0.081}           & 0.395\,$\bullet$\,0.11\,$\bullet$\,0.032                             \\
    Silicate                      & 0                    & \textbf{1.12}\,$\bullet$\,\textbf{1.575}\,$\bullet$\,1.968 & 0.111\,$\bullet$\,0.032\,$\bullet$\,0.008         & 0.117\,$\bullet$\,0.047\,$\bullet$\,\textbf{0.037}                   & 0.122\,$\bullet$\,0.043\,$\bullet$\,0.018                            & 0.111\,$\bullet$\,0.031\,$\bullet$\,0.01                             \\
    Total dissolved iron          & 0                    & \textbf{1.507}\,$\bullet$\,3.771\,$\bullet$\,5.516         & 0.21\,$\bullet$\,0.038\,$\bullet$\,0.011          & 0.218\,$\bullet$\,0.052\,$\bullet$\,\textbf{0.027}                   & 0.235\,$\bullet$\,\textbf{0.055}\,$\bullet$\,0.019                   & 0.22\,$\bullet$\,0.04\,$\bullet$\,0.011                              \\
    Oxygen                        & 0                    & 1.704\,$\bullet$\,5.864\,$\bullet$\,9.023                  & 0.372\,$\bullet$\,0.107\,$\bullet$\,0.032         & 0.429\,$\bullet$\,\textbf{0.213}\,$\bullet$\,\textbf{0.185}          & \textbf{0.435}\,$\bullet$\,0.199\,$\bullet$\,0.139                   & 0.38\,$\bullet$\,0.112\,$\bullet$\,0.037                             \\
    \textit{Group mean}           & 0                    & \textbf{1.402}\,$\bullet$\,3.517\,$\bullet$\,5.194         & 0.303\,$\bullet$\,0.079\,$\bullet$\,0.022         & 0.322\,$\bullet$\,0.124\,$\bullet$\,\textbf{0.086}                   & 0.337\,$\bullet$\,\textbf{0.126}\,$\bullet$\,0.068                   & 0.303\,$\bullet$\,0.081\,$\bullet$\,0.025                            \\
    \midline
    \textbf{Biology}              &                      &                                                            &                                                   &                                                                      &                                                                      &                                                                      \\
    Mesozooplankton               & 0                    & 1.857\,$\bullet$\,5.392\,$\bullet$\,7.55                   & 0.309\,$\bullet$\,0.063\,$\bullet$\,0.026         & 0.284\,$\bullet$\,0.081\,$\bullet$\,\textbf{0.072}                   & \textbf{0.321}\,$\bullet$\,\textbf{0.099}\,$\bullet$\,0.057          & 0.301\,$\bullet$\,0.075\,$\bullet$\,0.041                            \\
    Microzooplankton              & 0                    & 2.035\,$\bullet$\,5.671\,$\bullet$\,8.284                  & \textbf{0.314}\,$\bullet$\,0.058\,$\bullet$\,0.03 & 0.307\,$\bullet$\,\textbf{0.083}\,$\bullet$\,\textbf{0.055}          & \textbf{0.314}\,$\bullet$\,0.081\,$\bullet$\,0.045                   & 0.294\,$\bullet$\,0.056\,$\bullet$\,0.026                            \\
    Diatoms                       & 0                    & 2.058\,$\bullet$\,5.253\,$\bullet$\,6.964                  & \textbf{0.25}\,$\bullet$\,0.056\,$\bullet$\,0.026 & 0.228\,$\bullet$\,0.067\,$\bullet$\,\textbf{0.066}                   & 0.239\,$\bullet$\,\textbf{0.072}\,$\bullet$\,0.045                   & 0.226\,$\bullet$\,0.054\,$\bullet$\,0.036                            \\
    Nanophytoplankton             & 0                    & 1.939\,$\bullet$\,4.378\,$\bullet$\,5.78                   & 0.173\,$\bullet$\,0.039\,$\bullet$\,0.022         & \textbf{0.197}\,$\bullet$\,\textbf{0.067}\,$\bullet$\,\textbf{0.055} & 0.196\,$\bullet$\,0.064\,$\bullet$\,0.043                            & 0.168\,$\bullet$\,0.042\,$\bullet$\,0.025                            \\
    Chlorophyll                   & 0                    & 1.9\,$\bullet$\,4.647\,$\bullet$\,6.375                    & 0.207\,$\bullet$\,0.041\,$\bullet$\,0.02          & 0.208\,$\bullet$\,0.062\,$\bullet$\,\textbf{0.053}                   & \textbf{0.224}\,$\bullet$\,\textbf{0.067}\,$\bullet$\,0.039          & 0.203\,$\bullet$\,0.048\,$\bullet$\,0.028                            \\
    Net primary production        & 0                    & 1.465\,$\bullet$\,3.127\,$\bullet$\,4.187                  & 0.638\,$\bullet$\,0.259\,$\bullet$\,0.173         & 0.706\,$\bullet$\,\textbf{0.372}\,$\bullet$\,\textbf{0.397}          & \textbf{0.731}\,$\bullet$\,0.345\,$\bullet$\,0.284                   & 0.642\,$\bullet$\,0.23\,$\bullet$\,0.188                             \\
    \textit{Group mean}           & 0                    & 1.875\,$\bullet$\,4.745\,$\bullet$\,6.523                  & 0.315\,$\bullet$\,0.086\,$\bullet$\,0.049         & 0.322\,$\bullet$\,\textbf{0.122}\,$\bullet$\,\textbf{0.116}          & \textbf{0.338}\,$\bullet$\,0.121\,$\bullet$\,0.085                   & 0.306\,$\bullet$\,0.084\,$\bullet$\,0.057                            \\
    \midline
    \textbf{Carbon Pools}         &                      &                                                            &                                                   &                                                                      &                                                                      &                                                                      \\
    Dissolved inorganic carbon    & 0                    & 1.593\,$\bullet$\,4.239\,$\bullet$\,6.412                  & 0.446\,$\bullet$\,0.152\,$\bullet$\,0.06          & 0.465\,$\bullet$\,0.213\,$\bullet$\,0.13                             & \textbf{0.512}\,$\bullet$\,\textbf{0.255}\,$\bullet$\,\textbf{0.151} & 0.45\,$\bullet$\,0.169\,$\bullet$\,0.07                              \\
    Alkalinity                    & 0                    & \textbf{1.409}\,$\bullet$\,2.805\,$\bullet$\,3.847         & 0.404\,$\bullet$\,0.147\,$\bullet$\,0.055         & 0.447\,$\bullet$\,0.217\,$\bullet$\,\textbf{0.143}                   & 0.461\,$\bullet$\,\textbf{0.227}\,$\bullet$\,0.135                   & 0.425\,$\bullet$\,0.174\,$\bullet$\,0.081                            \\
    Calcium carbonate             & 0                    & \textbf{1.615}\,$\bullet$\,3.616\,$\bullet$\,5.31          & 0.179\,$\bullet$\,0.019\,$\bullet$\,0.003         & 0.164\,$\bullet$\,0.024\,$\bullet$\,0.009                            & 0.192\,$\bullet$\,\textbf{0.039}\,$\bullet$\,\textbf{0.014}          & 0.18\,$\bullet$\,0.028\,$\bullet$\,0.008                             \\
    Particulate organic carbon    & 0                    & 1.918\,$\bullet$\,5.558\,$\bullet$\,8.368                  & 0.319\,$\bullet$\,0.056\,$\bullet$\,0.019         & 0.316\,$\bullet$\,0.075\,$\bullet$\,\textbf{0.044}                   & \textbf{0.334}\,$\bullet$\,\textbf{0.086}\,$\bullet$\,0.038          & 0.317\,$\bullet$\,0.062\,$\bullet$\,0.024                            \\
    Dissolved organic carbon      & 0                    & \textbf{1.335}\,$\bullet$\,3.318\,$\bullet$\,4.974         & 0.416\,$\bullet$\,0.106\,$\bullet$\,0.028         & 0.464\,$\bullet$\,0.181\,$\bullet$\,0.07                             & 0.489\,$\bullet$\,\textbf{0.205}\,$\bullet$\,\textbf{0.095}          & 0.442\,$\bullet$\,0.128\,$\bullet$\,0.03                             \\
    \textit{Group mean}           & 0                    & \textbf{1.574}\,$\bullet$\,3.907\,$\bullet$\,5.782         & 0.353\,$\bullet$\,0.096\,$\bullet$\,0.033         & 0.371\,$\bullet$\,0.142\,$\bullet$\,0.079                            & 0.398\,$\bullet$\,\textbf{0.163}\,$\bullet$\,\textbf{0.087}          & 0.363\,$\bullet$\,0.112\,$\bullet$\,0.043                            \\
    \midline
    \textbf{Physics}              &                      &                                                            &                                                   &                                                                      &                                                                      &                                                                      \\
    Temperature                   & 0                    & 1.961\,$\bullet$\,8.162\,$\bullet$\,14.372                 & ---\,$\bullet$\,---\,$\bullet$\,---               & ---\,$\bullet$\,---\,$\bullet$\,---                                  & ---\,$\bullet$\,---\,$\bullet$\,---                                  & \textbf{0.553}\,$\bullet$\,\textbf{0.209}\,$\bullet$\,\textbf{0.063} \\
    Salinity                      & 0                    & \textbf{1.133}\,$\bullet$\,\textbf{1.77}\,$\bullet$\,2.307 & ---\,$\bullet$\,---\,$\bullet$\,---               & ---\,$\bullet$\,---\,$\bullet$\,---                                  & ---\,$\bullet$\,---\,$\bullet$\,---                                  & 0.48\,$\bullet$\,0.201\,$\bullet$\,\textbf{0.065}                    \\
    \textit{Group mean}           & 0                    & 1.547\,$\bullet$\,4.966\,$\bullet$\,8.34                   & ---\,$\bullet$\,---\,$\bullet$\,---               & ---\,$\bullet$\,---\,$\bullet$\,---                                  & ---\,$\bullet$\,---\,$\bullet$\,---                                  & \textbf{0.517}\,$\bullet$\,\textbf{0.205}\,$\bullet$\,\textbf{0.064} \\
    \botline
  \end{tabular}
  \caption{Per-variable anomaly variance ratio at lead times 1, 3, and 6 months ($x$\,$\bullet$\,$x$\,$\bullet$\,$x$), by group and model variant, averaged over 2020--2022. Physics variables evaluated only for climatology, persistence, and forecasted forcings. Best value at each lead time in bold: closest to 1.}\label{tab:forecast_varratio_by_var}
\end{sidewaystable}

\clearpage

\subsection*{Anomaly correlation coefficient (ACC) at 220m}

\begin{figure}[htbp]
  \centering
  \includegraphics[width=\textwidth]{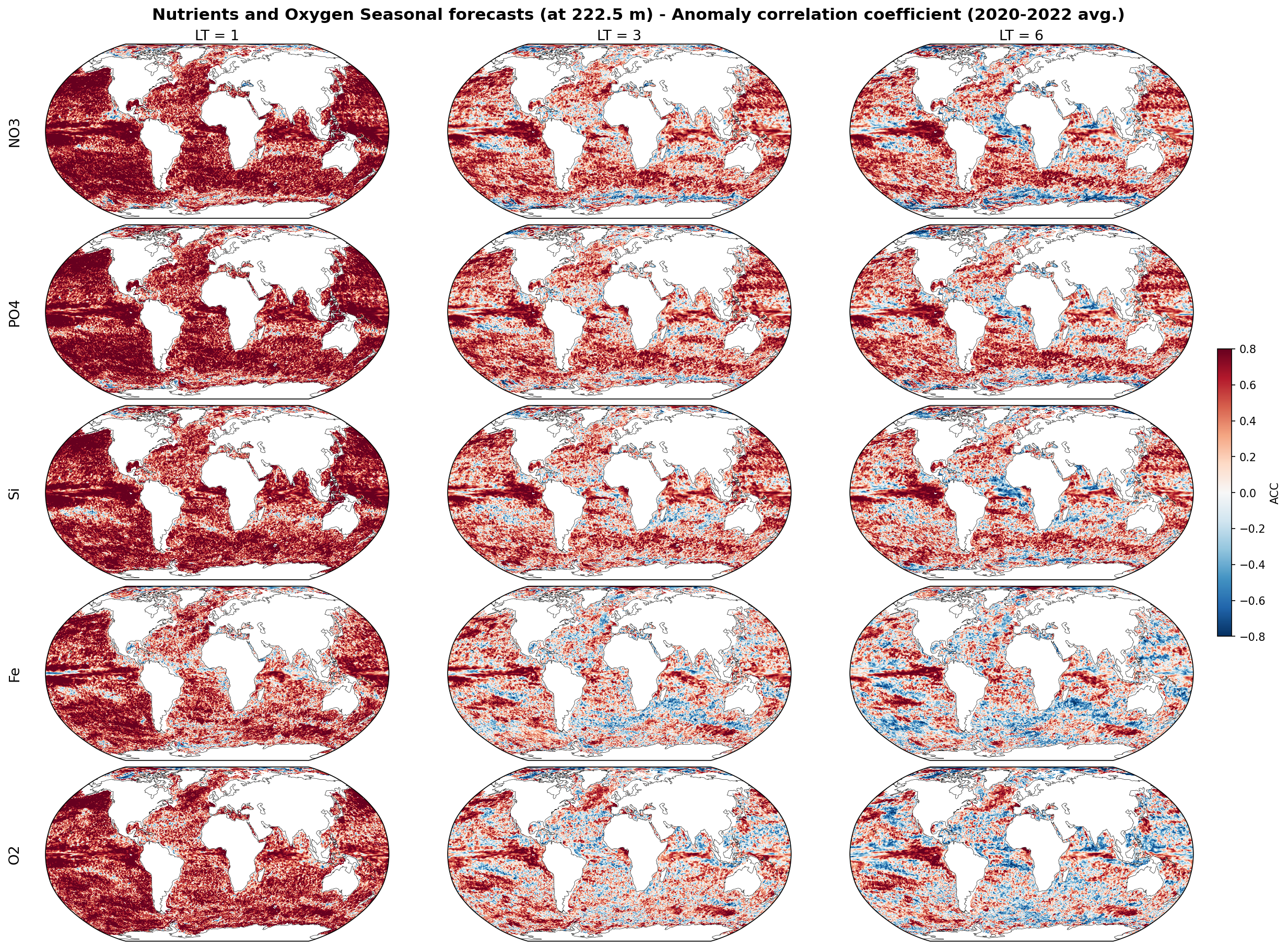}
  \caption{Anomaly correlation coefficient (ACC) for dissolved chemistry variables at 220\,m depth.}
  \label{fig:acc_chem220}
\end{figure}

\begin{figure}[htbp]
  \centering
  \includegraphics[width=\textwidth]{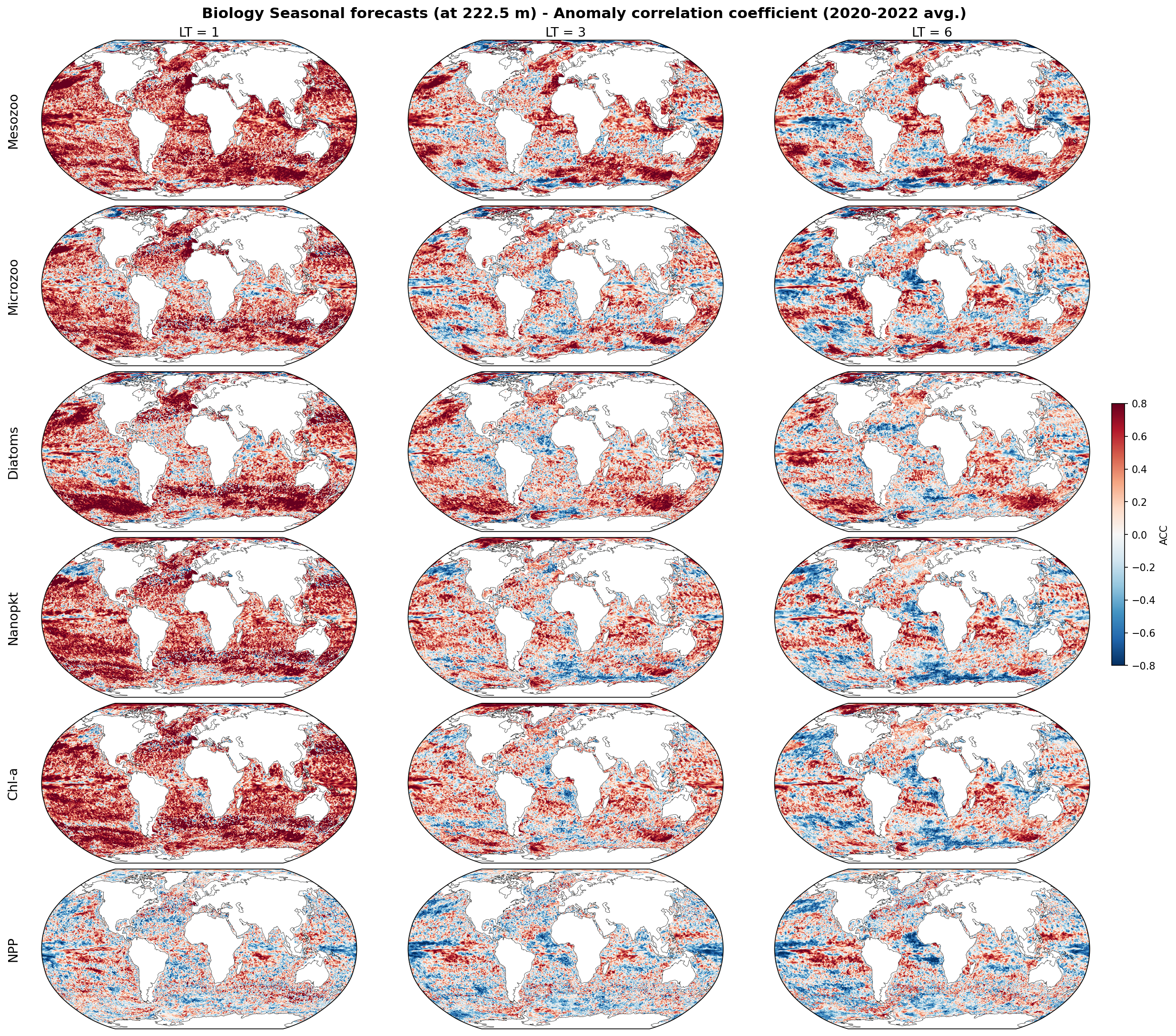}
  \caption{Anomaly correlation coefficient (ACC) for biology variables at 220\,m depth.}
  \label{fig:acc_biology220}
\end{figure}

\begin{figure}[htbp]
  \centering
  \includegraphics[width=\textwidth]{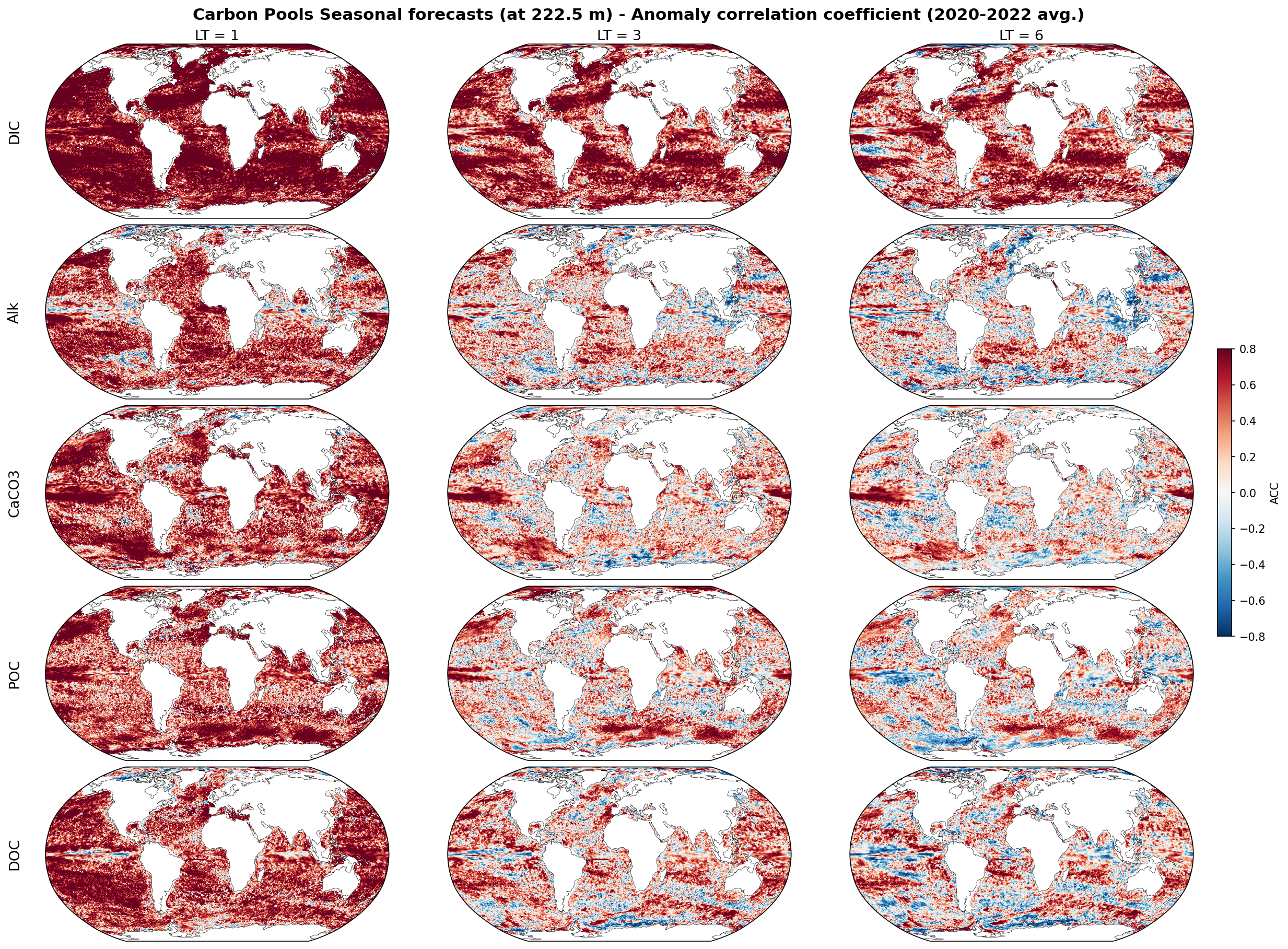}
  \caption{Anomaly correlation coefficient (ACC) for carbon variables at 220\,m depth.}
  \label{fig:acc_carbon220}
\end{figure}

\clearpage

\subsection*{Additional Hovmöller diagrams and forecasts samples}

\begin{figure}[htbp]
  \centering
  \includegraphics[width=\textwidth]{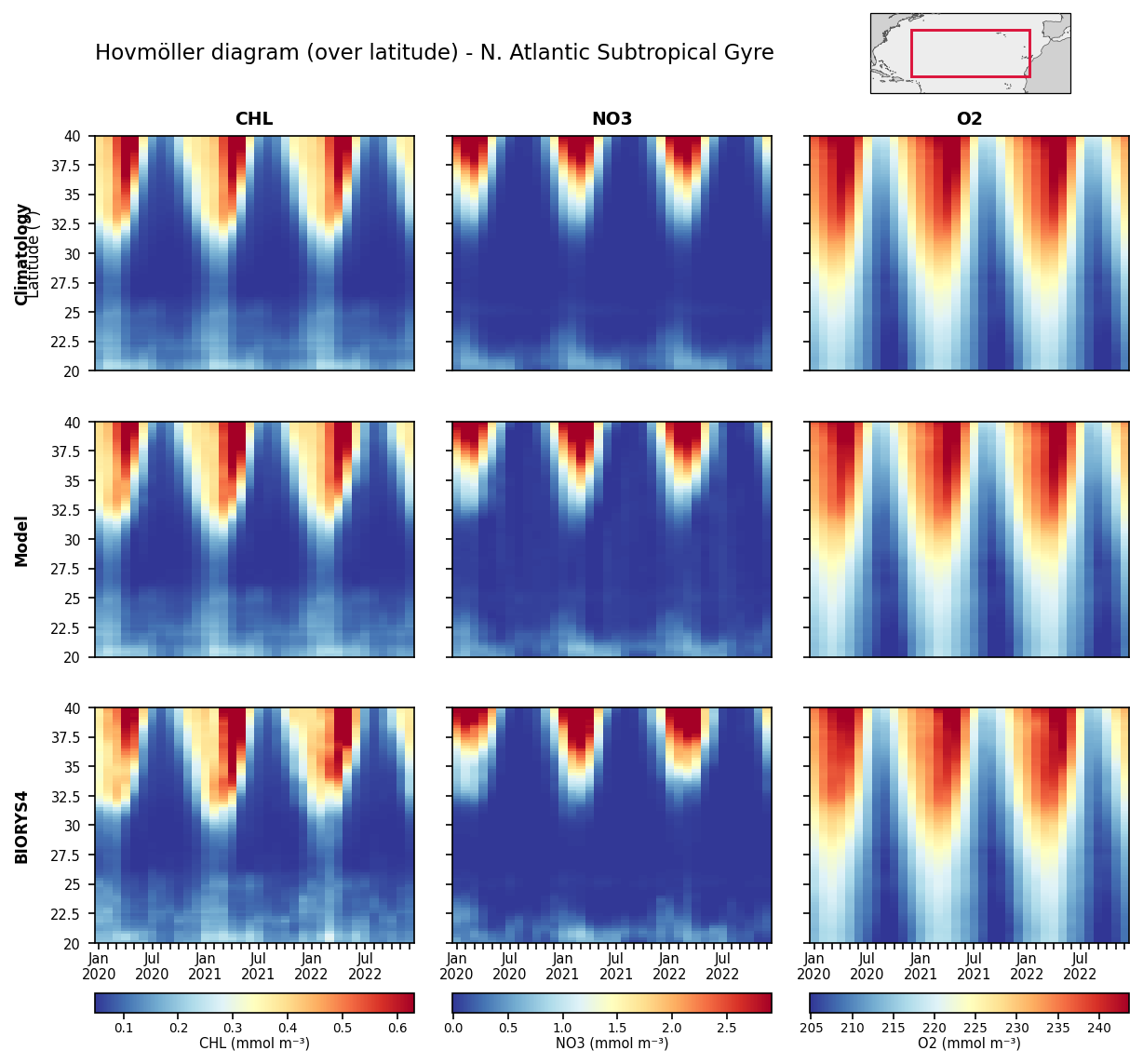}
  \caption{Hovmöller diagram (over latitude) of chlorophyll-a, nitrate and oxygen concentrations (at lead time one) in the North Atlantic Subtropical Gyre. The region is shown at the top of the figure, highlighted with a red box.}
  \label{fig:hovmoller_NASTG_lat}
\end{figure}

\begin{figure}[htbp]
  \centering
  \includegraphics[width=\textwidth]{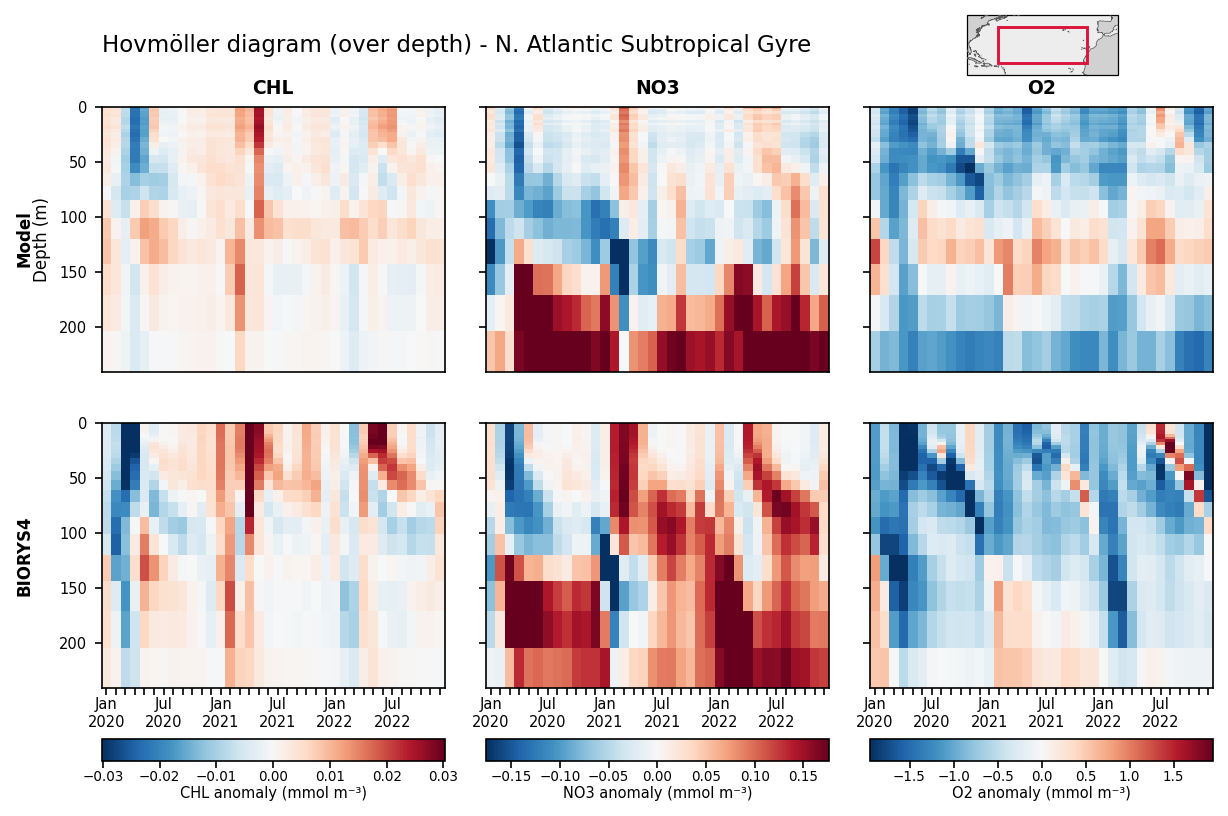}
  \caption{Hovmöller diagram (over depth) of chlorophyll-a, nitrate and oxygen anomalies (at lead time one) in the North Atlantic Subtropical Gyre.}
  \label{fig:hovmoller_NASTG_anom}
\end{figure}

\begin{figure}[htbp]
  \centering
  \includegraphics[width=\textwidth]{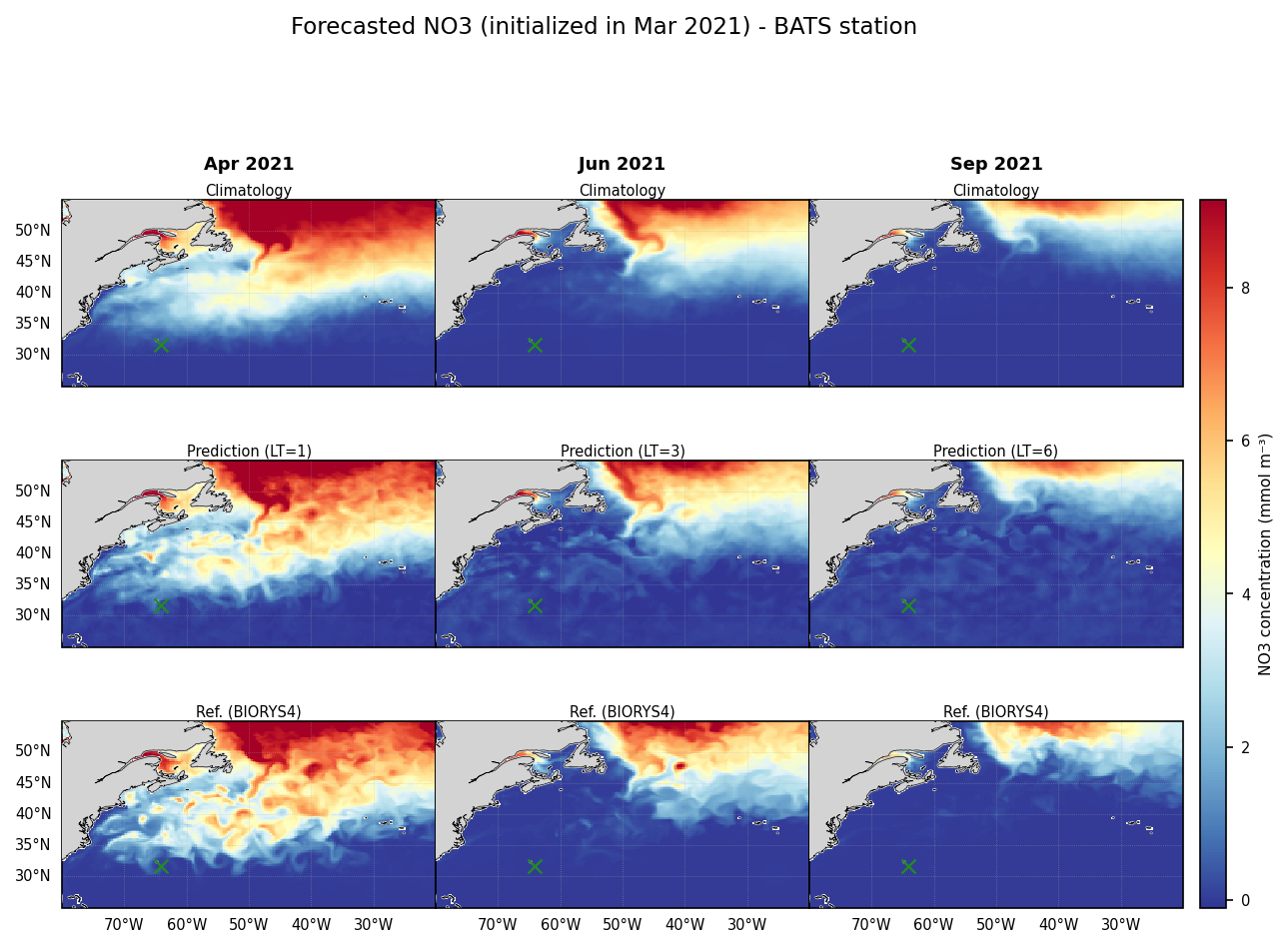}
  \caption{Forecasted nitrate (around BATS station)}
  \label{fig:BATS_no3}
\end{figure}

\begin{figure}[htbp]
  \centering
  \includegraphics[width=\textwidth]{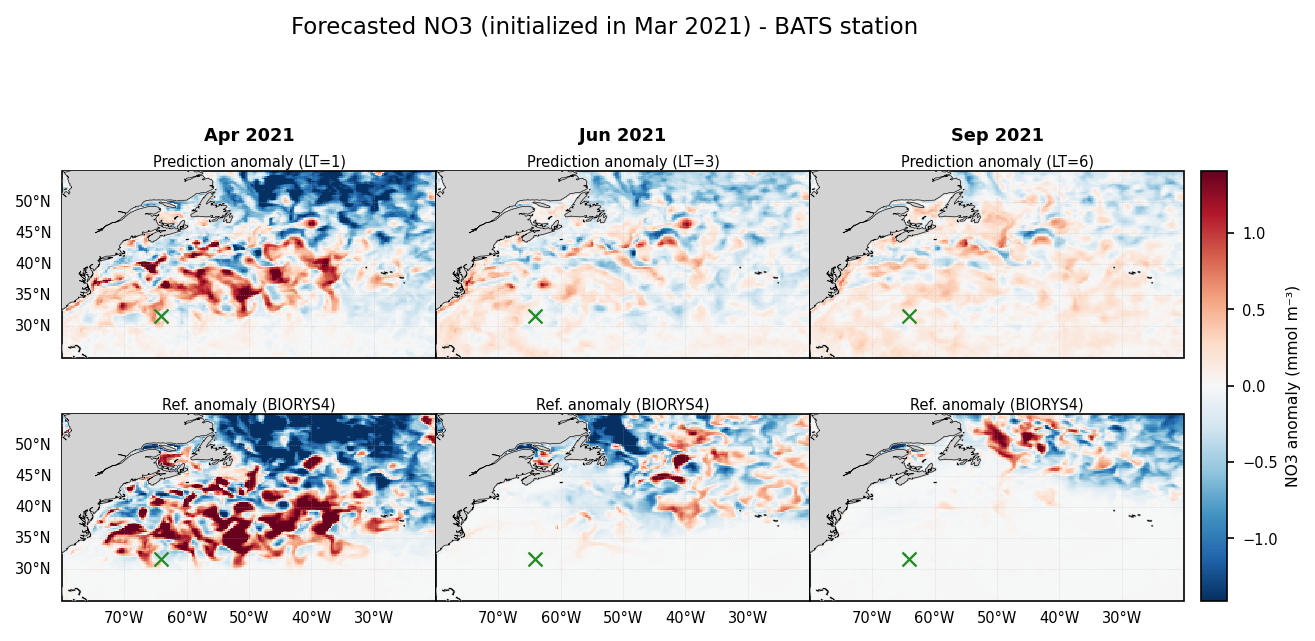}
  \caption{Forecasted nitrate anomalies (around BATS station)}
  \label{fig:BATS_no3_anom}
\end{figure}

\begin{figure}[htbp]
  \centering
  \includegraphics[width=\textwidth]{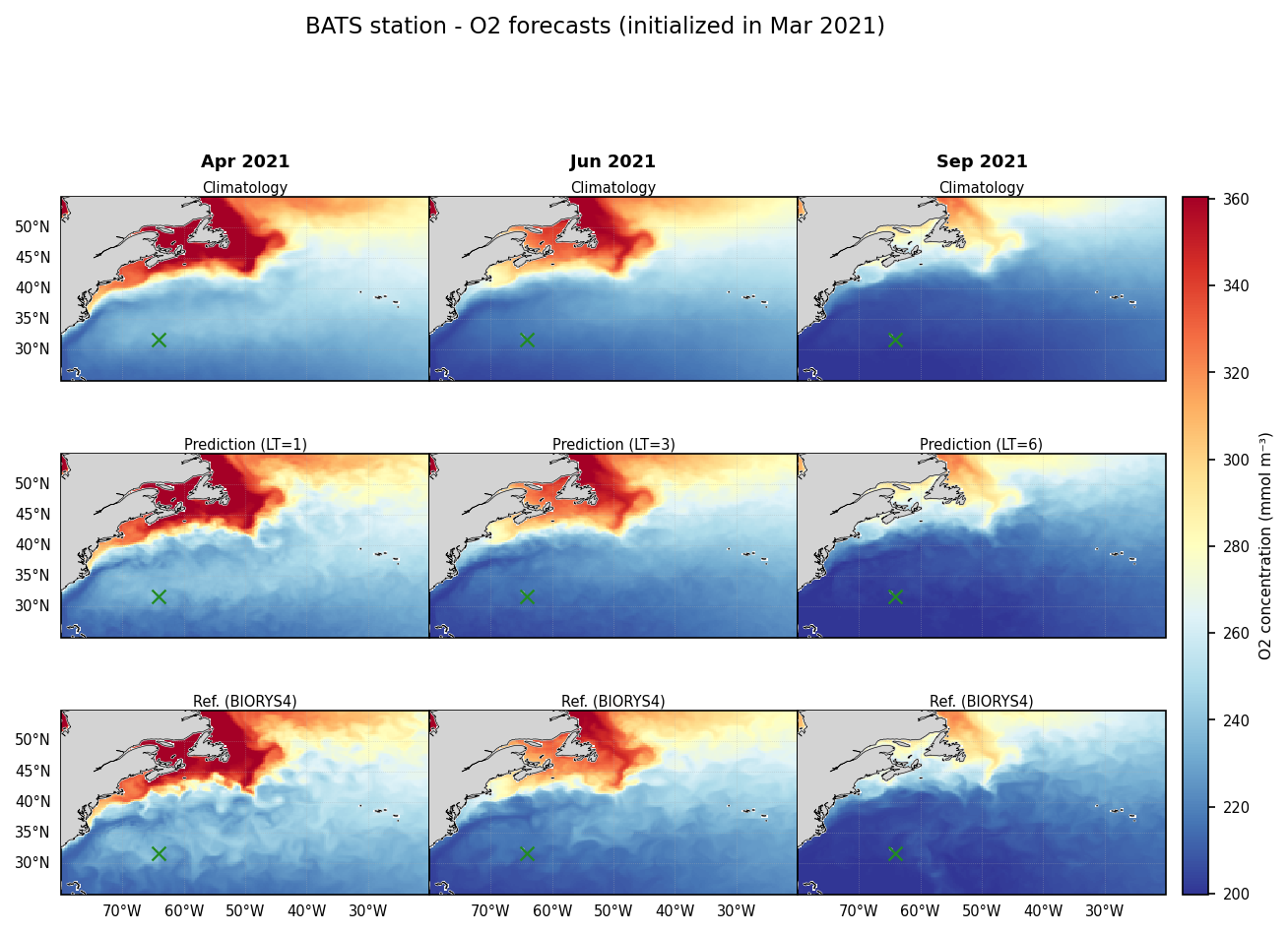}
  \caption{Forecasted oxygen (around BATS station)}
  \label{fig:BATS_o2}
\end{figure}

\begin{figure}[htbp]
  \centering
  \includegraphics[width=\textwidth]{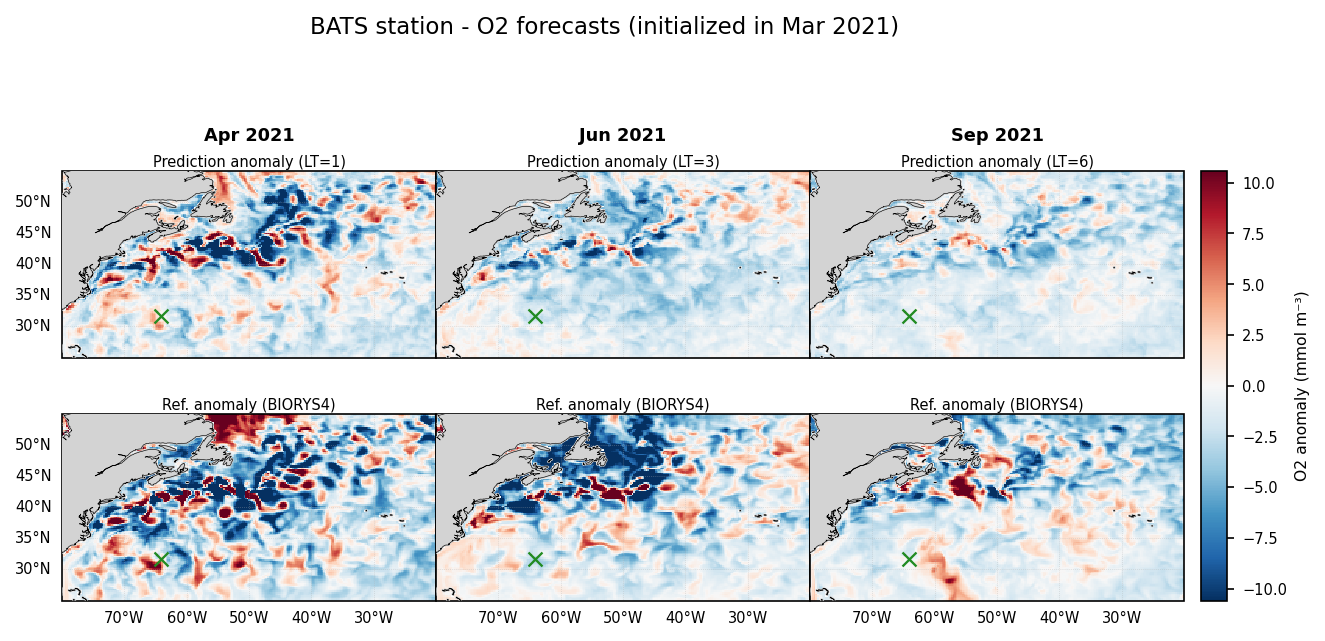}
  \caption{Forecasted oxygen anomalies (around BATS station)}
  \label{fig:BATS_o2_anom}
\end{figure}

\begin{figure}[htbp]
  \centering
  \includegraphics[width=\textwidth]{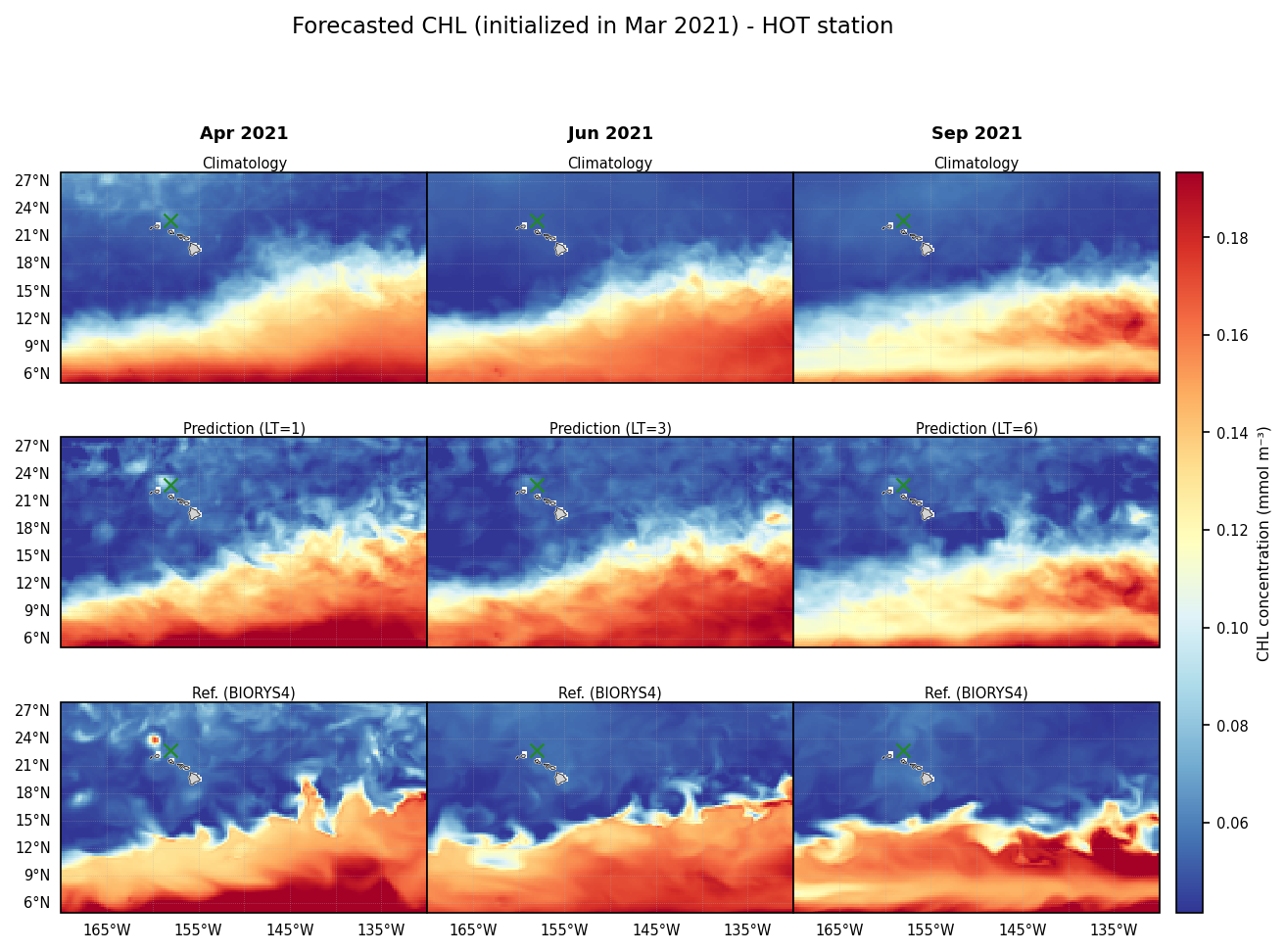}
  \caption{Forecasted chlorophyll-a (around HOT station)}
  \label{fig:HOT_chl}
\end{figure}

\begin{figure}[htbp]
  \centering
  \includegraphics[width=\textwidth]{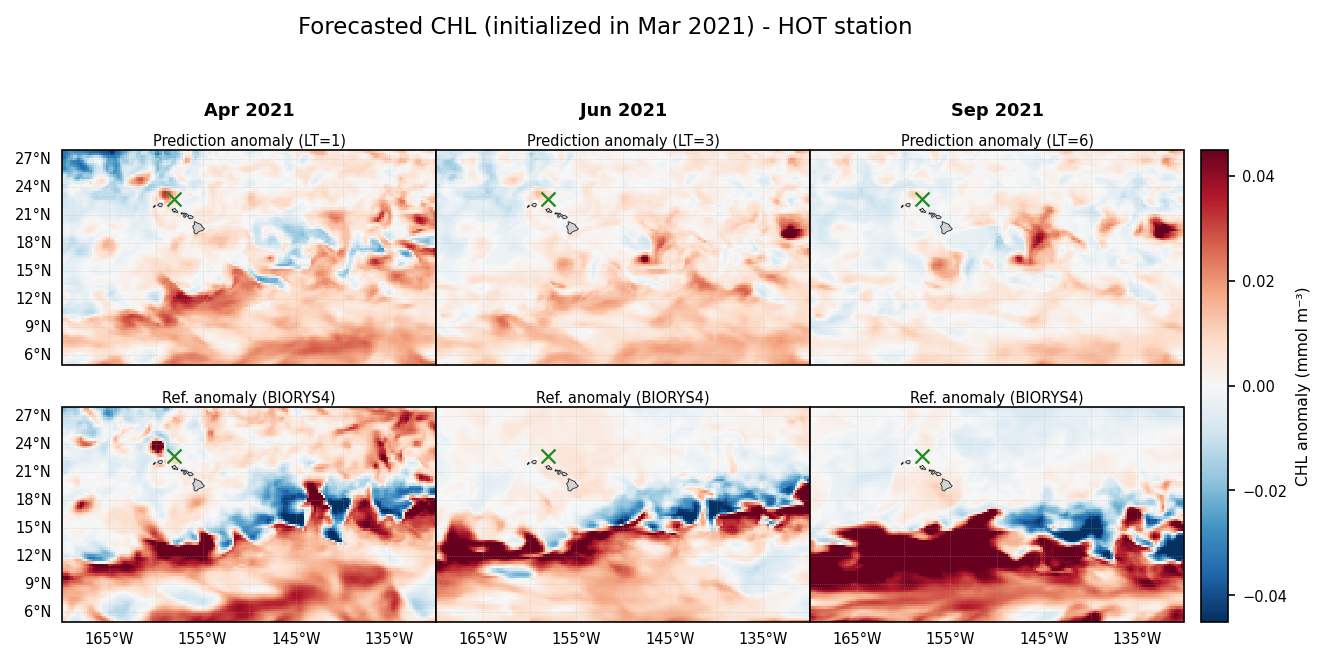}
  \caption{Forecasted chlorophyll-a anomalies (around HOT station)}
  \label{fig:HOT_chl_anom}
\end{figure}

\begin{figure}[htbp]
  \centering
  \includegraphics[width=\textwidth]{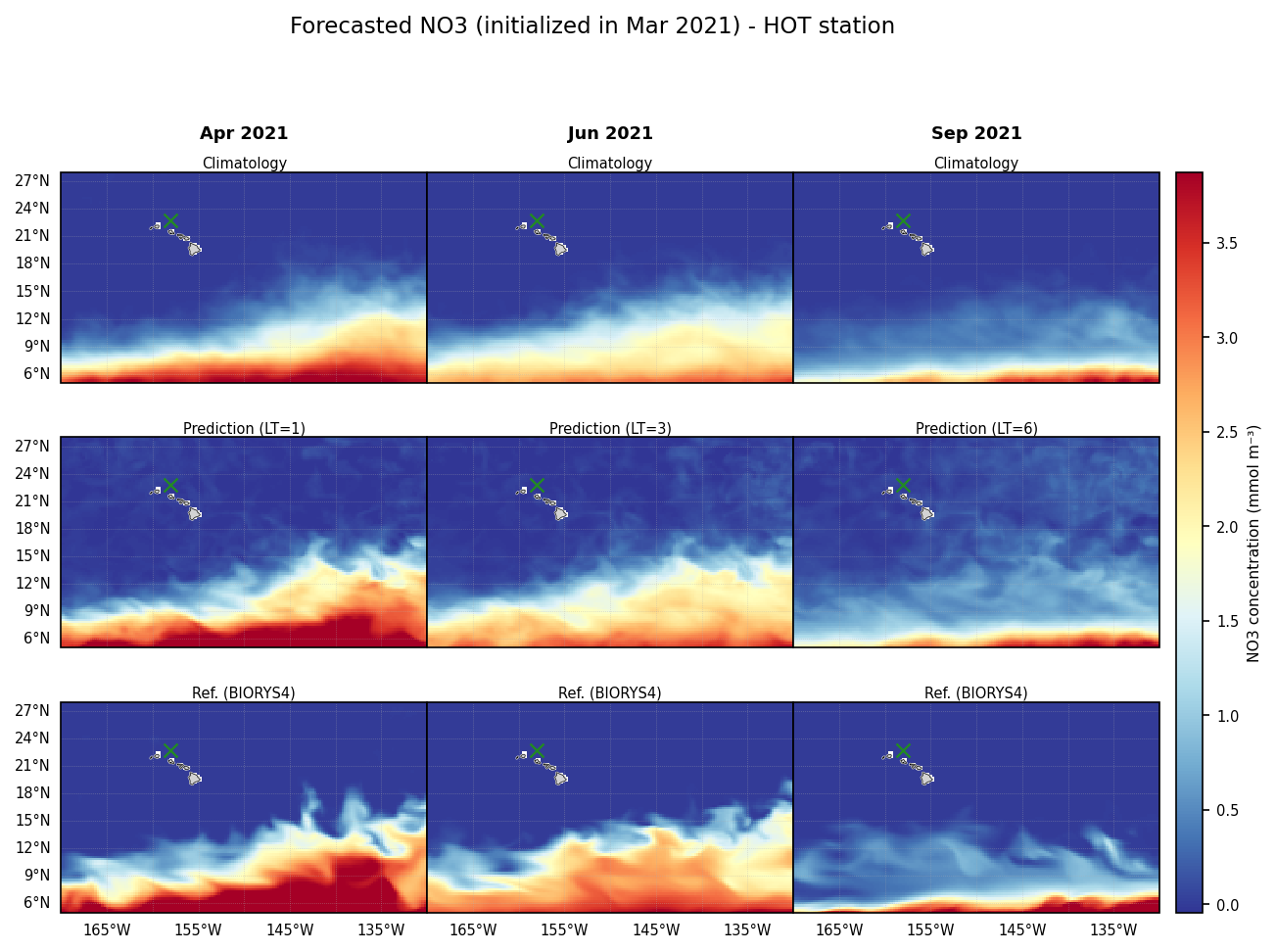}
  \caption{Forecasted nitrate (around HOT station)}
  \label{fig:HOT_no3}
\end{figure}

\begin{figure}[htbp]
  \centering
  \includegraphics[width=\textwidth]{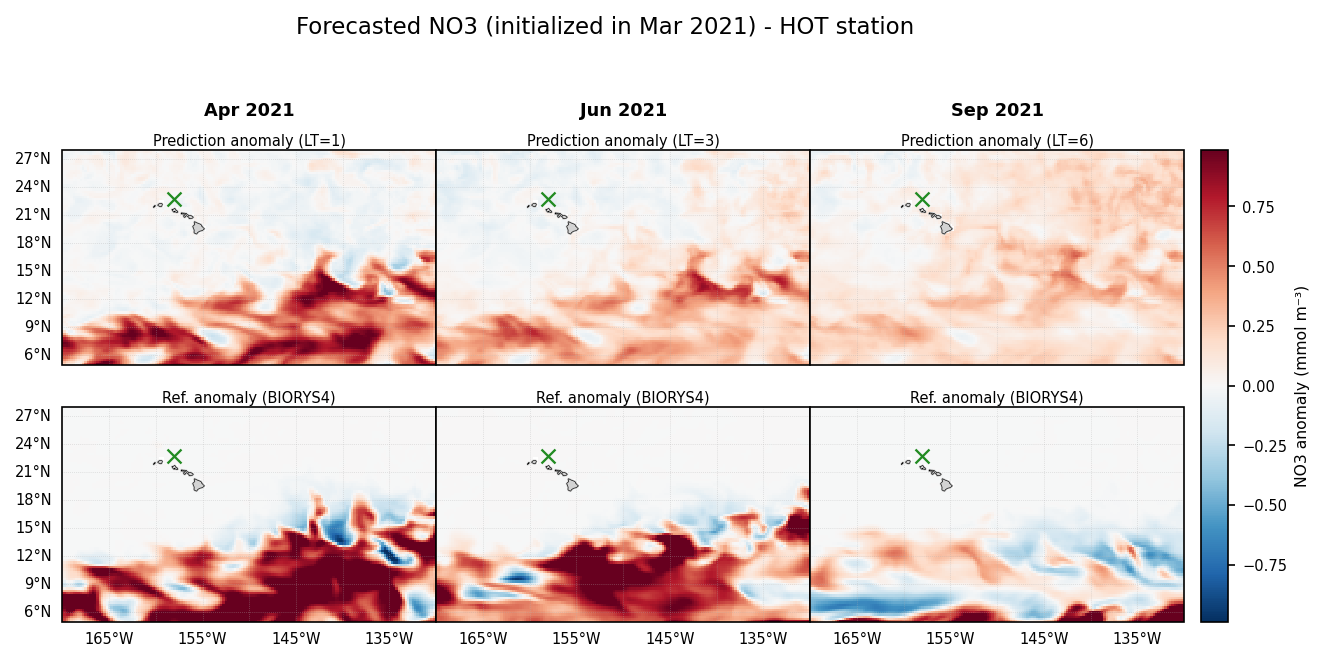}
  \caption{Forecasted nitrate anomalies (around HOT station)}
  \label{fig:HOT_no3_anom}
\end{figure}

\begin{figure}[htbp]
  \centering
  \includegraphics[width=\textwidth]{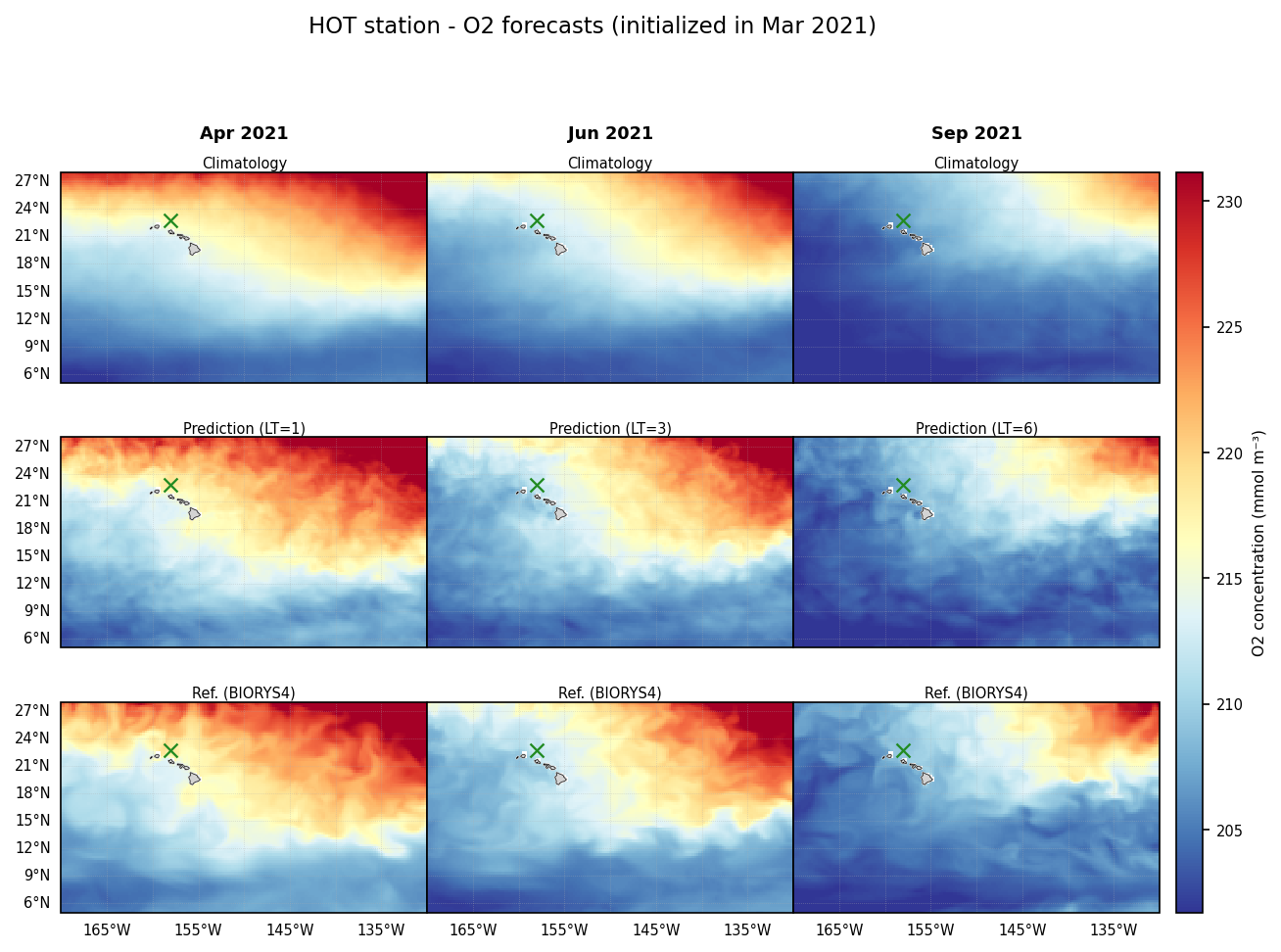}
  \caption{Forecasted oxygen (around HOT station)}
  \label{fig:HOT_o2}
\end{figure}

\begin{figure}[htbp]
  \centering
  \includegraphics[width=\textwidth]{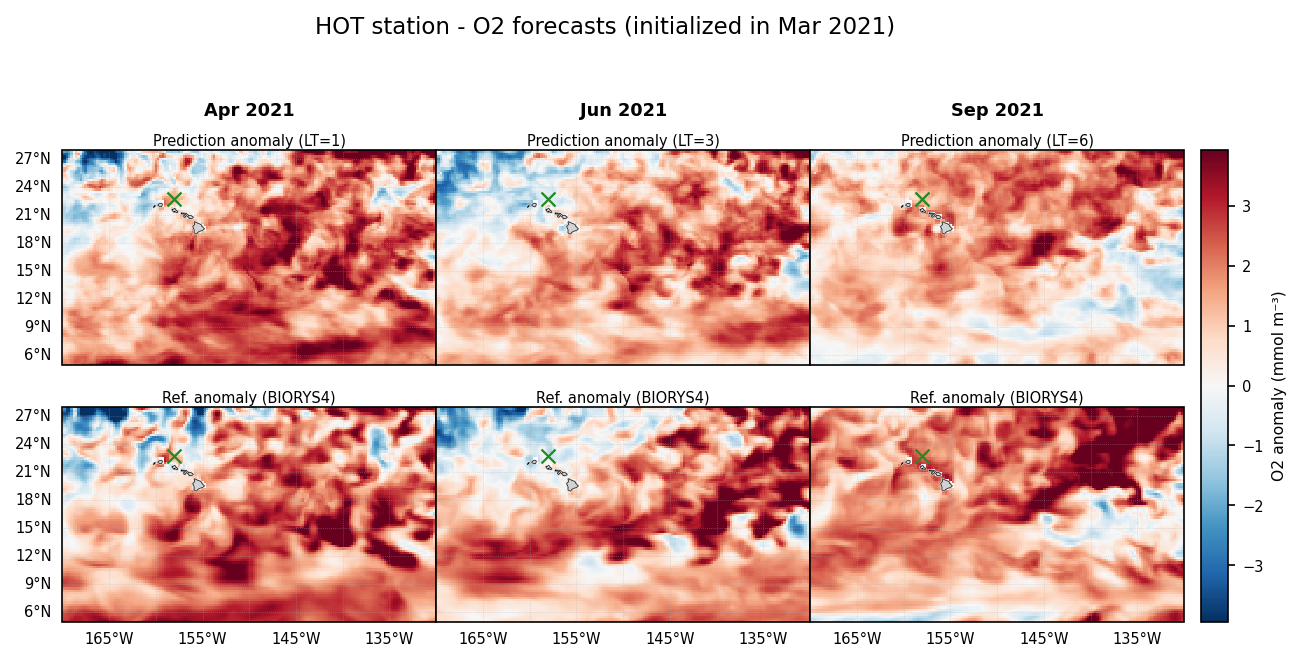}
  \caption{Forecasted oxygen anomalies (around HOT station)}
  \label{fig:HOT_o2_anom}
\end{figure}

\clearpage

\subsection*{Derived variables}

\paragraph*{Plankton and chlorophyll-a ratios}

BIORYS4 inherits two phytoplankton groups (diatoms and nanophytoplankton) and their corresponding chlorophyll-a tracers from PISCES. Since we also have tracers for total chlorophyll-a and phytoplankton biomass, we used these to calculate and verify the diatom and nanophytoplankton fractions of each total, averaged over the evaluation period. We calculate the global spatial mean of each field, then the component fractions. We also report the standard deviation of each field.

\begin{table}[h]
  \centering
  \begin{tabular}{lcc}
    \topline
                                                 & \textbf{BIORYS4 (reference)} & \textbf{Model-derived prediction} \\
    \midline
    Biomass (diatom / nano) mean fractions       & 0.3064 / 0.6936              & 0.3070 / 0.6930                   \\
    Chlorophyll-a (diatom / nano) mean fractions & 0.3629 / 0.6343              & 0.3608 / 0.6391                   \\
    Biomass standard deviation                   & 0.0178                       & 0.0021                            \\
    Chlorophyll-a standard deviation             & 0.0142                       & 0.0014                            \\
    \botline
  \end{tabular}
  \caption{Comparison of diatom and nanophytoplankton fractions between BIORYS4 and BG4Sea.}\label{tab:dn_ratios}
\end{table}

\paragraph*{Surface pressure of dissolved inorganic carbon (spCO$_2$)}
\mbox{}\\

\begin{table}[h]
  \centering
  \begin{tabular}{lccc}
    \topline
          & \textbf{Climatology}                                        & \textbf{Persistence}                                        & \textbf{Model-derived prediction}                  \\
    \midline
    ACC   & ---\,$\bullet$\,---\,$\bullet$\,---                         & \textbf{0.723}\,$\bullet$\,\textbf{0.372}\,$\bullet$\,0.275 & 0.521\,$\bullet$\,0.368\,$\bullet$\,\textbf{0.287} \\
    NRMSE & 0.487\,$\bullet$\,\textbf{0.487}\,$\bullet$\,\textbf{0.487} & \textbf{0.406}\,$\bullet$\,0.862\,$\bullet$\,1.095          & 0.481\,$\bullet$\,0.532\,$\bullet$\,0.557          \\
    R$^2$ & 0.763\,$\bullet$\,\textbf{0.763}\,$\bullet$\,\textbf{0.763} & \textbf{0.835}\,$\bullet$\,0.257\,$\bullet$\,$-$0.198       & 0.768\,$\bullet$\,0.717\,$\bullet$\,0.690          \\
    \botline
  \end{tabular}
  \caption{Forecast metrics for spCO$_2$ at lead times 1, 3, and 6 months ($x$\,$\bullet$\,$x$\,$\bullet$\,$x$), derived from prescribed physics, forecasted alkalinity and dissolved inorganic carbon.}\label{tab:spco2_metrics}
\end{table}

\subsection*{Physics forecasts}

\begin{figure}[htbp]
  \centering
  \includegraphics[width=\textwidth]{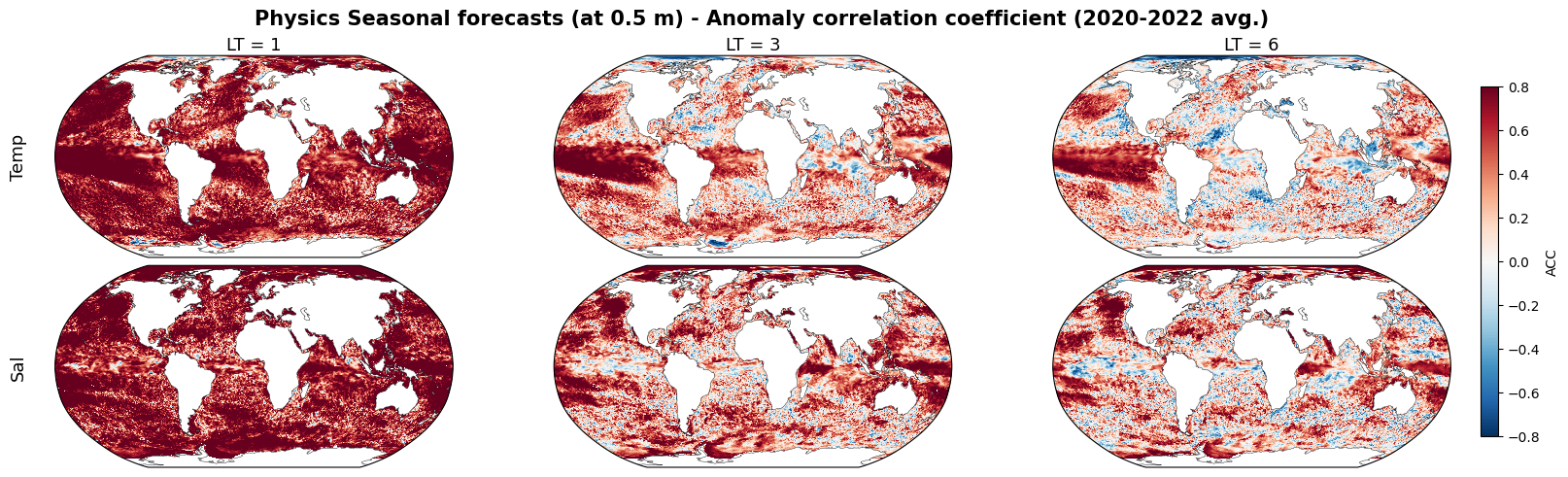}
  \caption{Anomaly correlation coefficient (ACC) for physical variables at the surface.}
  \label{fig:acc_physsurf}
\end{figure}

\begin{figure}[htbp]
  \centering
  \includegraphics[width=\textwidth]{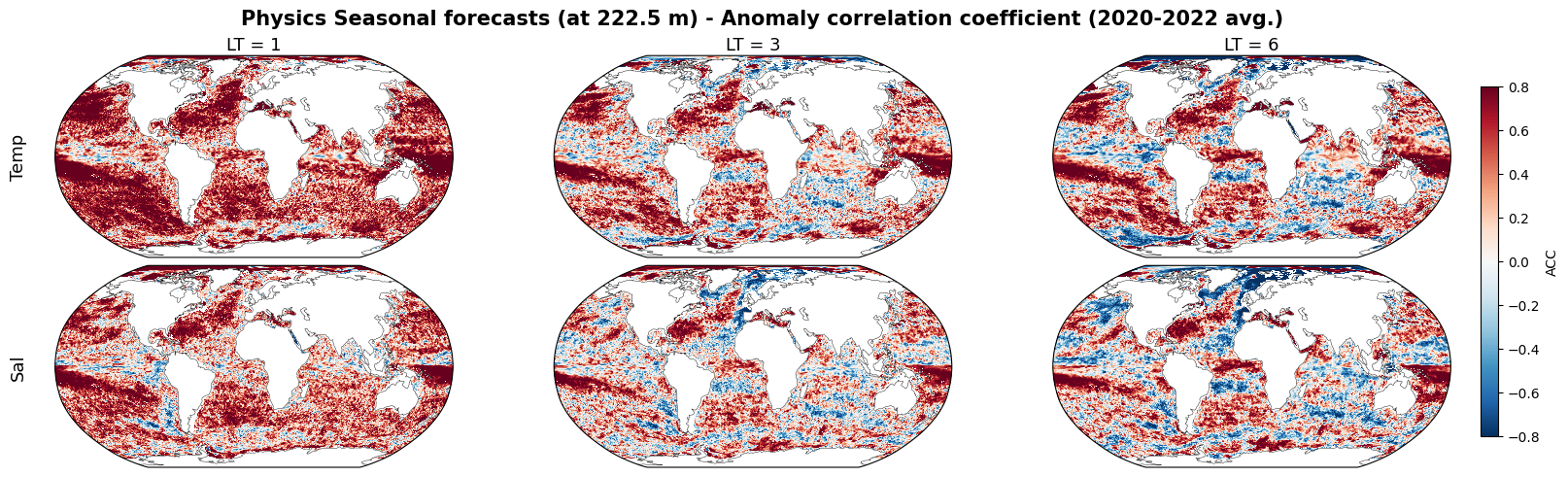}
  \caption{Anomaly correlation coefficient (ACC) for physical variables at 220m depth.}
  \label{fig:acc_phys220}
\end{figure}

\clearpage

\appendix[B]

\appendixtitle{Methodological details}

\subsection*{Data preprocessing and evaluation metrics}
\label{sec:supp_preprocessing_metrics}

\paragraph*{Monthly climatology and anomalies.}
For each variable and calendar month $m$, the climatology is
\begin{equation}
  \bar{\mathbf{k}}_m = \frac{1}{|\mathcal{T}_m|} \sum_{t \in \mathcal{T}_m} \mathbf{k}_t,
  \qquad \mathcal{T}_m = \{t : \mathrm{month}(t) = m\},
  \label{eq:clim}
\end{equation}
computed over the combined 2010--2019 window. We use both training and validation years to stabilise the monthly mean, as BIORYS4 does not extend prior to 2009 (which is regarded as a spin-up year). The anomaly is
\begin{equation}
  \mathbf{x}_t = \mathbf{k}_t - \bar{\mathbf{k}}_{m(t)}.
  \label{eq:anomaly}
\end{equation}

\paragraph*{Evaluation metrics.}
Let $\hat{\mathbf{k}}$ and $\mathbf{k}$ denote predicted and reference values in physical space, $\bar{\mathbf{k}}_m$ the monthly climatology for calendar month $m$, and $\tilde{w}_i = w_i / \sum_j w_j$ normalized area weights, with $w_i$ proportional to the cosine of latitude (so $\sum_i \tilde{w}_i = 1$). Spatial aggregates are computed as weighted sums over grid points $i$. The anomaly correlation coefficient (ACC) is
\begin{equation}
  \mathrm{ACC}
  = \frac{\sum_i \tilde{w}_i (\hat{k}_i - \bar{k}_m)(k_i - \bar{k}_m)}
  {\sqrt{\sum_i \tilde{w}_i (\hat{k}_i - \bar{k}_m)^2} \,
    \sqrt{\sum_i \tilde{w}_i (k_i - \bar{k}_m)^2}},
  \label{eq:acc}
\end{equation}
where $\bar{k}_m$ is evaluated at the calendar month of the target field. The coefficient of determination is
\begin{equation}
  R^2 = 1 - \frac{\sum_i \tilde{w}_i (\hat{k}_i - k_i)^2}{\sum_i \tilde{w}_i (k_i - \bar{k})^2},
  \label{eq:r2}
\end{equation}
with $\bar{k}$ the weighted mean of the reference field over the evaluation domain. Root mean squared error and normalised root mean squared error are
\begin{align}
  \mathrm{RMSE}  & = \sqrt{\sum_i \tilde{w}_i (\hat{k}_i - k_i)^2},
  \label{eq:rmse}                                                    \\
  \mathrm{NRMSE} & = \frac{\mathrm{RMSE}}{\sigma_k},
  \label{eq:nrmse}
\end{align}
where $\sigma_k = \sqrt{\sum_i \tilde{w}_i (k_i - \bar{k})^2}$ is the weighted standard deviation of the reference field over the same domain and period.

\subsection*{Base Autoencoder}
\label{sec:appendix_ae}

\paragraph*{Architecture}

The encoder applies $n_{conv}$ one-dimensional convolutional stages along the depth axis. Each stage increases the number of channels ($n_c$) by a fixed multiplicative factor $r$ while halving the depth dimension via max-pooling. Stages include residual connections. The resulting feature map is flattened and compressed to the latent dimension $L$ through a fully connected bottleneck MLP. The decoder is symmetric: the latent vector is expanded through a mirrored fully connected block, reshaped, and passed through $n_{conv}$ transposed-convolution and FiLM stages. The FiLM conditioning is constructed from the spatiotemporal embedding $\mathbf{e}_{D}$ as described below.

\begin{table}[h]
  \centering
  \begin{tabular}{ll}
    \topline
    \textbf{Component}                 & \textbf{Transformation, dimensionality}                                        \\
    \midline
    Month of year ($m$)                & $(\sin\frac{2\pi m}{12},\, \cos\frac{2\pi m}{12}), 2D$                         \\
    Latitude (degrees)                 & Linear map to $[-1,1], 1D$                                                     \\
    Longitude ($\lambda$ degrees)      & $(\sin\frac{2\pi\lambda}{360^\circ},\, \cos\frac{2\pi\lambda}{360^\circ}), 2D$ \\
    \midline
    \textit{Positional block:}         & \textit{Concatenation of above}, $5D$                                          \\
    \midline
    Depth profile (24 nominal depths), & Rescaled to $[-1,1]$, learned projection $(24 \to 16), 16D$                    \\
    \midline
    \textit{3D Positional block}       & \textit{Concatenation of positional block + depth}, $21D$                      \\
    \botline
  \end{tabular}
  \caption{Construction of the embedding $\mathbf{e}_{D}$. The lower half branches on whether surface conditioning is used.}
  \label{tab:ae_embedding}
\end{table}

Before being passed to the decoder, the combined embedding is projected onto a $L$-dimensional vector using a linear layer.

\paragraph*{Hyperparameter optimization (HPO) and training schedule}

Architecture, regularisation, and training hyperparameters of the column autoencoder were jointly tuned using Optuna~\citep{akiba2019optuna} with the Tree-structured Parzen Estimator (TPE) sampler and median pruning. We ran 50 startup trials to ensure adequate random exploration of the search space before model-based updates begin. The training objective was set to the mean squared error (MSE, Eq.~\ref{eq:ae_loss}). Trials used the same training configuration as the final training run. The selected architecture and training configuration is shown in Table~\ref{tab:hpo_space}. The final fixed training settings are listed in Table~\ref{tab:training_fixed}.

\begin{table}[h]
  \centering
  \begin{tabular}{p{0.50\linewidth}p{0.44\linewidth}}
    \topline
    \textbf{Hyperparameter}                    & \textbf{Candidates (selected in bold)}                                          \\
    \midline
    Encoder stages (depth-downsampling blocks) & 2,\ \textbf{3}                                                                  \\
    Base channels (first conv layer)           & 64,\ 128,\ \textbf{256}                                                         \\
    Channel multiplier (per stage)             & 1.5,\ 2.0,\ \textbf{2.5}                                                        \\
    Downsampling method                        & strided conv,\ \textbf{max pool}                                                \\
    Residual connections in encoder            & \textbf{yes},\ no                                                               \\
    Self-attention layer before bottleneck     & yes,\ \textbf{no}                                                               \\
    Bottleneck hidden layers                   & \textbf{3},\ 4,\ 5                                                              \\
    Bottleneck width (first FC layer)          & 256,\ \textbf{512},\ 1024                                                       \\
    Activation function                        & GELU,\ SiLU,\ Leaky ReLU,\ \textbf{ELU}                                         \\
    Normalisation                              & none,\ \textbf{LayerNorm},\ BatchNorm,\ Hybrid (BN$+$LN)                        \\
    Dropout                                    & \textbf{0.0},\ 0.05,\ 0.1,\ 0.2,\ 0.3                                           \\
    Weight decay                               & 0,\ $\mathbf{10^{-5}}$,\ $10^{-4}$,\ $10^{-3}$,\ $10^{-2}$                      \\
    Learning rate                              & log-uniform $[10^{-5},\ 5\times10^{-3}]$; \textbf{selected:} $2.5\times10^{-4}$ \\
    Optimiser                                  & \textbf{AdamW},\ Prodigy                                                        \\
    \botline
  \end{tabular}
  \caption{Autoencoder architecture and training hyperparameters. The HPO search space and selected values are shown; bold indicates the selected configuration.}
  \label{tab:hpo_space}
\end{table}

\begin{table}[h]
  \centering
  \begin{tabular}{ll}
    \topline
    \textbf{Setting}  & \textbf{Value}                                                                       \\
    \midline
    Optimiser         & AdamW                                                                                \\
    LR schedule       & Linear warm-up from $10^{-4}$ to $2.5 \times10^{-4}$ $\to$ constant during 10 epochs \\
    Batch size        & 260                                                                                  \\
    Maximum epochs    & 300                                                                                  \\
    Val loss patience & 30 validation checks                                                                 \\
    \botline
  \end{tabular}
  \caption{Autoencoder final run training settings.}
  \label{tab:training_fixed}
\end{table}

\subsection*{Latent Forecaster}
\label{sec:appendix_forecaster}

\paragraph*{Architecture}
The latent forecaster is a fully connected residual multilayer perceptron (MLP) operating in the $L$-dimensional latent space of the frozen autoencoder. It consists of $N_\mathrm{blocks} = 2$ residual blocks (fixed). Each block applies a linear transformation, normalization, and activation, with the initial hidden dimension $h_0$ scaled by a multiplier $m$. The model predicts the next latent vector by adding the output to the input (Eq.~\ref{eq:res_for}).

Before the residual MLP, the latent vector is augmented with a spatiotemporal embedding:
\begin{equation}
  \mathbf{z} \leftarrow \mathbf{z} + \mathbf{e}_{FC},
  \label{eq:latent_embed}
\end{equation}
where $\mathbf{e}_{FC}$ is the 5D positional block from Table~\ref{tab:ae_embedding} (month, latitude, and longitude only). The residual network $r_\delta$ therefore acts on $\tilde{\mathbf{z}}$; we suppress the tilde and refer to $\mathbf{z}$ instead for notational simplicity.

\paragraph*{HPO and training schedule}

The hyperparameters of the forecaster are tuned with the same Bayesian HPO framework used for the autoencoder. Only the forecaster weights are optimised during both HPO and training; the autoencoder is fully frozen. In order to avoid an additional learning rate search, the forecaster is trained with the Prodigy optimiser~\citep{mishchenko2024prodigy} (which adapts its own effective learning rate). The training objective was set to the same masked MSE as in Equation~\ref{eq:ae_loss}, and it is applied to the decoded prediction against the target anomaly field.

After initial evidence of overfitting, we introduced a penalty to the HPO objective to discourage those configurations. The adapted objective penalises end-of-trial degradation from the trial's best validation loss. Trials were run for longer than the autoencoder (60 epochs as opposed to 30) to further discourage overfitting. Otherwise, trials used the same training configuration as the final training run. The hyperparameter search space and selected values are given in Table~\ref{tab:hpo_latent} and the final training settings are listed in Table~\ref{tab:training_fixed2}.

\begin{table}[h]
  \centering
  \begin{tabular}{p{0.50\linewidth}p{0.44\linewidth}}
    \topline
    \textbf{Hyperparameter}         & \textbf{Candidates (selected in bold)}                     \\
    \midline
    Initial hidden dimension $h_0$  & 64,\ 128,\ \textbf{256}                                    \\
    Hidden dimension multiplier $m$ & 1.0,\ 1.5,\ \textbf{2.0}                                   \\
    Activation function             & GELU,\ SiLU,\ \textbf{Leaky ReLU},\ ELU                    \\
    Normalisation                   & \textbf{none},\ LayerNorm,\ BatchNorm                      \\
    Dropout                         & 0.0,\ 0.05,\ 0.1,\ 0.2,\ \textbf{0.3}                      \\
    Weight decay                    & \textbf{0},\ $10^{-5}$,\ $10^{-4}$,\ $10^{-3}$,\ $10^{-2}$ \\
    \botline
  \end{tabular}
  \caption{Latent forecaster hyperparameter search space. Bold indicates the selected configuration.}
  \label{tab:hpo_latent}
\end{table}

\begin{table}[h]
  \centering
  \begin{tabular}{ll}
    \topline
    \textbf{Setting}  & \textbf{Value}       \\
    \midline
    Optimiser         & Prodigy              \\
    Learning rate     & N/A                  \\
    Batch size        & 252                  \\
    Maximum epochs    & 300                  \\
    Val loss patience & 30 validation checks \\
    \botline
  \end{tabular}
  \caption{Latent forecaster final run training settings.}
  \label{tab:training_fixed2}
\end{table}

\subsection*{Surface-conditioning}
\label{sec:appendix_surface}

\paragraph*{Surface-conditioned autoencoder}

The surface-conditioned autoencoder uses the same convolutional architecture as the base AE, with identical hyperparameters (Table~\ref{tab:hpo_space}). The only modification is the FiLM conditioning: instead of the embedding-only $\mathbf{e}_{D} \in \mathbb{R}^{L}$, the decoder receives a 128-dimensional surface-aware vector $\mathbf{s}$. This is produced by a two-layer MLP ($\mathbf{s} = f_{\omega}(\mathbf{b},\, \mathbf{e}_{D}$)) which is applied to the concatenation of $\mathbf{e}_{D}$ and the standardised surface variables $\mathbf{b}$ (Table~\ref{tab:surface_cond}). $\mathbf{s}$ replaces $\mathbf{e}_{D}$ in the FiLM parameterisation at every decoder stage.

No separate HPO was conducted; the base AE configuration and training schedule were reused. A new instance of the base forecaster $r_\delta$ is trained with the surface-conditioned AE fully frozen. Architecture and hyperparameters are identical to Appendix~\ref{sec:appendix_forecaster} (Table~\ref{tab:hpo_latent}); training follows Table~\ref{tab:training_fixed2}.

\paragraph*{Surface-conditioned forecaster}

The surface-informed forecaster extends the previous architecture by concatenating a $d_s$-dimensional projection (where $d_s>L$) of the surface conditioning vector $\mathbf{s}$ with the latent state before the residual network, so the forecaster input dimension is $L + d_s$. All other hyperparameters are identical to the base architecture (Table~\ref{tab:hpo_latent}). Training follows Table~\ref{tab:training_fixed2}.

\paragraph*{Surface forecaster}

The surface forecaster $g_\epsilon$ is a residual MLP that takes the  $\mathbf{b}$ vector of physical variables as input to generate the prediction $\hat{\mathbf{b}}_{t} = \mathbf{b}_{t-1} + g_\epsilon(\mathbf{b}_{t-1})$ (advanced autoregressively at every lead time) before passing the prediction through the 128-dimensional surface embedding that conditions both the latent forecaster and the decoder.
The prediction $\hat{\mathbf{b}}_t$ is detached before being passed into the FiLM conditioning pathway, so $g_\epsilon$ receives gradients only from its own direct surface loss and is optimised independently of the column reconstruction loss:

\begin{equation}
  \mathcal{L}_{sf} = \frac{1}{N_{\text{train}}} \sum_{t, u} \left\| \mathbf{b}_{t, u} - \hat{\mathbf{b}}_{t, u} \right\|_F^2
  \label{eq:surf_loss}
\end{equation}

\begin{table}[h]
  \centering
  \begin{tabular}{ll}
    \topline
    \textbf{Hyperparameter}                     & \textbf{Value}       \\
    \midline
    Residual blocks $N_\mathrm{sf}$             & 2                    \\
    Initial hidden dimension $h_0^\mathrm{sf}$  & 16                   \\
    Hidden dimension multiplier $m^\mathrm{sf}$ & 2.0                  \\
    Positional embeddings                       & none                 \\
    Output dimension                            & $\in \mathbb{R}^{b}$ \\
    \botline
  \end{tabular}
  \caption{Surface forecaster ($g_\epsilon$) architecture. Width and depth were fixed; activation, normalisation, and dropout are inherited from the latent forecaster (Appendix~\ref{sec:appendix_forecaster}).}
  \label{tab:sf_forecaster}
\end{table}

\subsection*{Horizontal coupling}
\label{sec:appendix_horizontal}

\paragraph*{Architecture.}

We implement the horizontal context vector $\mathbf{n}^z_{t-1,u}$ with cross-attention over the latent neighbourhood ($n_h = 1$ head), with the center column providing a single query vector. Here, $\mathbf{z}_{t-1,j}$ are the latents of the $N=8$ columns surrounding the central $u$ in a $3\times3$ grid of immediate neighbours.

A query vector $ \mathbf{Q}_{t-1,u}$ is built from the center latent, while values $\mathbf{V}_{t-1,j}$ and keys $\mathbf{K}_{t-1,j}$ come from surrounding columns. A positional embedding $\mathbf{e}_p$ that includes the same coordinate encoding for latitude and longitude described in Table~\ref{tab:ae_embedding} is added to the surrounding latents before computing the keys.

The spatial context is thus given by:
\begin{equation}
  \mathbf{n}^z_{t-1,u}= \sum_{j \in \mathcal{P}_u} \alpha_{uj}\,\mathbf{V}_{t-1,j}, \qquad
  \alpha_{uj} = \mathrm{softmax}_{j}\!\left(\frac{\mathbf{Q}_{t-1,u}^{\top}\mathbf{K}_{t-1,j}}{\sqrt{d_h}}\right),
\end{equation}
where $d_h = L$ (single head) and $\mathcal{P}_u$ is the set of $N=8$ surrounding columns.

The surface context $\mathbf{n}^s_{t,u}$ is computed identically, with $\mathbf{s}_{t,u}$ as the query source and $\{\mathbf{s}_{t,j}\}_{j \in \mathcal{P}_u}$ as keys and values. Both $\mathbf{n}^z_{t-1,u}$ and $\mathbf{n}^s_{t,u}$ are projected to $d_s>L$ before concatenation with the latent state.

No HPO was conducted for this module. The neighbourhood size is fixed at $3\times3$ ($N=8$ neighbours). The cross-attention weights are trained jointly with the latent forecaster $r_{\delta sc}$; the autoencoder remains frozen throughout. Training follows the same schedule as the latent forecaster (Table~\ref{tab:training_fixed2}).

This shows how to enter the commands for making a bibliography using
BibTeX. It uses references.bib and the ametsocV6.bst file for the style.

\bibliographystyle{ametsocV6}
\bibliography{references}

@article{humphreysPyCO2SYSV182022,
  title   = {{PyCO2SYS} v1.8: marine carbonate system calculations in {Python}},
  volume  = {15},
  doi     = {10.5194/gmd-15-15-2022},
  number  = {1},
  journal = {Geoscientific Model Development},
  author  = {Humphreys, Matthew P. and Lewis, Ernie R. and Sharp, Jonathan D. and Pierrot, Denis},
  month   = jan,
  year    = {2022},
  pages   = {15--43}
}

@article{linkWhyWeNeed2023,
  title   = {Why we need weather forecast analogues for marine ecosystems},
  volume  = {80},
  doi     = {10.1093/icesjms/fsad143},
  number  = {8},
  journal = {ICES Journal of Marine Science},
  author  = {Link, J S and Thur, S and Matlock, G and Grasso, M},
  month   = oct,
  year    = {2023},
  pages   = {2087--2098}
}

@article{seferianTrackingImprovementSimulated2020,
  title   = {Tracking {Improvement} in {Simulated} {Marine} {Biogeochemistry} {Between} {CMIP5} and {CMIP6}},
  volume  = {6},
  doi     = {10.1007/s40641-020-00160-0},
  number  = {3},
  journal = {Current Climate Change Reports},
  author  = {Séférian, Roland and Berthet, Sarah and Yool, Andrew and Palmiéri, Julien and Bopp, Laurent and Tagliabue, Alessandro and Kwiatkowski, Lester and Aumont, Olivier and Christian, James and Dunne, John and Gehlen, Marion and Ilyina, Tatiana and John, Jasmin G. and Li, Hongmei and Long, Matthew C. and Luo, Jessica Y. and Nakano, Hideyuki and Romanou, Anastasia and Schwinger, Jörg and Stock, Charles and Santana-Falcón, Yeray and Takano, Yohei and Tjiputra, Jerry and Tsujino, Hiroyuki and Watanabe, Michio and Wu, Tongwen and Wu, Fanghua and Yamamoto, Akitomo},
  month   = aug,
  year    = {2020},
  pages   = {95--119}
}

@misc{perezFiLMVisualReasoning2017,
  title     = {{FiLM}: {Visual} {Reasoning} with a {General} {Conditioning} {Layer}},
  doi       = {10.48550/arXiv.1709.07871},
  publisher = {arXiv},
  author    = {Perez, Ethan and Strub, Florian and Vries, Harm de and Dumoulin, Vincent and Courville, Aaron},
  month     = dec,
  year      = {2017}
}

@misc{mishchenko2024prodigy,
  title  = {Prodigy: {An} {Expeditiously} {Adaptive} {Parameter}-{Free} {Learner}},
  author = {Mishchenko, Konstantin and Defazio, Aaron},
  year   = {2024},
  doi    = {10.48550/arXiv.2306.06101}
}

@inproceedings{akiba2019optuna,
  title     = {{Optuna}: {A} {Next-generation} {Hyperparameter} {Optimization} {Framework}},
  author    = {Akiba, Takuya and Sano, Shotaro and Yanase, Toshihiko and Ohta, Takeru and Koyama, Masanori},
  booktitle = {Proceedings of the 25th {ACM} {SIGKDD} International Conference on Knowledge Discovery and Data Mining},
  series    = {{KDD} '19},
  year      = {2019},
  location  = {Anchorage, {AK}, {USA}},
  publisher = {Association for Computing Machinery},
  address   = {New York, {NY}, {USA}},
  pages     = {2623--2631},
  doi       = {10.1145/3292500.3330701}
}

@article{gehlen_building_2015,
  title   = {Building the capacity for forecasting marine biogeochemistry and ecosystems: recent advances and future developments},
  volume  = {8},
  doi     = {10.1080/1755876X.2015.1022350},
  number  = {sup1},
  journal = {Journal of Operational Oceanography},
  author  = {Gehlen, M. and Barciela, R. and Bertino, L. and Brasseur, P. and Butenschön, M. and Chai, F. and Crise, A. and Drillet, Y. and Ford, D. and Lavoie, D. and Lehodey, P. and Perruche, C. and Samuelsen, A. and Simon, E.},
  month   = apr,
  year    = {2015},
  pages   = {s168--s187}
}

@article{fennel_ocean_2022,
  title   = {Ocean biogeochemical modelling},
  volume  = {2},
  doi     = {10.1038/s43586-022-00154-2},
  number  = {1},
  journal = {Nature Reviews Methods Primers},
  author  = {Fennel, Katja and Mattern, Jann Paul and Doney, Scott C. and Bopp, Laurent and Moore, Andrew M. and Wang, Bin and Yu, Liuqian},
  month   = sep,
  year    = {2022},
  pages   = {1--21}
}

@article{bopp_multiple_2013,
  title   = {Multiple stressors of ocean ecosystems in the 21st century: projections with {CMIP5} models},
  volume  = {10},
  doi     = {10.5194/bg-10-6225-2013},
  number  = {10},
  journal = {Biogeosciences},
  author  = {Bopp, L. and Resplandy, L. and Orr, J. C. and Doney, S. C. and Dunne, J. P. and Gehlen, M. and Halloran, P. and Heinze, C. and Ilyina, T. and Séférian, R. and Tjiputra, J. and Vichi, M.},
  month   = oct,
  year    = {2013},
  pages   = {6225--6245}
}

@article{park_seasonal_2019,
  title   = {Seasonal to multiannual marine ecosystem prediction with a global {Earth} system model},
  volume  = {365},
  doi     = {10.1126/science.aav6634},
  number  = {6450},
  journal = {Science},
  author  = {Park, Jong-Yeon and Stock, Charles A. and Dunne, John P. and Yang, Xiaosong and Rosati, Anthony},
  month   = jul,
  year    = {2019},
  pages   = {284--288}
}

@article{berardi_21st-century_2020,
  title   = {21st-century biogeochemical modeling: {Challenges} for {Century}-based models and where do we go from here?},
  volume  = {12},
  doi     = {10.1111/gcbb.12730},
  number  = {10},
  journal = {GCB Bioenergy},
  author  = {Berardi, Danielle},
  year    = {2020},
  pages   = {774--788}
}

@article{ismail_applications_2023,
  title   = {Applications of biogeochemical models in different marine environments: a review},
  volume  = {11},
  url     = {https://www.frontiersin.org/articles/10.3389/fenvs.2023.1198856},
  journal = {Frontiers in Environmental Science},
  author  = {Ismail, Kaltham A. and Al-Shehhi, Maryam R.},
  year    = {2023}
}

@article{aumont_pisces-v2_2015,
  title   = {{PISCES}-v2: an ocean biogeochemical model for carbon and ecosystem studies},
  volume  = {8},
  doi     = {10.5194/gmd-8-2465-2015},
  number  = {8},
  journal = {Geoscientific Model Development},
  author  = {Aumont, O. and Ethé, C. and Tagliabue, A. and Bopp, L. and Gehlen, M.},
  month   = aug,
  year    = {2015},
  pages   = {2465--2513}
}

@misc{pathak_fourcastnet_2022,
  title     = {{FourCastNet}: {A} {Global} {Data}-driven {High}-resolution {Weather} {Model} using {Adaptive} {Fourier} {Neural} {Operators}},
  doi       = {10.48550/arXiv.2202.11214},
  publisher = {arXiv},
  author    = {Pathak, Jaideep and Subramanian, Shashank and Harrington, Peter and Raja, Sanjeev and Chattopadhyay, Ashesh and Mardani, Morteza and Kurth, Thorsten and Hall, David and Li, Zongyi and Azizzadenesheli, Kamyar and Hassanzadeh, Pedram and Kashinath, Karthik and Anandkumar, Animashree},
  month     = feb,
  year      = {2022}
}

@article{lam_graphcast_2023,
  title   = {Learning skillful medium-range global weather forecasting},
  author  = {Lam, Remi and Sanchez-Gonzalez, Alvaro and Willson, Matthew and Wirnsberger, Peter and Fortunato, Meire and Pritzel, Alexander and Ravuri, Suman and Eaton-Rosen, Zach and Hu, Weihua and Merose, Alex and Hoyer, Stephan and Holland, George and Vinyals, Oriol and Stott, Jack and Paganini, Michela and Mohamed, Shakir and Boutall, Kathy and Bishop, Chris and Lattimer, Kevin and McKinlay, Scott and Kurth, Thorsten and Hall, David and Lane, Nathan and Martin, Chris and Suman, Suman and Li, Zongyi and Azizzadenesheli, Kamyar and Hassanzadeh, Pedram and Kashinath, Karthik and Anandkumar, Animashree},
  journal = {Science},
  volume  = {382},
  number  = {6670},
  pages   = {1416--1421},
  year    = {2023},
  doi     = {10.1126/science.adi2336}
}

@article{price_gencast_2025,
  title   = {Probabilistic weather forecasting with machine learning},
  author  = {Price, Ilan and Sanchez-Gonzalez, Alvaro and Alet, Ferran and Andersson, Tom R. and El-Kadi, Andrew and Masters, Dominic and Ewalds, Timo and Stott, Jacklynn and Mohamed, Shakir and Battaglia, Peter and Lam, Remi and Willson, Matthew},
  journal = {Nature},
  volume  = {637},
  pages   = {84--90},
  year    = {2025},
  doi     = {10.1038/s41586-024-08252-9}
}

@article{bi_accurate_2023,
  title   = {Accurate medium-range global weather forecasting with 3D neural networks},
  author  = {Bi, Kaifeng and Xie, Lingxi and Zhang, Hengheng and Chen, Xin and Gu, Xiaotao and Tian, Qi},
  journal = {Nature},
  volume  = {619},
  pages   = {533--538},
  year    = {2023},
  doi     = {10.1038/s41586-023-06185-3}
}

@article{claustre2020,
  title   = {The Scientific Rationale, Design and Implementation Plan for a Biogeochemical-Argo float array},
  author  = {Claustre, Hervé and Johnson, Kenneth S. and Takeshita, Yuichiro and Boss, Emmanuel and Briggs, Nathan and Argo Steering Team},
  journal = {Frontiers in Marine Science},
  volume  = {7},
  pages   = {546},
  year    = {2020},
  doi     = {10.3389/fmars.2020.00546}
}

@article{johnson_seas5_2019,
  title   = {{SEAS5}: the new {ECMWF} seasonal forecast system},
  volume  = {12},
  doi     = {10.5194/gmd-12-1087-2019},
  number  = {3},
  journal = {Geoscientific Model Development},
  author  = {Johnson, Stephanie J. and Stockdale, Timothy N. and Ferranti, Laura and Balmaseda, Magdalena A. and Molteni, Franco and Magnusson, Linus and Tietsche, Steffen and Decremer, Damien and Weisheimer, Antje and Balsamo, Gianpaolo and Keeley, Sarah P. E. and Mogensen, Kristian and Zuo, Hao and Monge-Sanz, Beatriz M.},
  month   = mar,
  year    = {2019},
  pages   = {1087--1117}
}

@article{lellouche_copernicus_2021,
  title   = {The {Copernicus} {Global} 1/12° {Oceanic} and {Sea} {Ice} {GLORYS12} {Reanalysis}},
  volume  = {9},
  doi     = {10.3389/feart.2021.698876},
  journal = {Frontiers in Earth Science},
  author  = {Lellouche, Jean-Michel and Eric, Greiner and Romain, Bourdallé-Badie and Gilles, Garric and Angélique, Melet and Marie, Drévillon and Clément, Bricaud and Mathieu, Hamon and Olivier, Le Galloudec and Charly, Regnier and Tony, Candela and Charles-Emmanuel, Testut and Florent, Gasparin and Giovanni, Ruggiero and Mounir, Benkiran and Yann, Drillet and Pierre-Yves, Le Traon},
  month   = jul,
  year    = {2021}
}

@misc{cmems_bio,
  address   = {Marine Data Store (MDS)},
  title     = {Global {Ocean} {Biogeochemistry} {Analysis} and {Forecast}},
  doi       = {10.48670/moi-00015},
  publisher = {E.U. Copernicus Marine Service Information (CMEMS)},
  author    = {CMEMS},
  year      = {2023}
}

@misc{cmems_phys,
  address   = {Marine Data Store (MDS)},
  title     = {Global {Ocean} {Physics} {Reanalysis}},
  doi       = {10.48670/moi-00021},
  publisher = {E.U. Copernicus Marine Service Information (CMEMS)},
  author    = {Copernicus Marine Environment Monitoring Service (CMEMS)},
  year      = {2021}
}

@misc{gregoire_review_2026,
  title  = {Review and {Synthesis}: operational prediction of ocean biogeochemistry in the {Copernicus Marine Service}. {Main} achievements and perspectives},
  author = {Gr\'egoire, Marilaure and Cossarini, Gianpiero and Derval, Corinne and Gutknecht, Elodie and Kay, Susan and Lamouroux, Julien and Morrison, Helen and Perruche, Coralie and Samuelsen, Annette and Spruch, Lena and Teruzzi, Anna and Vandenbulcke, Luc and {von Schuckmann}, Karina and Wakamatsu, Tsuyoshi},
  year   = {2026},
  note   = {Preprint, EGUsphere},
  doi    = {10.5194/egusphere-2026-813},
  url    = {https://egusphere.copernicus.org/preprints/2026/egusphere-2026-813/}
}

@article{eppleyParticulateOrganicMatter1979,
  title   = {Particulate Organic Matter Flux and Planktonic New Production in the Deep Ocean},
  volume  = {282},
  number  = {5740},
  journal = {Nature},
  author  = {Eppley, Richard W. and Peterson, Bruce J.},
  year    = {1979},
  pages   = {677--680},
  doi     = {10.1038/282677a0}
}

@article{martinGlacialInterglacialCO21990,
  title   = {Glacial-Interglacial {CO$_2$} Change: The Iron Hypothesis},
  volume  = {5},
  number  = {1},
  journal = {Paleoceanography},
  author  = {Martin, John H.},
  year    = {1990},
  pages   = {1--13},
  doi     = {10.1029/PA005i001p00001}
}

@article{zehrNitrogenCyclingOcean2002,
  title   = {Nitrogen Cycling in the Ocean: New Perspectives on Processes and Paradigms},
  volume  = {68},
  number  = {3},
  journal = {Applied and Environmental Microbiology},
  author  = {Zehr, Jonathan P. and Ward, Bess B.},
  year    = {2002},
  pages   = {1015--1024},
  doi     = {10.1128/AEM.68.3.1015-1024.2002}
}

@article{gruberWarmingTurningSour2011,
  title   = {Warming up, turning sour, losing breath: ocean biogeochemistry under global change},
  volume  = {369},
  number  = {1943},
  journal = {Philosophical Transactions of the Royal Society A: Mathematical, Physical and Engineering Sciences},
  author  = {Gruber, Nicolas},
  year    = {2011},
  pages   = {1980--1996},
  doi     = {10.1098/rsta.2011.0003}
}

@article{moraBioticHumanVulnerability2013,
  title   = {Biotic and Human Vulnerability to Projected Changes in Ocean Biogeochemistry over the 21st Century},
  volume  = {11},
  number  = {10},
  journal = {{PLoS} Biology},
  author  = {Mora, Camilo and Wei, Chih-Lin and Rollo, Audrey and Amaro, Teresa and Baco, Amy R. and Billett, David and Bopp, Laurent and Chen, Qi and Collier, Mark and Danovaro, Roberto and Gooday, Andrew J. and Grupe, Benjamin M. and Halloran, Paul R. and Ingels, Jeroen and Jones, Daniel O. B. and Levin, Lisa A. and Nakano, Hideyuki and Norling, Karl and Ramirez-Llodra, Eva and Rex, Michael and Ruhl, Henry A. and Smith, Craig R. and Sweetman, Andrew K. and Thurber, Andrew R. and Tjiputra, Jerry F. and Usseglio, Paolo and Watling, Les and Wu, Tongwen and Yasuhara, Moriaki},
  year    = {2013},
  month   = oct,
  pages   = {e1001682},
  doi     = {10.1371/journal.pbio.1001682}
}

@article{terhaarAssessmentGlobalOceanBiogeochemistry2024,
  title   = {Assessment of Global Ocean Biogeochemistry Models for Ocean Carbon Sink Estimates in {RECCAP2} and Recommendations for Future Studies},
  volume  = {16},
  number  = {3},
  journal = {Journal of Advances in Modeling Earth Systems},
  author  = {Terhaar, Jens and Goris, Nadine and Müller, Jens D. and DeVries, Tim and Gruber, Nicolas and Hauck, Judith and Pérez, Fiz F. and Séférian, Roland},
  year    = {2024},
  month   = mar,
  pages   = {e2023MS003840},
  doi     = {10.1029/2023MS003840}
}

@article{hagstromImpactDynamicPhytoplanktonStoichiometry2024,
  title   = {Impact of Dynamic Phytoplankton Stoichiometry on Global Scale Patterns of Nutrient Limitation, Nitrogen Fixation, and Carbon Export},
  volume  = {38},
  number  = {5},
  journal = {Global Biogeochemical Cycles},
  author  = {Hagstrom, George I. and Stock, Charles A. and Luo, Jessica Y. and Levin, Simon A.},
  year    = {2024},
  month   = may,
  pages   = {e2023GB007991},
  doi     = {10.1029/2023GB007991}
}

@article{fuEvaluationOceanBiogeochemistryCMIP2022,
  title   = {Evaluation of Ocean Biogeochemistry and Carbon Cycling in {CMIP} Earth System Models With the International Ocean Model Benchmarking ({IOMB}) Software System},
  volume  = {127},
  number  = {10},
  journal = {Journal of Geophysical Research: Oceans},
  author  = {Fu, Weiwei and Moore, J. Keith and Primeau, François and Collier, Nathan and Ogunro, Oluwaseun O. and Hoffman, Forrest M. and Randerson, James T.},
  year    = {2022},
  pages   = {e2022JC018965},
  doi     = {10.1029/2022JC018965}
}

@article{payne_uncertainties_projecting_2016,
  title   = {Uncertainties in projecting climate-change impacts in marine ecosystems},
  volume  = {73},
  number  = {5},
  journal = {ICES Journal of Marine Science},
  author  = {Payne, Mark R. and Barange, Manuel and Cheung, William W. L. and MacKenzie, Brian R. and Batchelder, Harold P. and Cormon, Xochitl and Eddy, Tyler D. and Fernandes, Jose A. and Hollowed, Anne B. and Jones, Miranda C. and Link, Jason S. and Neubauer, Philipp and Ortiz, Ivonne and Queirós, Ana M. and Paula, José Ricardo},
  year    = {2016},
  month   = may,
  pages   = {1272--1282},
  doi     = {10.1093/icesjms/fsv231}
}

@article{frolicher_sources_uncertainties_2016,
  title   = {Sources of uncertainties in 21st century projections of potential ocean ecosystem stressors},
  volume  = {30},
  number  = {8},
  journal = {Global Biogeochemical Cycles},
  author  = {Frölicher, Thomas L. and Rodgers, Keith B. and Stock, Charles A. and Cheung, William W. L.},
  year    = {2016},
  month   = aug,
  pages   = {1224--1243},
  doi     = {10.1002/2015GB005338}
}

@article{tagliabueHowWellGlobal2016,
  title   = {How well do global ocean biogeochemistry models simulate dissolved iron distributions?},
  volume  = {30},
  number  = {2},
  journal = {Global Biogeochemical Cycles},
  author  = {Tagliabue, Alessandro and Aumont, Olivier and DeAth, Ros and Dunne, John P. and Dutkiewicz, Stephanie and Galbraith, Eric and Misumi, Kazuhiro and Moore, J. Keith and Ridgwell, Andy and Sherman, Elliot and Stock, Charles and Vichi, Marcello and V{\"o}lker, Christoph and Yool, Andrew},
  year    = {2016},
  month   = feb,
  pages   = {149--174},
  doi     = {10.1002/2015GB005289}
}

@misc{suttonBitterLesson2019,
  author       = {Sutton, Richard S.},
  title        = {The Bitter Lesson},
  year         = {2019},
  howpublished = {Incomplete Ideas (blog essay)},
  url          = {http://www.incompleteideas.net/IncIdeas/BitterLesson.html}
}

\end{document}